\documentclass[twoside,11pt]{article}

\usepackage[utf8]{inputenc} %
\usepackage[T1]{fontenc}    %
\usepackage{url}            %
\usepackage{booktabs}       %
\usepackage{mathtools,amssymb}
\usepackage{amsfonts}       %
\usepackage{nicefrac}       %
\usepackage{microtype}      %
\usepackage{pgfplots,pgfplotstable}
\pgfplotsset{compat=1.14}
\usepackage{array,colortbl}
\usepackage{xcolor}
\usepackage{algorithm,algorithmicx,algpseudocode}
\usepackage{graphbox}
\usepackage{placeins}
\usepackage{wrapfig}
\usepackage{subcaption}
\usepackage{etoolbox}
\usepackage{makecell}
\usepackage{enumitem}
\usepackage[export]{adjustbox}
\usepackage{xspace}

\usepackage[abbrvbib, preprint]{jmlr2e}
\usepackage[capitalise]{cleveref}

\newtoggle{hqfigures}
\toggletrue{hqfigures}

\newcommand{\defeq}{\coloneqq}
\newcommand{\grad}{\nabla}
\newcommand{\E}{\mathbb{E}}

\newcommand{\Eb}[2]{\E_{#1}\!\left[#2\right]}

\newcommand{\kl}[2]{D_{\mathrm{KL}}\!\left(#1 ~ \| ~ #2\right)}

\newcommand{\bI}{\mathbf{I}}

\newcommand{\bzero}{\mathbf{0}}

\newcommand{\bc}{\mathbf{c}}

\newcommand{\bx}{\mathbf{x}}

\newcommand{\bz}{\mathbf{z}}

\newcommand{\bepsilon}{{\boldsymbol{\epsilon}}}
\newcommand{\bmu}{{\boldsymbol{\mu}}}

\newcommand{\bSigma}{{\boldsymbol{\Sigma}}}

\def\eg{{\em e.g.,}\xspace}

\newcommand{\website}{\url{https://cascaded-diffusion.github.io/}}

\jmlrheading{1}{2021}{1-48}{4/00}{10/00}{meila00a}{Jonathan Ho, Chitwan Saharia, William Chan, David J. Fleet, Mohammad Norouzi, Tim Salimans}

\ShortHeadings{Cascaded Diffusion Models}{Ho, Saharia, Chan, Fleet, Norouzi and Salimans}
\firstpageno{1}

\begin{document}

\title{Cascaded Diffusion Models \\ for High Fidelity Image Generation}

\author{\name Jonathan Ho\thanks{Equal contribution} \email jonathanho@google.com
       \AND
       \name Chitwan Saharia\footnotemark[1] \email sahariac@google.com
       \AND
       \name William Chan \email williamchan@google.com
       \AND
       \name David J. Fleet \email davidfleet@google.com
       \AND
       \name Mohammad Norouzi \email mnorouzi@google.com
       \AND
       \name Tim Salimans \email salimans@google.com
       \AND
       \addr Google, 1600 Amphitheatre Parkway, Mountain View, CA 94043
    }

\editor{TODO editor}

\maketitle

\begin{abstract}%
We show that cascaded diffusion models are capable of generating high fidelity images on the class-conditional ImageNet generation benchmark, without any assistance from auxiliary image classifiers to boost sample quality. A cascaded diffusion model comprises a pipeline of multiple diffusion models that generate images of increasing resolution, beginning with a standard diffusion model at the lowest resolution, followed by one or more super-resolution diffusion models that successively upsample the image and add higher resolution details. We find that the sample quality of a cascading pipeline relies crucially on conditioning augmentation, our proposed method of data augmentation of the lower resolution conditioning inputs to the super-resolution models. Our experiments show that conditioning augmentation prevents compounding error during sampling in a cascaded model, helping us to train cascading pipelines achieving FID scores of 1.48 at 64$\times$64, 3.52 at 128$\times$128 and 4.88 at 256$\times$256 resolutions, outperforming BigGAN-deep, and classification accuracy scores of 63.02\% (top-1) and 84.06\% (top-5) at 256$\times$256, outperforming VQ-VAE-2.
\end{abstract}


\begin{figure}[h!]
\vspace*{-1.5em}
\small
\begin{center}
\input{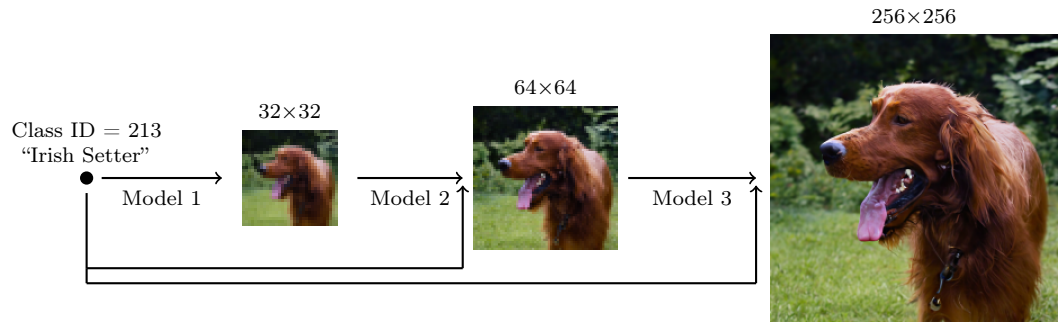}
\end{center}
\vspace*{-0.4em}
\caption{\small A cascaded diffusion model comprising a base model and two super-resolution models.}
\vspace*{-0.1cm}
\label{fig:cascade_fig}
\end{figure}

\section{Introduction}

\begin{figure} \centering
\setlength{\tabcolsep}{0pt}
\renewcommand{\arraystretch}{0} 
\begin{tabular}{cccccc}
\includegraphics[width=0.165\textwidth]{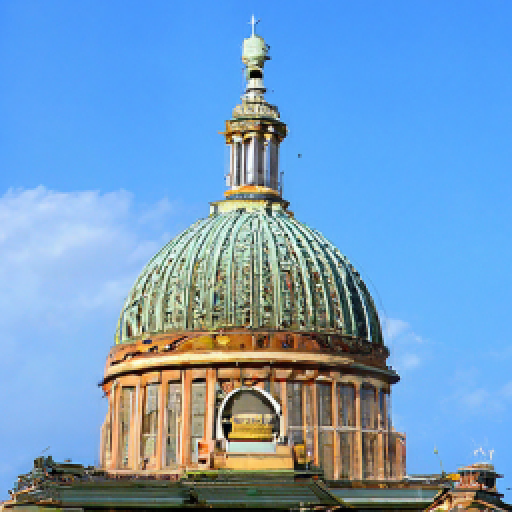} &
\includegraphics[width=0.165\textwidth]{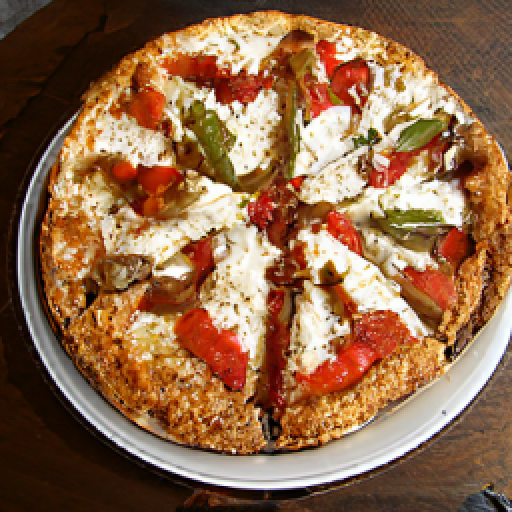} & 
\includegraphics[width=0.165\textwidth]{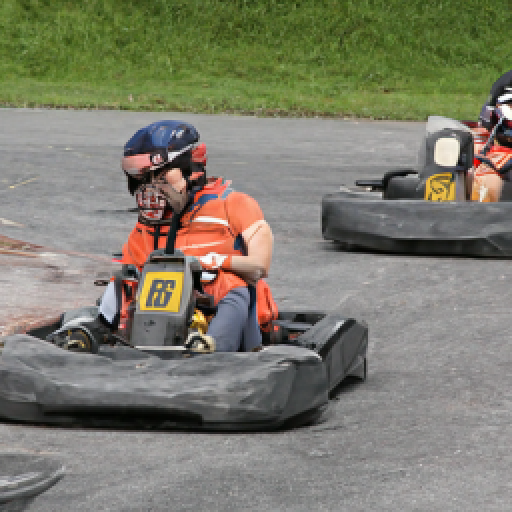} & 
\includegraphics[width=0.165\textwidth]{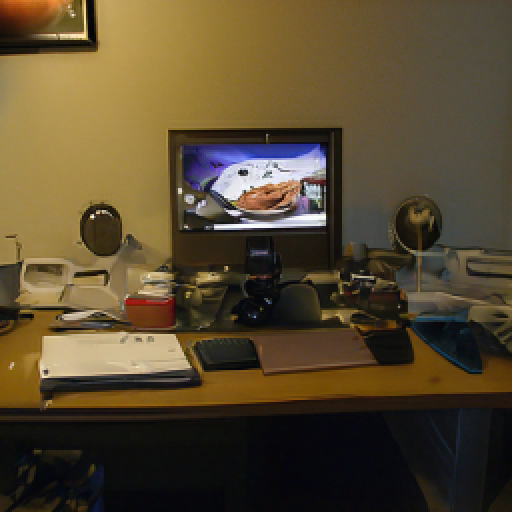} & 
\includegraphics[width=0.165\textwidth]{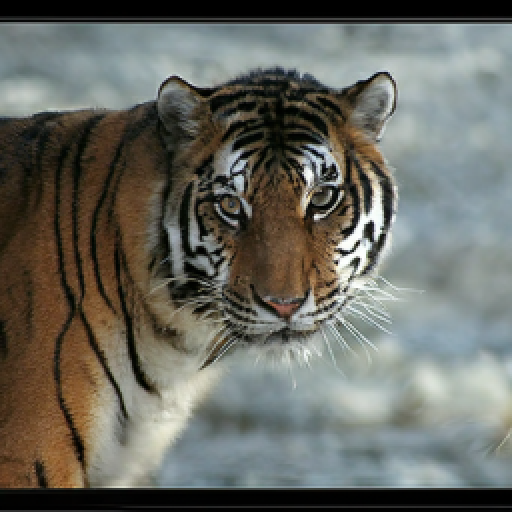} & 
\includegraphics[width=0.165\textwidth]{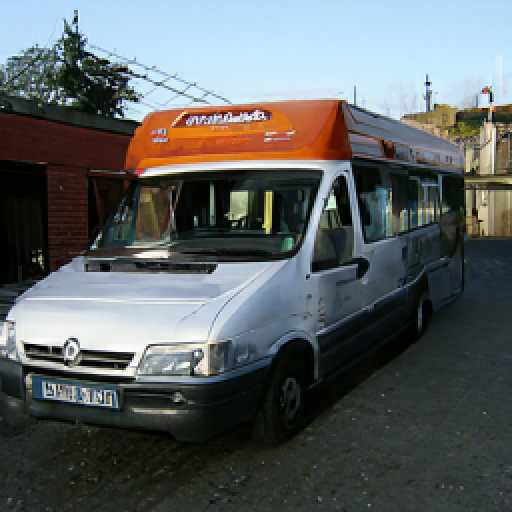} \\

\includegraphics[width=0.165\textwidth]{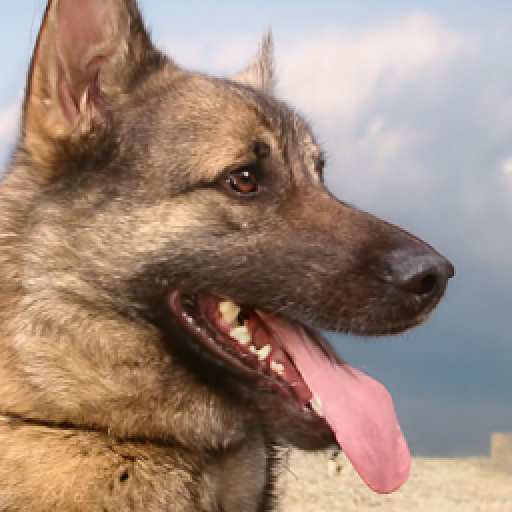} &
\includegraphics[width=0.165\textwidth]{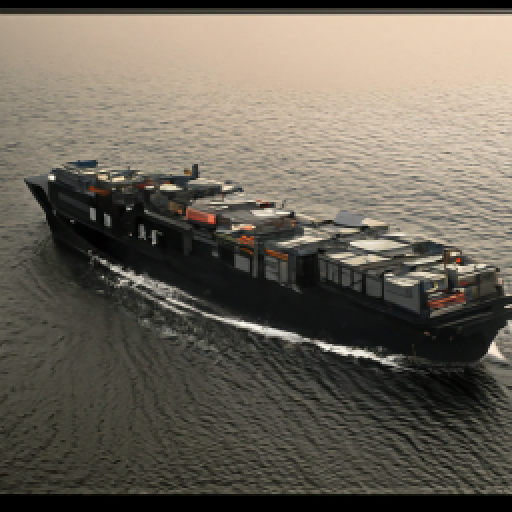} & 
\includegraphics[width=0.165\textwidth]{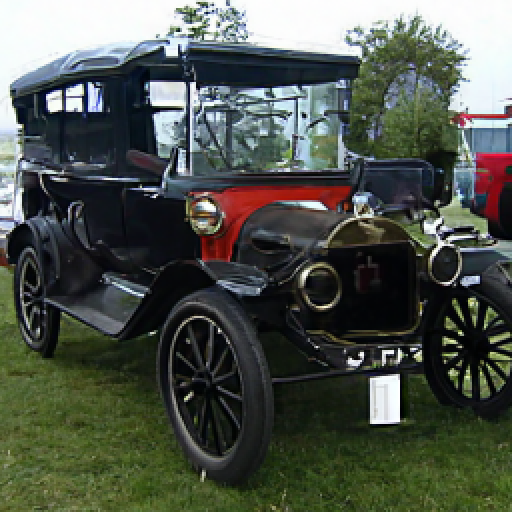} & 
\includegraphics[width=0.165\textwidth]{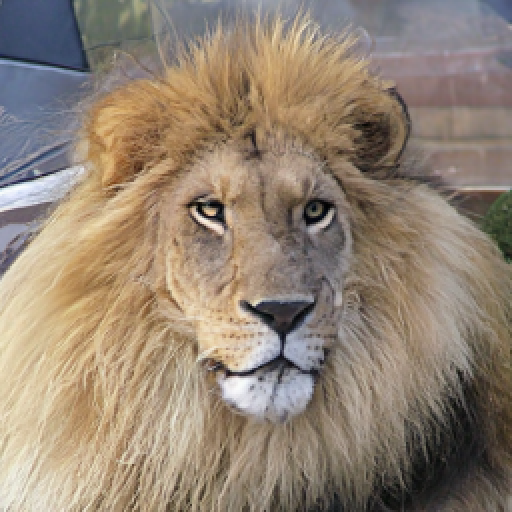} & 
\includegraphics[width=0.165\textwidth]{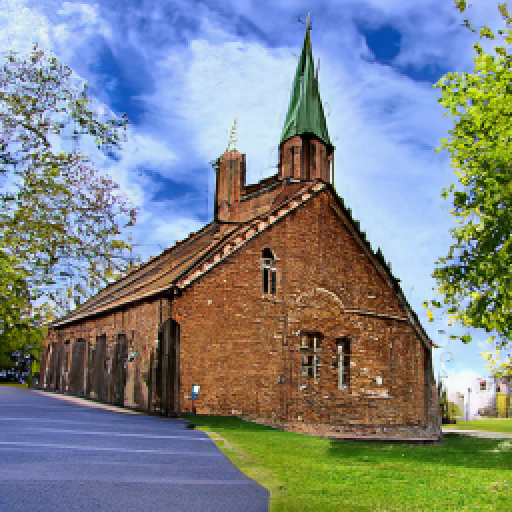} & 
\includegraphics[width=0.165\textwidth]{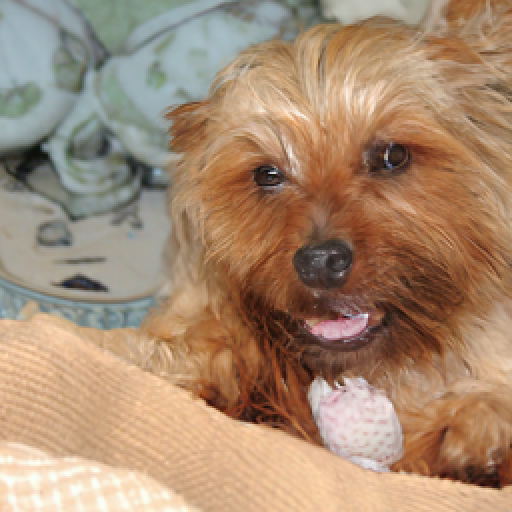} \\

\includegraphics[width=0.165\textwidth]{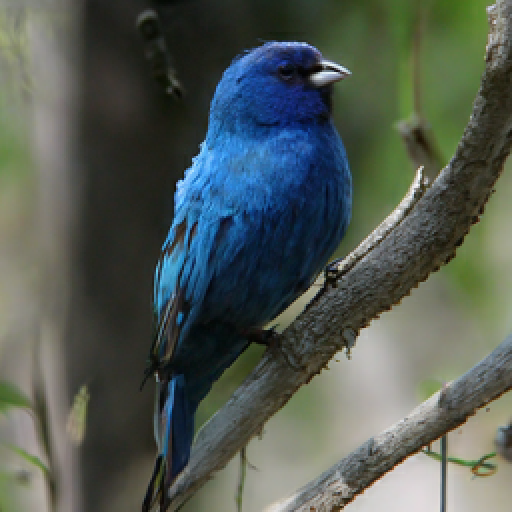} &
\includegraphics[width=0.165\textwidth]{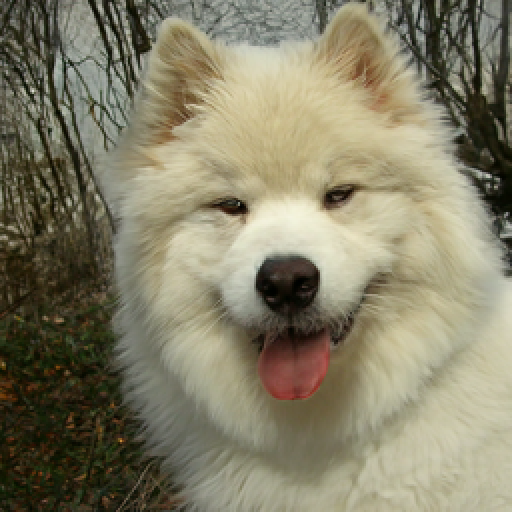} & 
\includegraphics[width=0.165\textwidth]{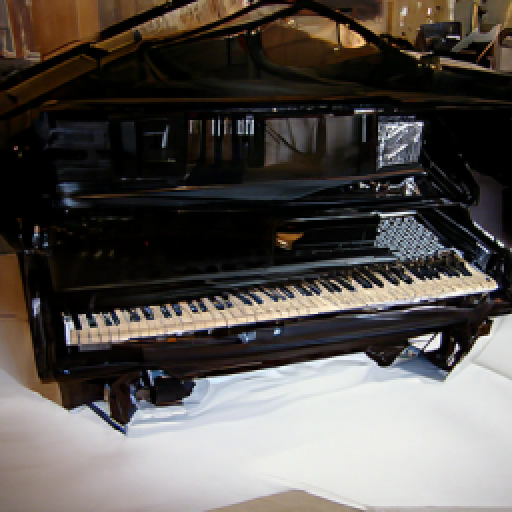} & 
\includegraphics[width=0.165\textwidth]{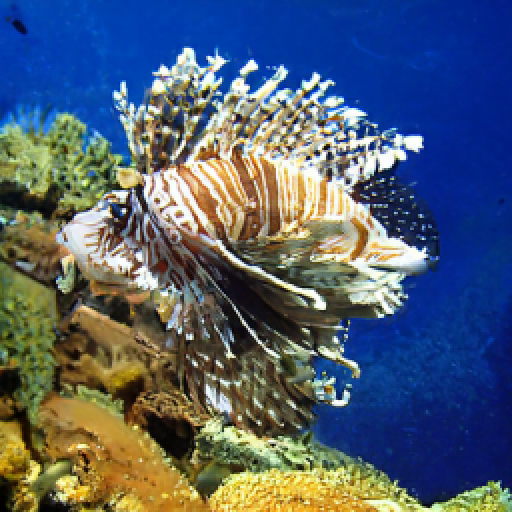} & 
\includegraphics[width=0.165\textwidth]{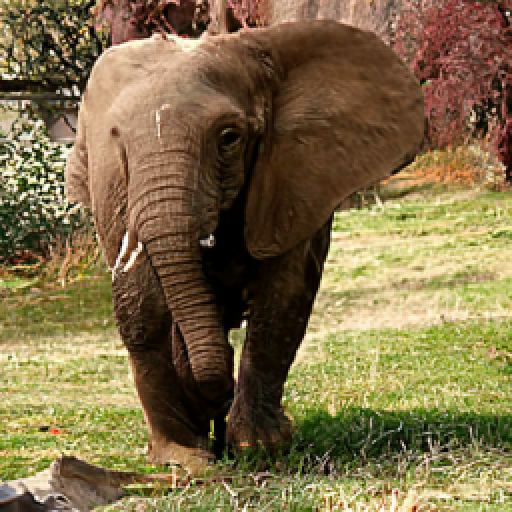} & 
\includegraphics[width=0.165\textwidth]{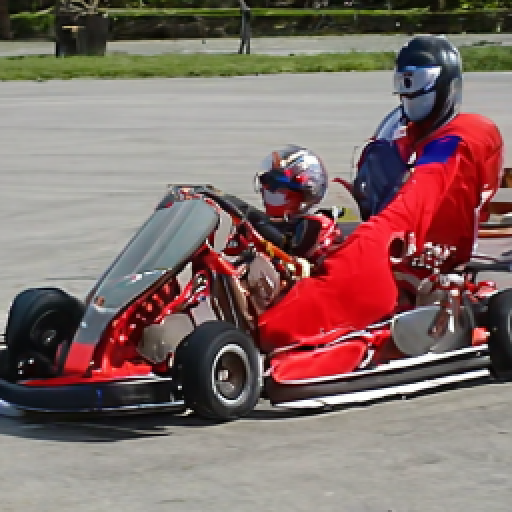} \\
\end{tabular}
\vspace*{-0.2cm}
\label{fig:imagenet_256x_montage}
\caption{Selected synthetic 256$\times$256 ImageNet samples.
}
\end{figure}

Diffusion models~\citep{sohl2015deep} have recently been shown to be capable of synthesizing high quality images and audio~\citep{chen2020wavegrad,ho2020denoising,kong2020diffwave,song2020improved}: an application of machine learning that has long been dominated by other classes of generative models such as autoregressive models, GANs, VAEs, and flows~\citep{brock2018large,dinh2016density,goodfellow2014generative,ho2019flow++,kingma2018glow,kingma2013auto,razavi2019generating,oord2016wavenet,oord2016pixel,oord2017neural}. Most previous work on diffusion models demonstrating high quality samples has focused on data sets of modest size, or data with strong conditioning signals.  
Our goal is to improve the sample quality of diffusion models on large high-fidelity data sets for which no strong conditioning information is available. 
To showcase the capabilities of the original diffusion formalism, we focus on simple, straightforward techniques to improve the sample quality of diffusion models; for example, we avoid using extra image classifiers to boost sample quality metrics~\citep{dhariwal2021diffusion,razavi2019generating}.

Our key contribution is the use of \emph{cascades} to improve the sample quality of diffusion models on class-conditional ImageNet .
Here, cascading refers to a simple technique to model high resolution data by learning a pipeline of separately trained models at multiple resolutions; a base model  generates low resolution samples, followed by super-resolution models  that upsample low resolution samples into high resolution samples. 
Sampling from a cascading pipeline occurs sequentially, first sampling from the low resolution base model, followed by sampling from super-resolution models in order of increasing resolution. 
While any type of generative model could be used in a cascading pipeline~\citep[\eg][]{menick2018generating,razavi2019generating},
here we restrict ourselves to diffusion models. Cascading has been shown in recent prior work to improve the sample quality of diffusion models~\citep{saharia2021image,nichol2021improved}; our work here concerns the improvement of diffusion cascading pipelines to attain the best possible sample quality.

The simplest and most effective technique we found to improve cascading diffusion pipelines is to apply strong data augmentation to the conditioning input of each super-resolution model. 
We refer to this technique as \emph{conditioning augmentation}. In our experiments, conditioning augmentation is crucial for our cascading pipelines to generate high quality samples at the highest resolution.
With this approach we attain FID scores on class-conditional ImageNet generation that are better than BigGAN-Deep~\citep{brock2018large} at any truncation value, and classification accuracy scores that are better than \mbox{VQ-VAE-2}~\citep{razavi2019generating}. We empirically find that conditioning augmentation is effective because it alleviates compounding error in cascading pipelines due to train-test mismatch, sometimes referred to as exposure bias in the sequence modeling literature~\citep{bengio2015scheduled,ranzato2015sequence}.

The key contributions of this paper are as follows:
\begin{itemize} %
    \item We show that our \textbf{C}ascaded \textbf{D}iffusion \textbf{M}odels (CDM) yield high fidelity samples superior to BigGAN-deep~\citep{brock2018large} and VQ-VAE-2~\citep{razavi2019generating} in terms of FID score~\citep{heusel2017gans} and classification accuracy score~\citep{ravuris2019cas}, the latter by a large margin.
    We achieve these results with pure generative models that are not combined with any classifier.
    \item We introduce conditioning augmentation for our super-resolution models, and find it critical towards achieving high sample fidelity. We perform an in-depth exploration of augmentation policies, and find Gaussian augmentation to be a key ingredient for low resolution upsampling, and Gaussian blurring for high resolution upsampling. We also show how to efficiently train models amortized over varying levels of conditioning augmentation to enable post-training hyperparameter search for optimal sample quality.
\end{itemize}

\Cref{sec:background} reviews recent work on diffusion models. \Cref{sec:cond_aug} describes the most effective types of conditioning augmentation that we found for class-conditional ImageNet generation. \Cref{sec:experiments} contains our sample quality results, ablations, and experiments on additional datasets. \Cref{appendix:samples} contains extra samples and \Cref{appendix:hyperparams} contains details on hyperparameters and architectures. High resolution figures and additional supplementary material can be found at \website.

\section{Background} \label{sec:background}

We begin with background on diffusion models, their extension to conditional generation, and their associated neural network architectures.

\subsection{Diffusion Models} A diffusion model~\citep{sohl2015deep,ho2020denoising} is defined by a forward process that gradually destroys data $\bx_0 \sim q(\bx_0)$ over the course of $T$ timesteps
\begin{align*}
    q(\bx_{1:T}|\bx_0) &= \prod_{t=1}^T q(\bx_{t} | \bx_{t-1}), \quad \quad
    q(\bx_{t} | \bx_{t-1}) = \mathcal{N}(\bx_{t}; \sqrt{1-\beta_t}\bx_{t-1}, \beta_t \bI)
\end{align*}
and a parameterized reverse process $p_\theta(\bx_0) = \int p_\theta(\bx_{0:T}) \, d \bx_{1:T}$, where
\begin{align*}
    p_\theta(\bx_{0:T}) &= p(\bx_T) \prod_{t=1}^T p_\theta(\bx_{t-1}|\bx_t) , \quad p_\theta(\bx_{t-1}|\bx_t) = \mathcal{N}(\bx_{t-1}; \bmu_\theta(\bx_{t},t), \bSigma_\theta(\bx_{t},t)).
\end{align*}
The forward process hyperparameters $\beta_t$ are set so that $\bx_T$ is approximately distributed according to a standard normal distribution, so $p(\bx_T)$ is set to a standard normal prior as well.
The reverse process is trained to match the joint distribution of the forward process by optimizing the evidence lower bound (ELBO) $-L_\theta(\bx_0) \leq \log p_\theta(\bx_0)$:
\begin{align}
    L_\theta(\bx_0) &= \Eb{q}{ 
    L_T(\bx_0) + \sum_{t > 1} \kl{q(\bx_{t-1} | \bx_t, \bx_0)}{p_\theta(\bx_{t-1}|\bx_t)} - \log p_\theta(\bx_0 | \bx_1) } \label{eq:elbo}
\end{align}
where $L_T(\bx_0) = \kl{q(\bx_T|\bx_0)}{p(\bx_T)}$.
The forward process posteriors $q(\bx_{t-1}|\bx_t,\bx_0)$ and marginals $q(\bx_t | \bx_0)$ are Gaussian, and the KL divergences in the ELBO can be calculated in closed form. Thus it is possible to train the diffusion model by taking stochastic gradient steps on random terms of \cref{eq:elbo}. As previously suggested~\citep{ho2020denoising,nichol2021improved}, we use the reverse process parameterizations
\begin{eqnarray*}
    \bmu_\theta(\bx_{t},t) &=& \frac{1}{\sqrt{\alpha_t}}\left( \bx_t - \frac{\beta_t}{\sqrt{1-\bar\alpha_t}}\bepsilon_\theta(\bx_t,t) \right) \\
    \Sigma^{ii}_\theta(\bx_{t},t) &=& \exp(\log \tilde\beta_t + (\log \beta_t - \log \tilde\beta_t) v^i_\theta(\bx_t,t))
\end{eqnarray*}
where $\alpha_t=1-\beta_t$, $\bar\alpha_t = \prod_{s=1}^t \alpha_s$, and $\tilde\beta_t = \frac{1-\bar\alpha_{t-1}}{1-\bar\alpha_t}\beta_t$.

Sample quality can be improved, at the cost of log likelihood, by optimizing modified losses instead of the ELBO. The particular form of the modified loss depends on whether we are learning $\bSigma_\theta$ or treating it as a fixed hyperparameter (and whether $\bSigma_\theta$ is learned is itself considered a hyperparameter choice that we set experimentally). For the case of non-learned $\bSigma_\theta$, we use the simplified loss
\begin{align*}
    L_\mathrm{simple}(\theta) = \Eb{\bx_0,\bepsilon\sim \mathcal{N}(\bzero, \bI),t \sim \mathcal{U}(\{1, \dotsc, T\})}{ \left\| \bepsilon_\theta(\sqrt{\bar\alpha_t}\bx_0 + \sqrt{1-\bar\alpha_t}\bepsilon,t) - \bepsilon \right\|^2}
\end{align*}
which is a weighted form of the ELBO that resembles denoising score matching over multiple noise scales~\citep{ho2020denoising,song2019generative}. For the case of learned $\bSigma_\theta$, we employ a hybrid loss~\citep{nichol2021improved} implemented using the expression
\begin{align*}
    L_\mathrm{hybrid}(\theta) = L_\mathrm{simple}(\theta) + \lambda L_\mathrm{vb}(\theta)
\end{align*}
where $L_\mathrm{vb} = \Eb{\bx_0}{L_\theta(\bx_0)}$ and a stop-gradient is applied to the $\bepsilon_\theta$ term inside $L_\theta$. Optimizing this hybrid loss has the effect of simultaneously learning $\bmu_\theta$ using $L_\mathrm{simple}$ and learning $\bSigma_\theta$ using the ELBO.

\subsection{Conditional Diffusion Models} \label{sec:background_conditioning}
In the conditional generation setting, the data $\bx_0$ has an associated conditioning signal $\bc$, for example a label in the case of class-conditional generation, or a low resolution image in the case of super-resolution~\citep{saharia2021image,nichol2021improved}. The goal is then to learn a conditional model $p_\theta(\bx_0|\bc)$. To do so, we modify the diffusion model to include $\bc$ as input to the reverse process:
\begin{align*}
  p_\theta(\bx_{0:T} | \bc) &= p(\bx_T) \prod_{t=1}^T p_\theta(\bx_{t-1}|\bx_t,\bc), \quad p_\theta(\bx_{t-1}|\bx_t,\bc) = \mathcal{N}(\bx_{t-1}; \bmu_\theta(\bx_{t},t,\bc), \bSigma_\theta(\bx_{t},t,\bc)) \nonumber \\
  L_\theta(\bx_0|\bc) &= \Eb{q}{ L_T(\bx_0) + \sum_{t > 1} \kl{q(\bx_{t-1} | \bx_t, \bx_0)}{p_\theta(\bx_{t-1}|\bx_t,\bc)} - \log p_\theta(\bx_0 | \bx_1,\bc) }.
\end{align*}
The data and conditioning signal $(\bx_0, \bc)$ are sampled jointly from the data distribution, now called $q(\bx_0, \bc)$, and the forward process $q(\bx_{1:T}|\bx_0)$ remains unchanged. The only modification that needs to be made is to inject $\bc$ as a extra input to the neural network function approximators: instead of $\bmu_\theta(\bx_t,t)$ we now have $\bmu_\theta(\bx_t,t,\bc)$, and likewise for $\bSigma_\theta$. The particular architectural choices for injecting these extra inputs depends on the type of the conditioning $\bc$, as described next.

\subsection{Architectures} 
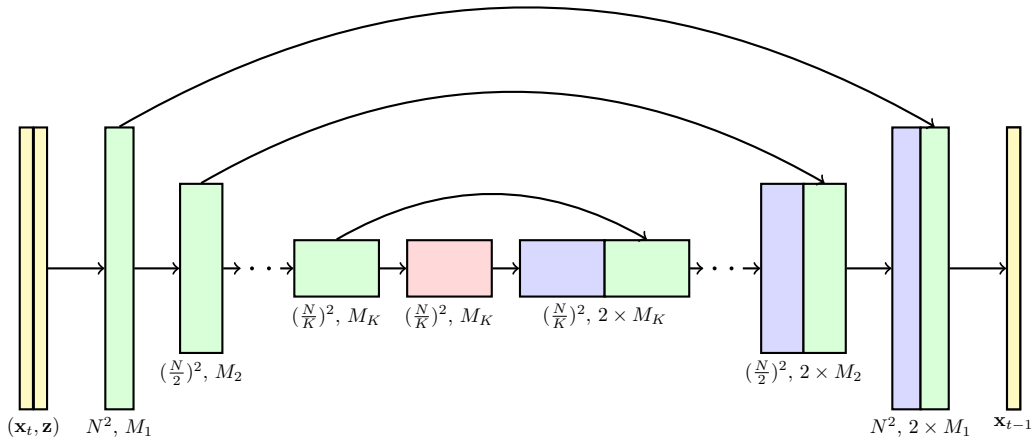
\begin{figure}[htb]
\vspace*{-1.5em}
\small
\begin{center}
\begin{tikzpicture}[
scale=0.75, every node/.style={scale=0.75},
dot/.style = {circle, fill, minimum size=#1,
              inner sep=0pt, outer sep=0pt},
dot/.default = 2pt]

\tikzstyle{layer} = [rectangle, thick, minimum width=4cm, minimum height=0.5cm, align=center, draw=black]
\tikzstyle{arrow} = [thick,->]

\node[fill=yellow!30] (input_high_res1) at (-1.9,0) [layer,thick,minimum width=0.1cm,minimum height=5cm] {};
\node[label=below:{$(\bx_t, \bz)$}, text width=0.0cm, minimum height=5cm] (temp) at (-1.775,0)  {};
\node[label, text width=0.0cm, minimum height=5cm] (temp) at (-1.775,0)  {};
\node[fill=yellow!30] (input_low_res1) at (-1.65,0) [layer,thick,minimum width=0.1cm,minimum height=5cm] {};
\node[fill=green!15, label=below:{$N^2$, $M_1$}] (first_res_d) at (-0.25,0) [layer,thick,minimum width=0.5cm,minimum height=5cm] {};
\draw[arrow] (input_low_res1.east) -- (first_res_d.west); 
\node[fill=green!15, label=below:{$(\frac{N}{2})^2$, $M_2$}] (second_res_d) at (1.2,0) [layer,thick,minimum width=0.75cm,minimum height=3cm] {};
\draw[arrow] (first_res_d.east) -- (second_res_d.west); 
\draw[arrow] (second_res_d.east) -- (1.9,0); 
\node[dot] at (2.1,0) {};
\node[dot] at (2.4,0) {};
\node[fill=green!15, label=below:{$(\frac{N}{K})^2$, $M_K$}] (last_res_d) at (3.6,0) [layer,thick,minimum width=1.5cm,minimum height=1cm] {};
\draw[arrow] (2.6,0) -- (last_res_d.west); 
\node[fill=red!15, label=below:{$(\frac{N}{K})^2$, $M_K$}] (middle_res) at (5.6,0) [layer,thick,minimum width=1.5cm,minimum height=1cm] {};
\draw[arrow] (last_res_d.east) -- (middle_res.west); 
\node[fill=blue!15] (first_res_u) at (7.6,0) [layer,thick,minimum width=1.5cm,minimum height=1cm] {};
\node[label=below:{$(\frac{N}{K})^2$, $2 \times M_K$}, text width=0.0cm, minimum height=1cm] (temp1) at (8.35,0)  {};
\node[fill=green!15] (last_res_d_concat) at (9.1,0) [layer,thick,minimum width=1.5cm,minimum height=1cm] {};
\draw[arrow] (middle_res.east) -- (first_res_u.west); 

\node[dot] at (10.3,0) {};
\node[dot] at (10.6,0) {};
\draw[arrow] (last_res_d_concat.east) -- (10.1, 0);

\node[fill=blue!15] (second_res_u) at (11.5,0) [layer,thick,minimum width=0.75cm,minimum height=3cm] {};
\node[label=below:{$(\frac{N}{2})^2$, $2 \times M_2$}, text width=0.0cm, minimum height=3cm] (temp2) at (11.875,0)  {};
\node[fill=green!15] (second_res_d_concat) at (12.25,0) [layer,thick,minimum width=0.75cm,minimum height=3cm] {};
\draw[arrow] (10.8,0) -- (second_res_u.west);
\node[fill=blue!15] (last_res_u) at (13.7,0) [layer,thick,minimum width=0.5cm,minimum height=5cm] {};
\node[label=below:{$N^2$, $2 \times M_1$}, text width=0.0cm, minimum height=5cm] (temp3) at (13.95,0)  {};
\node[fill=green!15] (first_res_d_concat) at (14.2,0) [layer,thick,minimum width=0.5cm,minimum height=5cm] {};
\draw[arrow] (second_res_d_concat.east) -- (last_res_u.west);
\node[label=below:$\bx_{t-1}$,fill=blue!20] (output) at (15.6,0) [layer,thick,minimum width=0.15cm,minimum height=5cm] {};
\node[label,fill=yellow!30] (output) at (15.6,0) [layer,thick,minimum width=0.15cm,minimum height=5cm] {};
\draw[arrow] (first_res_d_concat.east) -- (output.west);

\draw[arrow] (first_res_d.north) to [out=30,in=150] (first_res_d_concat.north);
\draw[arrow] (second_res_d.north) to [out=30,in=150] (second_res_d_concat.north);
\draw[arrow] (last_res_d.north) to [out=30,in=150] (last_res_d_concat.north);

\end{tikzpicture}
\end{center}
\vspace*{-0.4em}
\caption{\small The U-Net architecture used in each model of a CDM pipeline. The first model is a class-conditional diffusion model that receives the noisy image $\bx_t$ and the class label $y$ and as input. (The class label $y$ and timestep $t$ are injected into each block as an embedding, not depicted here). The remaining models in the pipeline are class-conditional super-resolution models that receive $\bx_t$, $y$, and an additional upsampled low-resolution image $\bz$ as input. The downsampling/upsampling blocks adjust the image input resolution $N\times N$ by a factor of 2 through each of the $K$ blocks. The channel count at each block is specified using channel multipliers $M_1$, $M_2$, ..., $M_K$, and the upsampling pass has concatenation skip connections to the downsampling pass.}
\vspace*{-0.1cm}
\label{fig:unet_fig}
\end{figure}

The current best architectures for image diffusion models are U-Nets~\citep{ronneberger2015unet,salimans2017pixelcnn++}, which are a natural choice to map corrupted data $\bx_t$ to reverse process parameters $(\bmu_\theta, \bSigma_\theta)$ that have the same spatial dimensions as $\bx_t$. Scalar conditioning, such as a class label or a diffusion timestep $t$, is provided by adding embeddings into intermediate layers of the network~\citep{ho2020denoising}. Lower resolution image conditioning is provided by channelwise concatenation of the low resolution image, processed by bilinear or bicubic upsampling to
the desired resolution, with the reverse process input $\bx_t$, as in the SR3~\citep{saharia2021image} and Improved DDPM~\citep{nichol2021improved} models. See~\cref{fig:unet_fig} for an illustration of the SR3-based architecture that we use in this work.

\section{Conditioning Augmentation in Cascaded Diffusion Models}
\label{sec:cond_aug}

\begin{figure}[htb]
\vspace*{-1.5em}
\small
\begin{center}
\scalebox{.75}{\input{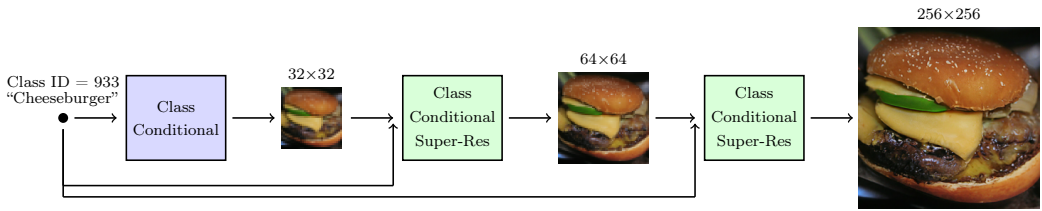}}
\end{center}
\vspace*{-1.0em}
\caption{\small Detailed CDM pipeline for generation of class conditional 256$\times$256 images. The first model is a class-conditional diffusion model, and it is followed by a sequence of two class-conditional super-resolution diffusion models. Each model has a U-Net architecture as depicted in~\cref{fig:unet_fig}.}
\vspace*{-0.1cm}
\label{fig:cascade_unet_fig}
\end{figure}

Suppose $\bx_0$ is high resolution data and $\bz_0$ is its low resolution counterpart. We use the term \emph{cascading pipeline} to refer to a sequence of generative models. At the low resolution we have a diffusion model $p_\theta(\bz_0)$, and at the high resolution, a super-resolution diffusion model $p_\theta(\bx_0|\bz_0)$. The cascading pipeline forms a latent variable model for high resolution data; i.e., $p_\theta(\bx_0) = \int p_\theta(\bx_0|\bz_0) p_\theta(\bz_0) \,d \bz_0$. It is straightforward to extend this to more than two resolutions. It is also straightforward to condition an entire cascading pipeline on class information or other conditioning information $\bc$: the models take on the form $p_\theta(\bz_0|\bc)$ and $p_\theta(\bx_0|\bz_0,\bc)$, each using the conditioning mechanism described in \cref{sec:background_conditioning}. An example cascading pipeline is depicted in~\cref{fig:cascade_unet_fig}.

Cascading pipelines have been shown to be useful with other generative model families~\citep{menick2018generating,razavi2019generating}. A major benefit to training a cascading pipeline over training a standard model at the highest resolution is that most of the modeling capacity can be dedicated to low resolutions, which empirically are the most important for sample quality, and  training and sampling at low resolutions tends to be the most computationally efficient. In addition, cascading allows the individual models to be trained independently, and architecture choices can be tuned at each specific resolution for the best performance of the entire pipeline.

The most effective technique we found to improve the sample quality of cascading pipelines is to train each  super-resolution model using data augmentation on its low resolution input. We refer to this general technique as \emph{conditioning augmentation}. At a high level, for some super-resolution model $p_\theta(\bx_0|\bz)$ from a low resolution image $\bz$ to a high resolution image $\bx_0$, conditioning augmentation refers to applying some form of data augmentation to $\bz$. This augmentation can take any form, but what we found most effective at low resolutions is adding Gaussian noise (forward process noise), and for high resolutions, randomly applying Gaussian blur to $\bz$. In some cases, we found it more practical to train super-resolution models amortized over the strength of conditioning augmentation and pick the best strength in a post-training hyperparameter search for optimal sample quality. Details on conditioning augmentation and its realization during training and sampling are given in the following sections.

\subsection{Blurring Augmentation}
\label{sec:blur_aug}

One simple instantiation of conditioning augmentation is augmentation of $\bz$ by blurring. We found this to be most effective for upsampling to images with resolution 128$\times$128 and 256$\times$256. More specifically, we apply a Gaussian filter of size $k$ and sigma $\sigma$ to obtain $\bz_b$. We use a filter size of $k = (3,3)$ and randomly sample $\sigma$ from a fixed range during training. We perform hyper-parameter search to find the range for $\sigma$. During training, we apply this blurring augmentation to 50\% of the examples. During inference, no augmentation is applied to low resolution inputs. We explored applying different amounts of blurring augmentations during inference, but did not find it helpful in initial experiments. 

\subsection{Truncated Conditioning Augmentation}
\label{sec:truncated_gaussian_aug}

Here we describe what we call \emph{truncated conditioning augmentation}, a form of conditioning augmentation that requires a simple modification to the training and architecture of the super-resolution models, but no change to the low resolution model at the initial stage of the cascade. We found this method to be most useful at resolutions smaller than 128$\times$128.
Normally, generating a high resolution sample $\bx_0$ involves first generating $\bz_0$ from the low resolution model $p_\theta(\bz_0)$, then feeding that result into the super-resolution model $p_\theta(\bx_0|\bz_0)$. In other words, generating a high resolution sample is performed using ancestral sampling from the latent variable model
\begin{align*}
    p_\theta(\bx_0) = \int p_\theta(\bx_0|\bz_0) p_\theta(\bz_0) \,d \bz_0 = \int p_\theta(\bx_0|\bz_0) p_\theta(\bz_{0:T}) \,d \bz_{0:T} .
\end{align*}
(For simplicity, we have assumed that the low resolution and super-resolution models both use the same number of timesteps $T$.) 
Truncated conditioning augmentation refers to truncating the low resolution reverse process to stop at timestep $s > 0$, instead of $0$; i.e.,
\begin{align}
    p^s_\theta(\bx_0) = \int p_\theta(\bx_0|\bz_s) p_\theta(\bz_s) \,d \bz_s = \int p_\theta(\bx_0|\bz_s) p_\theta(\bz_{s:T}) \,d \bz_{s:T} . \label{eq:p_s_theta}
\end{align}
The base model is now $p_\theta(\bz_s) = \int p_\theta(\bz_{s:T}) d\bz_{s+1:T}$, and the super-resolution model is now $p_\theta(\bx_0 | \bz_s) = \int p(\bx_T) \prod_{t=1}^T p_\theta(\bx_{t-1}|\bx_t,\bz_s) d\bx_{1:T}$, where
\begin{align*}
p_\theta(\bx_{t-1}|\bx_t,\bz_s) = \mathcal{N}(\bx_{t-1}; \bmu_\theta(\bx_{t},t,\bz_s,s), \bSigma_\theta(\bx_{t},t,\bz_s,s)).
\end{align*}
The reason truncating the low resolution reverse process is a form of data augmentation is that the training procedure for $p_\theta(\bx_0|\bz_s)$ involves conditioning on noisy $\bz_s \sim q(\bz_s|\bz_0)$, which, up to scaling, is $\bz_0$ augmented with Gaussian noise.
To be more precise about training a cascading pipeline with truncated conditioning augmentation, let us examine the ELBO for $p_\theta^s(\bx_0)$ in \cref{eq:p_s_theta}. We can treat $p_\theta^s(\bx_0)$ as a VAE with a diffusion model prior, a diffusion model decoder, and the approximate posterior
\begin{align*}
    q(\bx_{1:T}, \bz_{1:T} | \bx_0, \bz_0) = \prod_{t=1}^T q(\bx_t|\bx_{t-1}) q(\bz_t|\bz_{t-1}) ,
\end{align*}
which runs forward processes independently on a low and high resolution pair. The ELBO is
\begin{align*}
    -\log p^s_\theta(\bx_0) \leq \Eb{q}{ L_T(\bz_0)  + \sum_{t > s} \kl{q(\bz_{t-1} | \bz_t, \bz_0)}{p_\theta(\bz_{t-1}|\bz_t)} - \log p_\theta(\bx_0 | \bz_s) },
\end{align*}
where $L_T(\bz_0) = \kl{q(\bz_T|\bz_0)}{p(\bz_T)}$. Note that the sum over $t$ is truncated at $s$, and the decoder $p_\theta(\bx_0|\bz_s)$ is the super-resolution model conditioned on  $\bz_s$. The decoder itself has an ELBO of the form $-\log p_\theta(\bx_0|\bz_s) \leq L_\theta(\bx_0 | \bz_s)$, where
\begin{align*}
  L_\theta(\bx_0 | \bz_s) = \Eb{q}{ L_T(\bx_0) + \sum_{t > 1} \! \kl{q(\bx_{t-1} | \bx_t, \bx_0)}{p_\theta(\bx_{t-1}|\bx_t,\bz_s)} - \log p_\theta(\bx_0 | \bx_1,\bz_s) }.
\end{align*}
Thus we have an ELBO for the combined model
\begin{align}
    -\log p^s_\theta(\bx_0) \leq \Eb{q}{ L_T(\bz_0) + \sum_{t > s} \kl{q(\bz_{t-1} | \bz_t, \bz_0)}{p_\theta(\bz_{t-1}|\bz_t)} + L_\theta(\bx_0|\bz_s) }. \label{eq:combined_elbo}
\end{align}
It is apparent that optimizing \cref{eq:combined_elbo} trains the low and high resolution models separately. For a fixed value of $s$, the low resolution process is trained up to the truncation timestep $s$, and the super-resolution model is trained on a conditioning signal corrupted using the low resolution forward process stopped at timestep $s$.

In practice, since we pursue sample quality as our main objective, we do not use these ELBO expressions directly when training models with learnable reverse process variances. Rather, we train on the ``simple'' unweighted loss or the hybrid loss described in~\cref{sec:background}, and the particular loss we use is considered a hyperparameter reported in \cref{appendix:hyperparams}.

We would like to search over multiple values of $s$ to select for the best sample quality. To make this search practical, we avoid retraining models by amortizing a single super-resolution model over uniform random $s$ at training time.
Because each possible truncation time corresponds to a distinct super-resolution task, the super-resolution model for $\bmu_\theta$ and $\bSigma_\theta$ must take $\bz_s$ as input along with $s$, and this can be accomplished using a single network with an extra time embedding input for $s$. 
We leave the low resolution model training unchanged, because the standard diffusion training procedure already trains with random $s$.  The complete training procedure for a two-stage cascading pipeline is listed in \cref{alg:training}.

\subsection{Non-truncated Conditioning Augmentation}
\label{sec:non_truncated_gaussian_aug}

Another form of conditioning augmentation, which we call \emph{non-truncated conditioning augmentation}, 
uses the same model modifications and training procedure as truncated conditioning augmentation (\cref{sec:truncated_gaussian_aug}). The only difference is at sampling time. Instead of truncating the low resolution reverse process, in non-truncated conditioning augmentation we always sample $\bz_0$ using the full, non-truncated low resolution reverse process; then we corrupt $\bz_0$ using the forward process into $\bz'_s \sim q(\bz_s | \bz_0)$ and feed the corrupted $\bz'_s$ into the super-resolution model.

The main advantage of non-truncated conditioning augmentation over truncated conditioning augmentation is a practical one during the search phase over $s$. In the case of truncated augmentation, if we want to run the super-resolution model over all $s$ in parallel, we must store all low resolution samples $\bz_s$ for all values of $s$ considered. In the case of non-truncated augmentation, we need to store the low resolution samples just once, since sampling $\bz'_s \sim q(\bz_s | \bz_0)$ is computationally inexpensive. These sampling procedures are listed in~\cref{alg:sampling}.

Truncated and non-truncated conditioning augmentation should perform similarly because~$\bz_s$~and~$\bz'_s$ should have similar marginal distributions if the low resolution model is trained well enough. Indeed, in \cref{sec:cond_aug_ablations}, we empirically find that sample quality metrics are similar for both truncated and non-truncated conditioning augmentation.

\begin{algorithm}[tbp]
  \caption{Training a two-stage CDM with Gaussian conditioning augmentation} \label{alg:training}
  \small
  \begin{algorithmic}[1]
    \Repeat  \Comment{Train base model}
      \State $(\bz_0,\bc) \sim p(\bz,\bc)$ \Comment{Sample low-resolution image and label}
      \State $t \sim \mathcal{U}(\{1, \dotsc, T\})$
      \State $\bepsilon\sim\mathcal{N}(\bzero,\bI)$
      \State $\bz_t = \sqrt{\bar\alpha_t}\bz_0 + \sqrt{1-\bar\alpha_t}\bepsilon$
      \State $\theta \leftarrow \theta - \eta\grad_\theta \left\| \bepsilon_\theta(\bz_t,t,\bc) - \bepsilon \right\|^2$ \Comment{Simple loss (can be replaced with a hybrid loss)}
    \Until{converged}
    \Repeat \Comment{Train super-resolution model (in parallel with the base model)}
      \State $(\bx_0,\bz_0,\bc) \sim p(\bx,\bz,\bc)$ \Comment{Sample low- and high-resolution images and label}
      \State $s,t \sim \mathcal{U}(\{1, \dotsc, T\})$
      \State $\bepsilon_\bz, \bepsilon_\bx \sim\mathcal{N}(\bzero,\bI)$ \Comment{Note: $\bepsilon_\bz,\bepsilon_\bx$ should have the same shapes as $\bz_0,\bx_0$, respectively}
      \State $\bz_t = \sqrt{\bar\alpha_s}\bz_0 + \sqrt{1-\bar\alpha_s}\bepsilon_\bz$ \Comment{Apply Gaussian conditioning augmentation}
      \State $\bx_t = \sqrt{\bar\alpha_t}\bx_0 + \sqrt{1-\bar\alpha_t}\bepsilon_\bx$
      \State $\theta \leftarrow \theta - \eta\grad_\theta \left\| \bepsilon_\theta(\bx_t,t,\bz_s,s,\bc) - \bepsilon_\bx \right\|^2$
    \Until{converged}
  \end{algorithmic}
\end{algorithm}

\begin{algorithm}[tbp]
  \caption{Sampling from a two-stage CDM with Gaussian conditioning augmentation} \label{alg:sampling}
  \small
  \begin{algorithmic}[1]
    \Require $\bc$: class label
    \Require $s$: conditioning augmentation truncation time
    \State $\bz_{T} \sim \mathcal{N}(\bzero, \bI)$
    \If{using truncated conditioning augmentation}
      \For{$t=T,\dotsc,s+1$}
        \State $\bz_{t-1} \sim p_\theta(\bz_{t-1}|\bz_t,\bc)$
      \EndFor
    \Else
      \For{$t=T,\dotsc,1$}
        \State $\bz_{t-1} \sim p_\theta(\bz_{t-1}|\bz_t,\bc)$
      \EndFor
      \State $\bz_s \sim q(\bz_s | \bz_0)$  \Comment{Overwrite previously sampled value of $\bz_s$}
    \EndIf
    \State $\bx_{T} \sim \mathcal{N}(\bzero, \bI)$
    \For{$t=T,\dotsc,1$}
        \State $\bx_{t-1} \sim p_\theta(\bx_{t-1}|\bx_t,\bz_s,\bc)$
    \EndFor
    \State \textbf{return} $\bx_0$
  \end{algorithmic}
\end{algorithm}

\section{Experiments}
\label{sec:experiments}

\begin{figure}[p]
\vspace*{-1cm}
\setlength{\tabcolsep}{1.25pt}
\centering
\begin{tabular}{ccccc}
\includegraphics[width=0.2\textwidth]{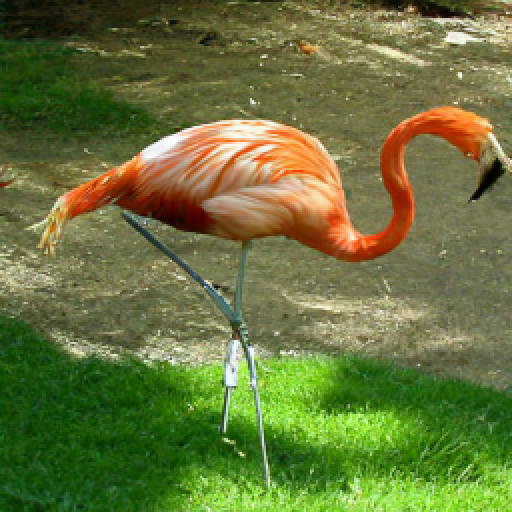} &
\includegraphics[width=0.2\textwidth]{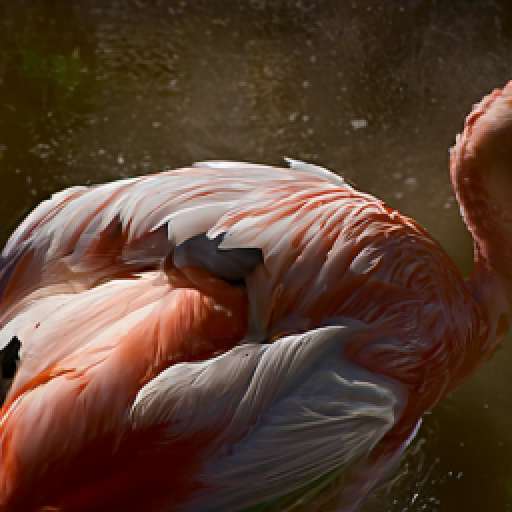} &
\includegraphics[width=0.2\textwidth]{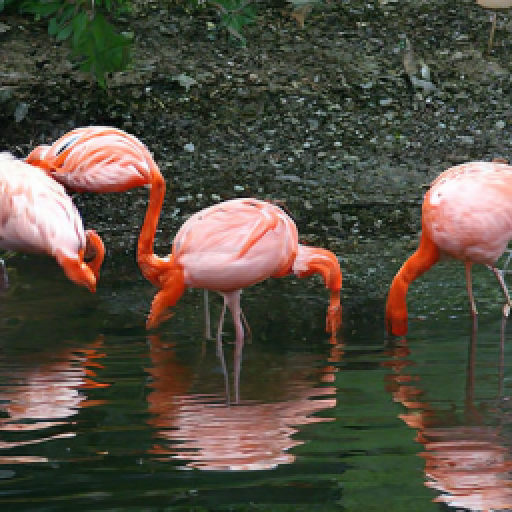} &
\includegraphics[width=0.2\textwidth]{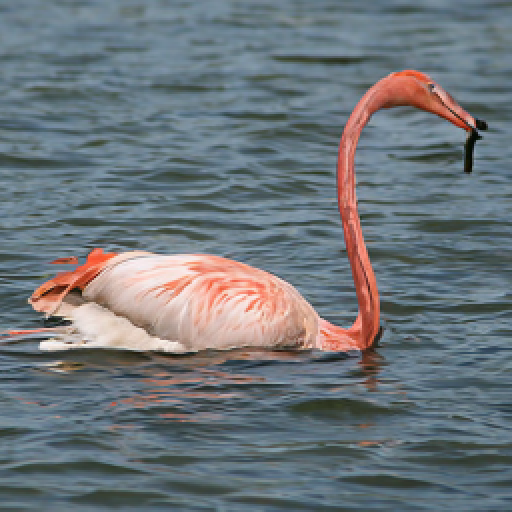} &
\includegraphics[width=0.2\textwidth]{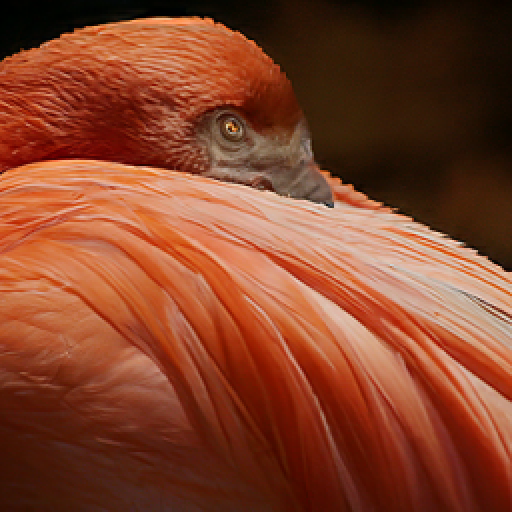} \\

\includegraphics[width=0.2\textwidth]{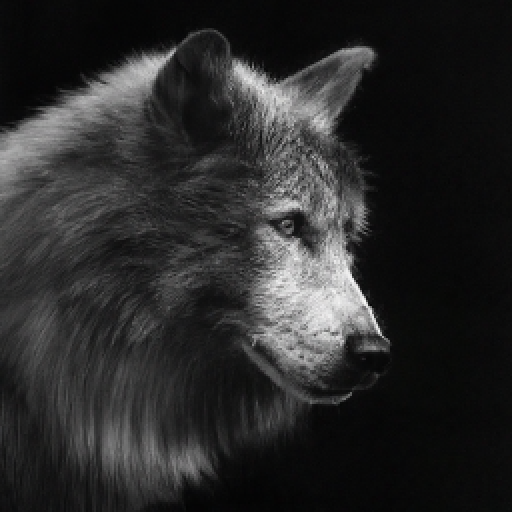} &
\includegraphics[width=0.2\textwidth]{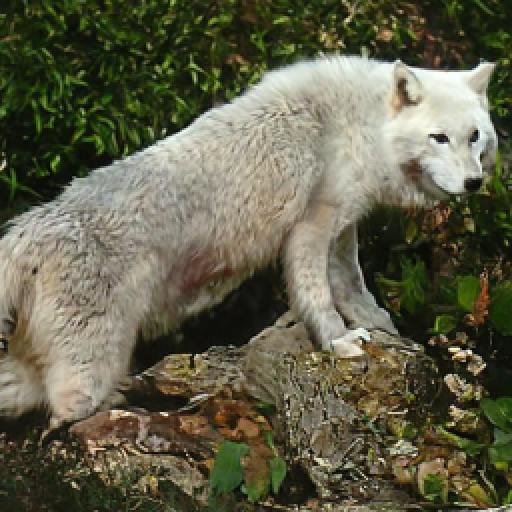} &
\includegraphics[width=0.2\textwidth]{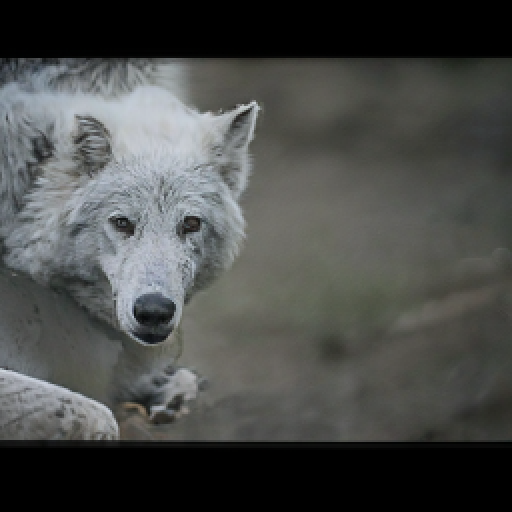} &
\includegraphics[width=0.2\textwidth]{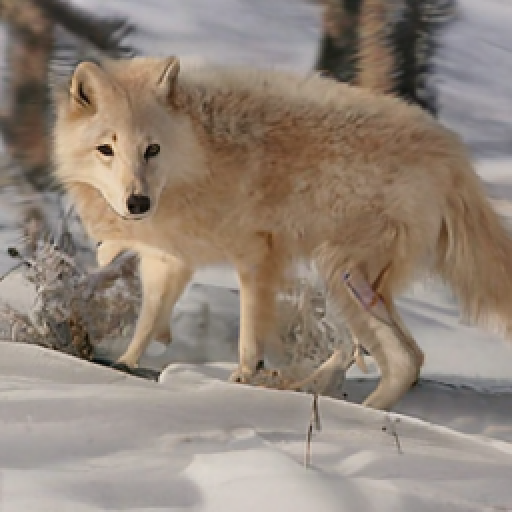} &
\includegraphics[width=0.2\textwidth]{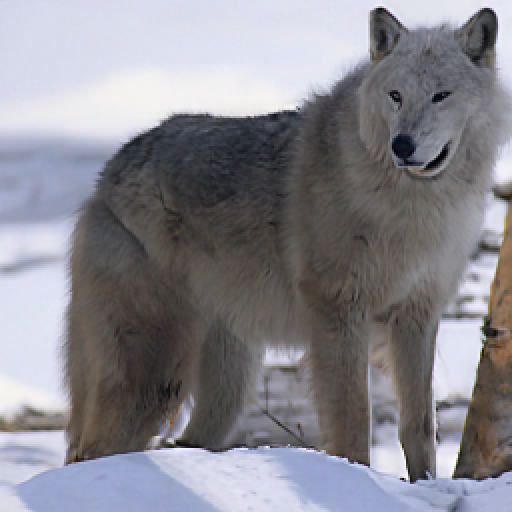} \\

\includegraphics[width=0.2\textwidth]{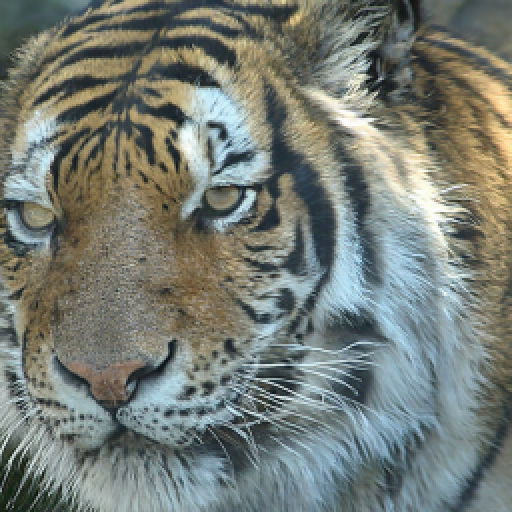} &
\includegraphics[width=0.2\textwidth]{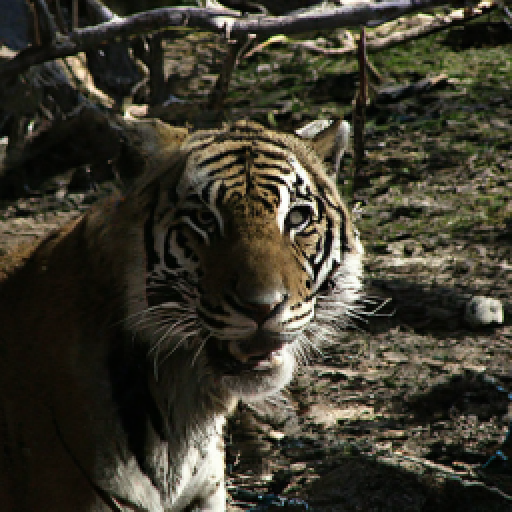} &
\includegraphics[width=0.2\textwidth]{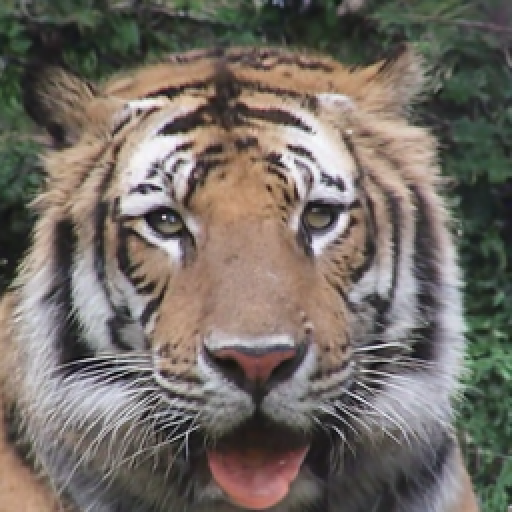} &
\includegraphics[width=0.2\textwidth]{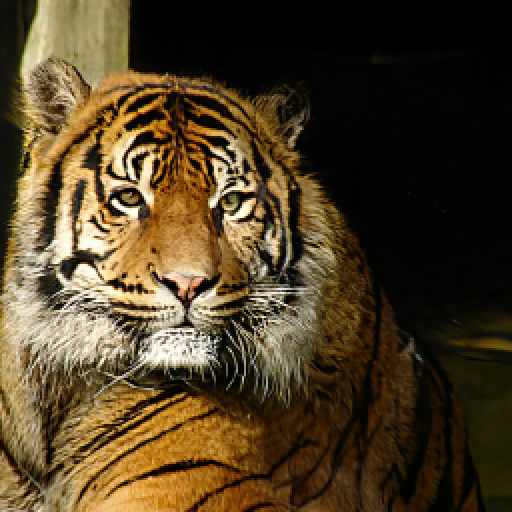} &
\includegraphics[width=0.2\textwidth]{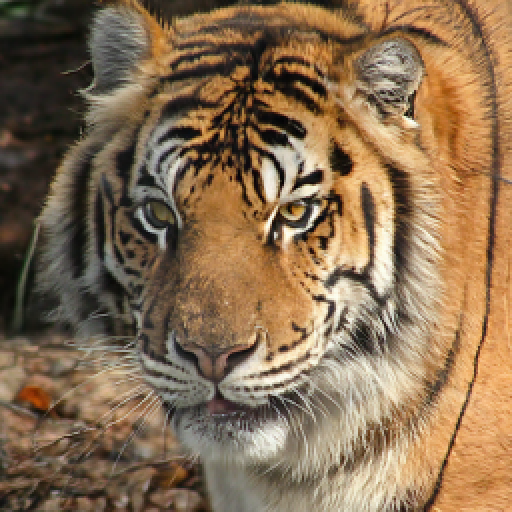} \\

\includegraphics[width=0.2\textwidth]{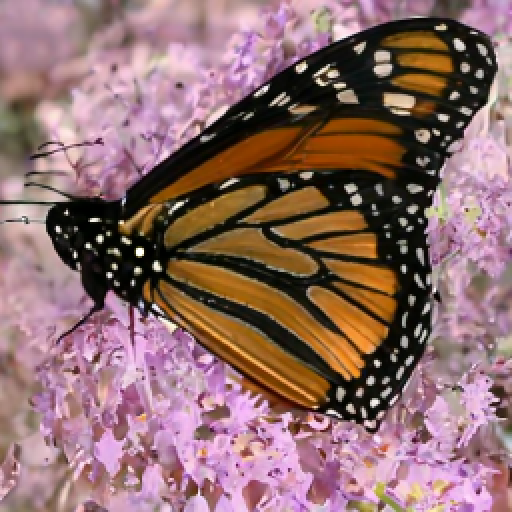} &
\includegraphics[width=0.2\textwidth]{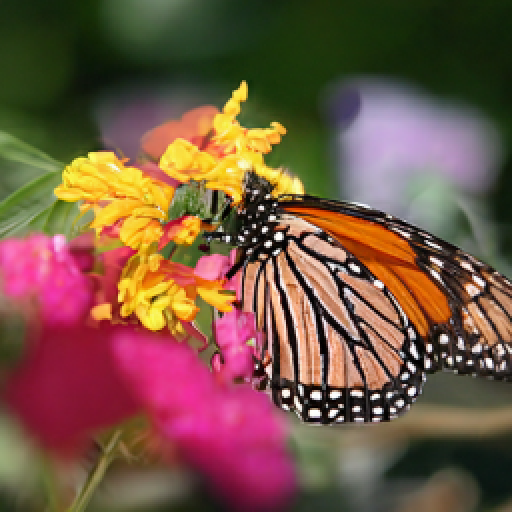} &
\includegraphics[width=0.2\textwidth]{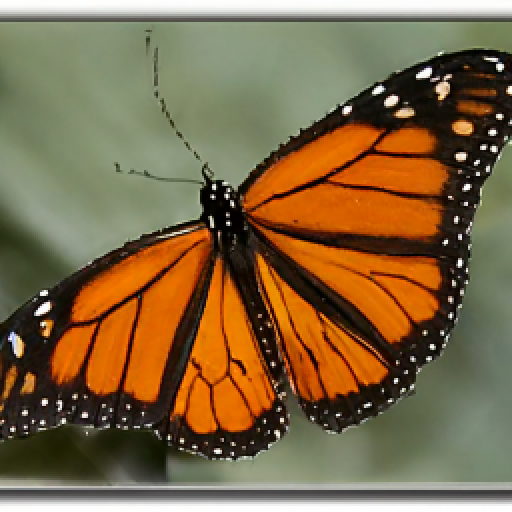} &
\includegraphics[width=0.2\textwidth]{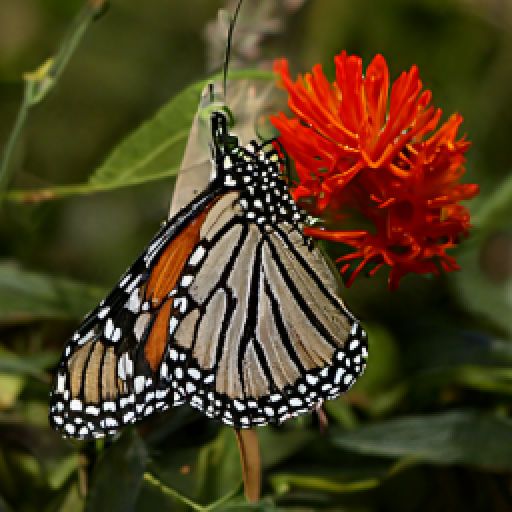} &
\includegraphics[width=0.2\textwidth]{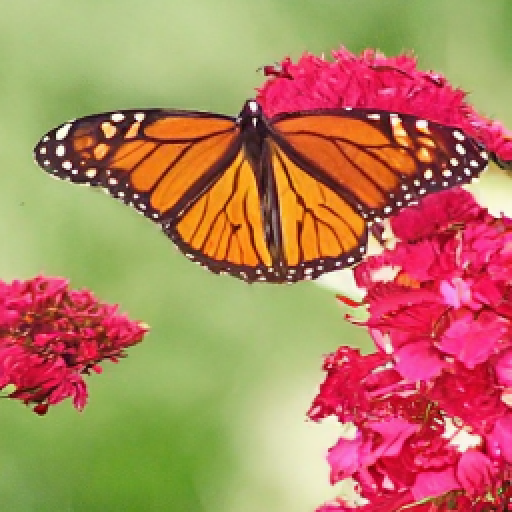} \\

\includegraphics[width=0.2\textwidth]{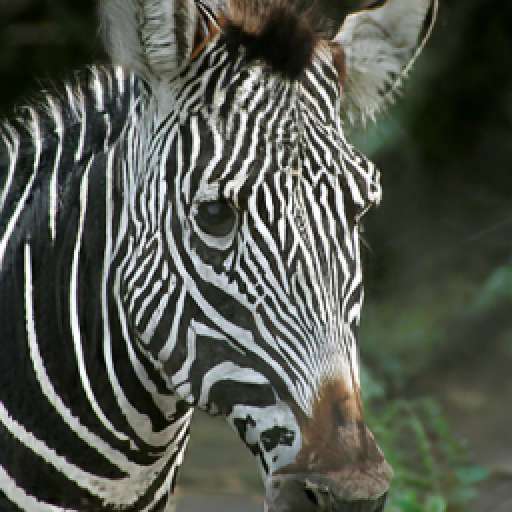} &
\includegraphics[width=0.2\textwidth]{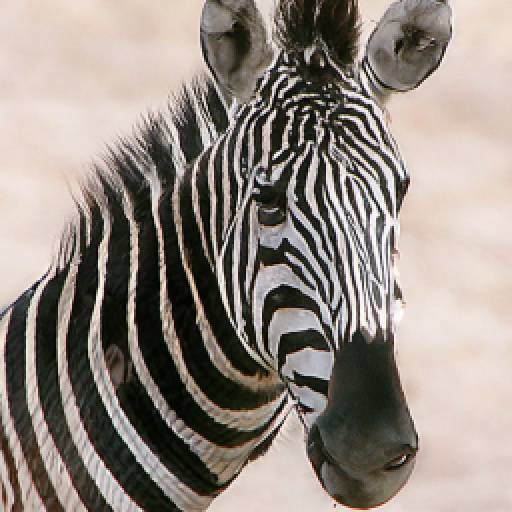} &
\includegraphics[width=0.2\textwidth]{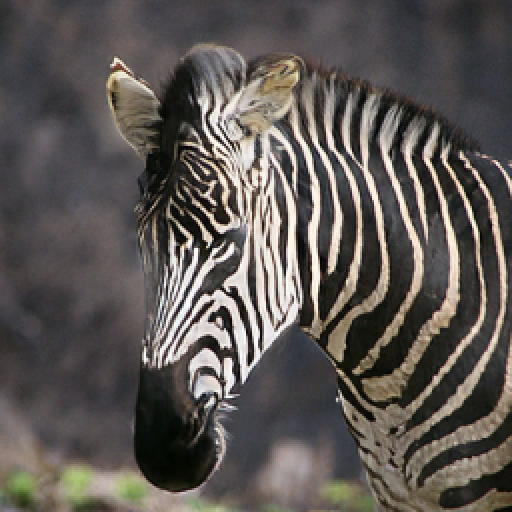} &
\includegraphics[width=0.2\textwidth]{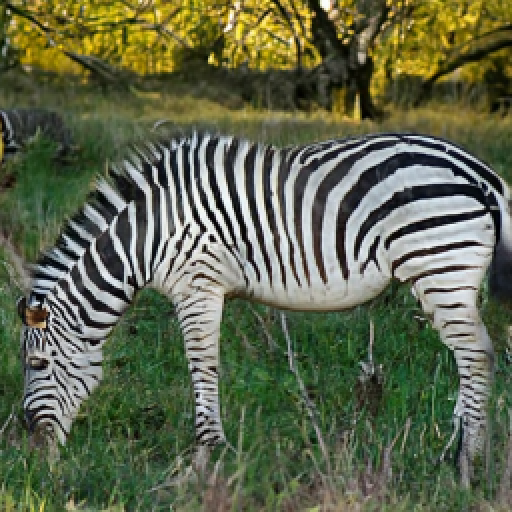} &
\includegraphics[width=0.2\textwidth]{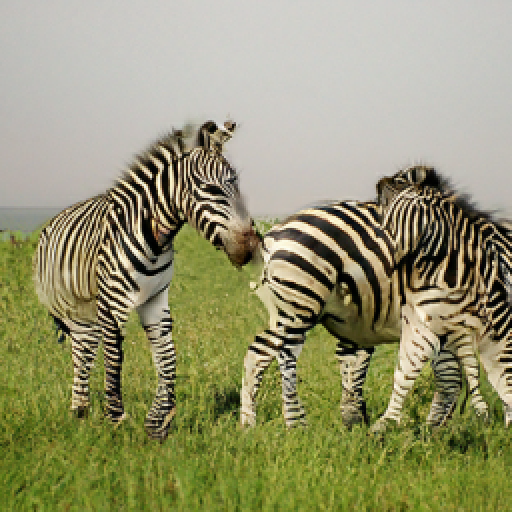} \\

\includegraphics[width=0.2\textwidth]{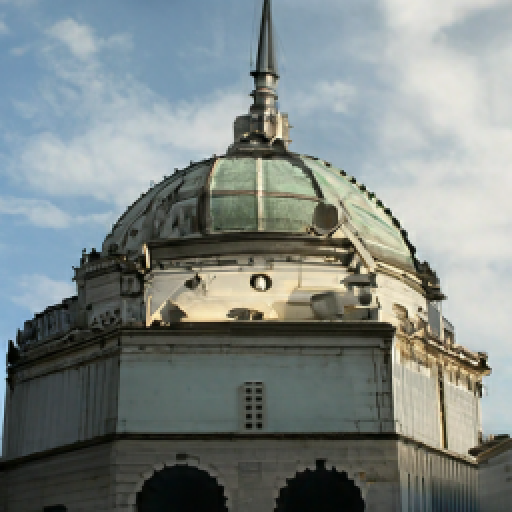} &
\includegraphics[width=0.2\textwidth]{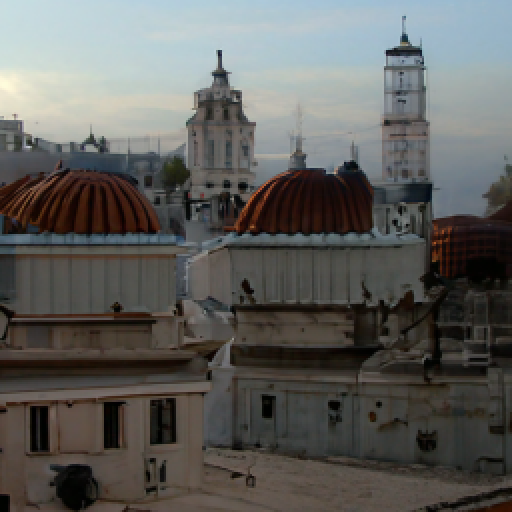} &
\includegraphics[width=0.2\textwidth]{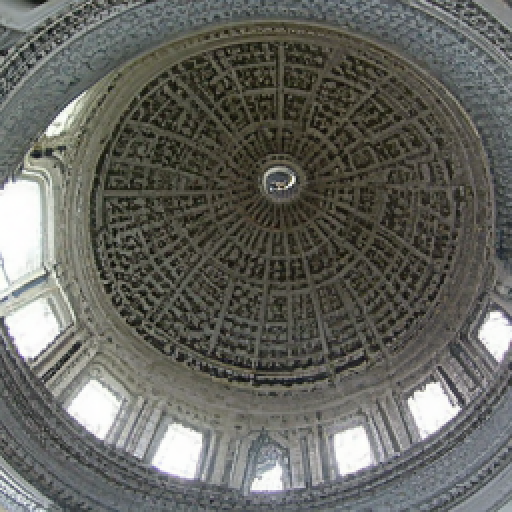} &
\includegraphics[width=0.2\textwidth]{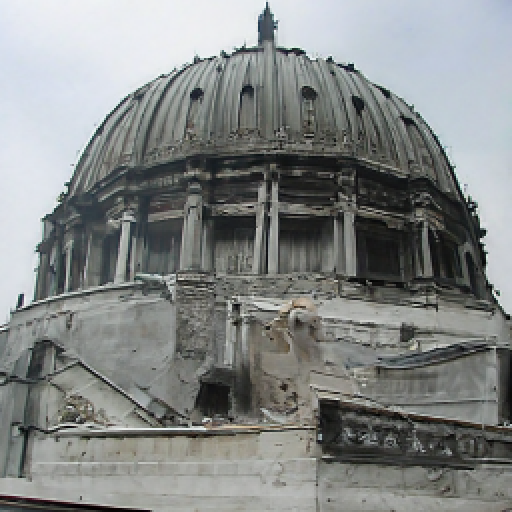} &
\includegraphics[width=0.2\textwidth]{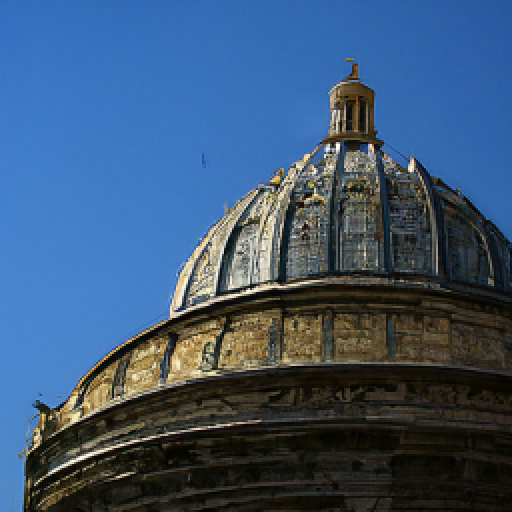}
\end{tabular}
\vspace*{-0.4cm}
\caption{Classwise Synthetic 256$\times$256 ImageNet images. Each row represents a specific ImageNet class. Classes from top to bottom - Flamingo (130), White Wolf (270), Tiger (292), Monarch Butterfly (323), Zebra (340) and Dome (538).}
\vspace*{-0.2cm}
\label{fig:imagenet_classwise_1}
\end{figure}

\begin{figure}[p]
\vspace*{-1cm}
\setlength{\tabcolsep}{1.25pt}
\centering
\begin{tabular}{ccccc}
\includegraphics[width=0.2\textwidth]{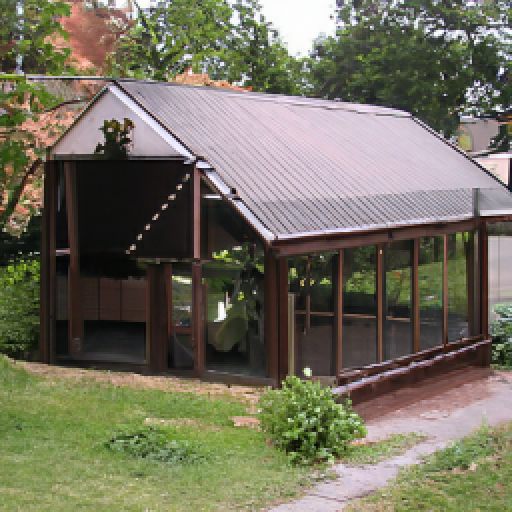} &
\includegraphics[width=0.2\textwidth]{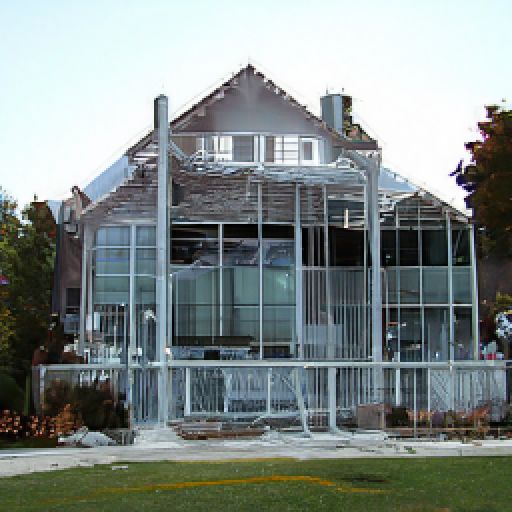} &
\includegraphics[width=0.2\textwidth]{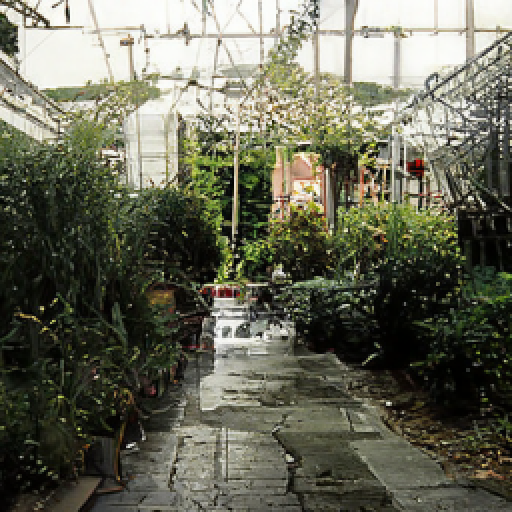} &
\includegraphics[width=0.2\textwidth]{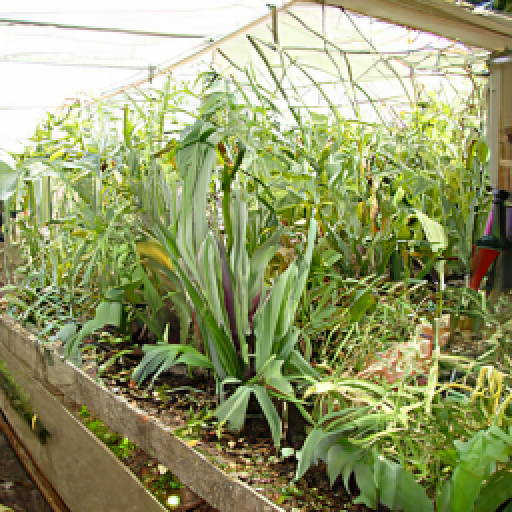} &
\includegraphics[width=0.2\textwidth]{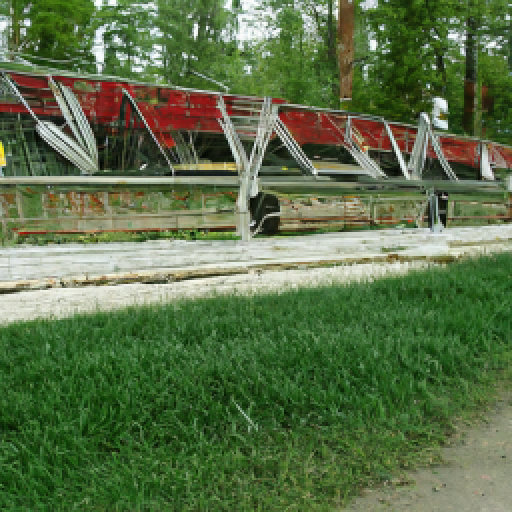} \\

\includegraphics[width=0.2\textwidth]{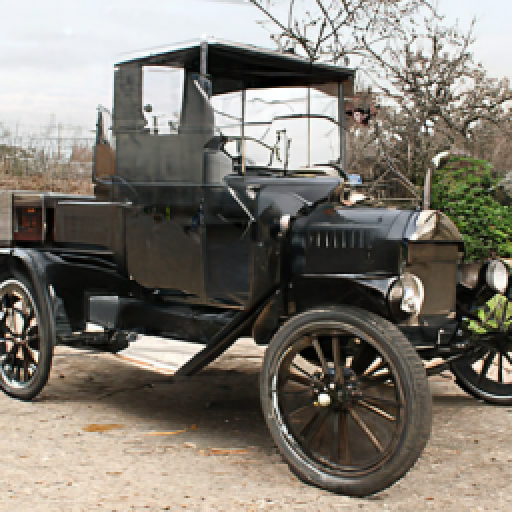} &
\includegraphics[width=0.2\textwidth]{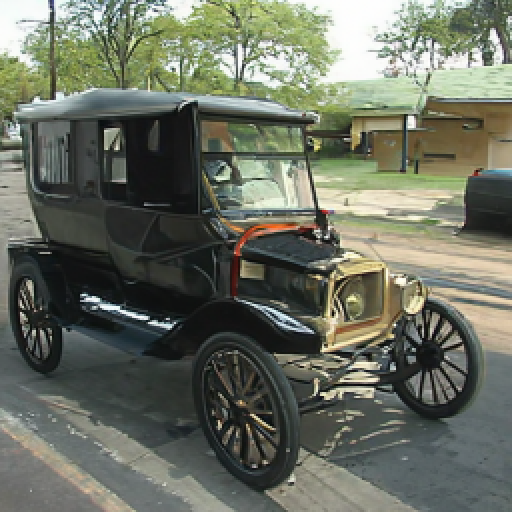} &
\includegraphics[width=0.2\textwidth]{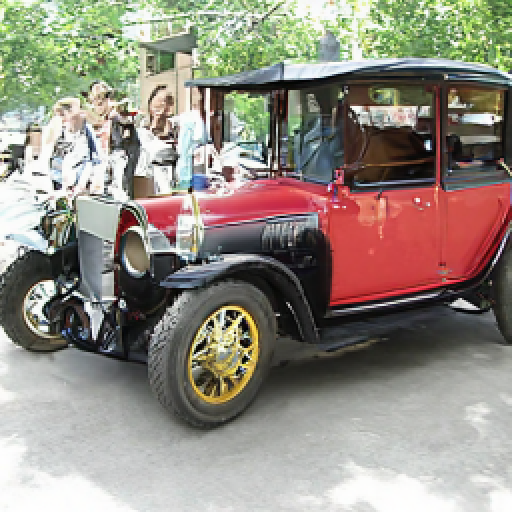} &
\includegraphics[width=0.2\textwidth]{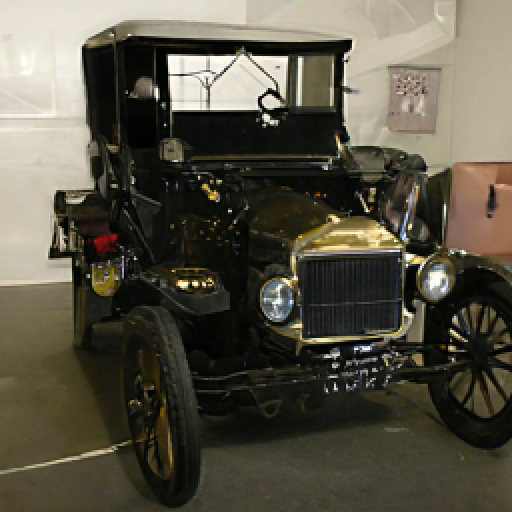} &
\includegraphics[width=0.2\textwidth]{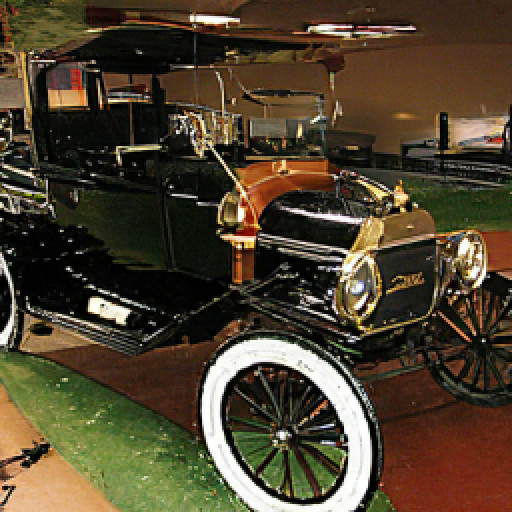} \\

\includegraphics[width=0.2\textwidth]{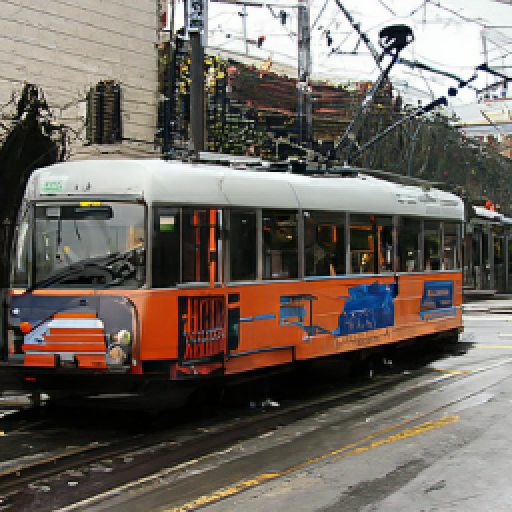} &
\includegraphics[width=0.2\textwidth]{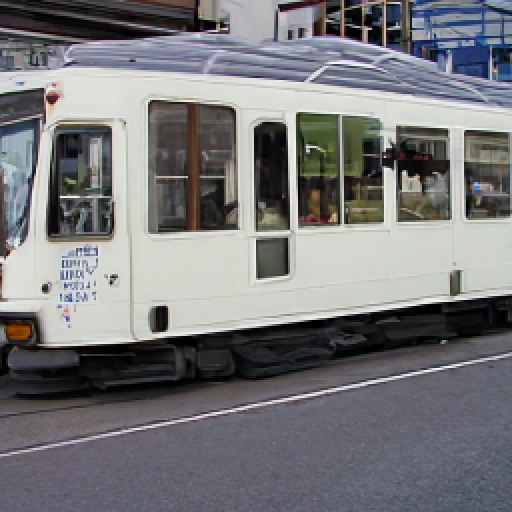} &
\includegraphics[width=0.2\textwidth]{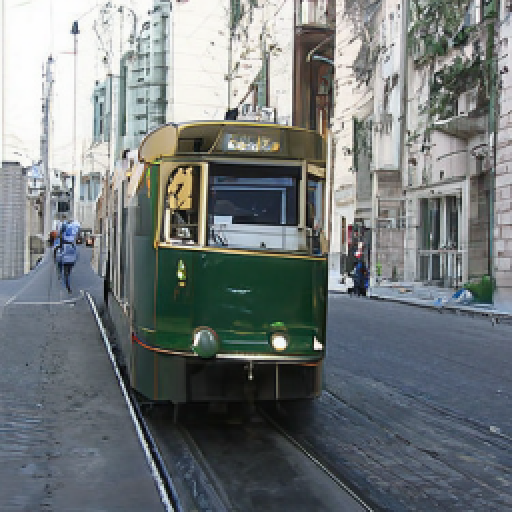} &
\includegraphics[width=0.2\textwidth]{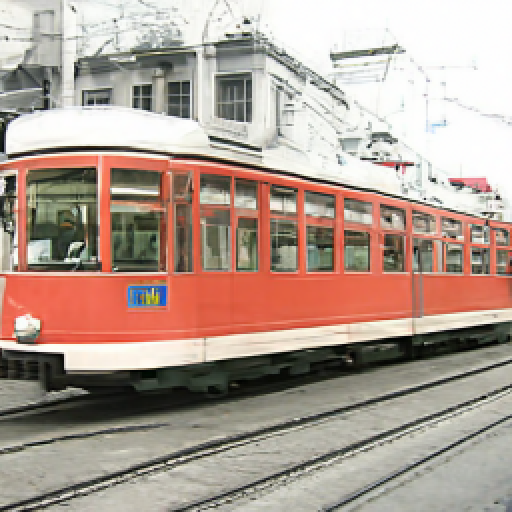} &
\includegraphics[width=0.2\textwidth]{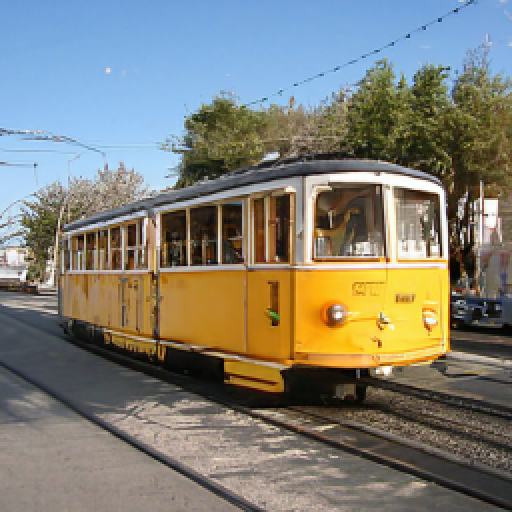} \\

\includegraphics[width=0.2\textwidth]{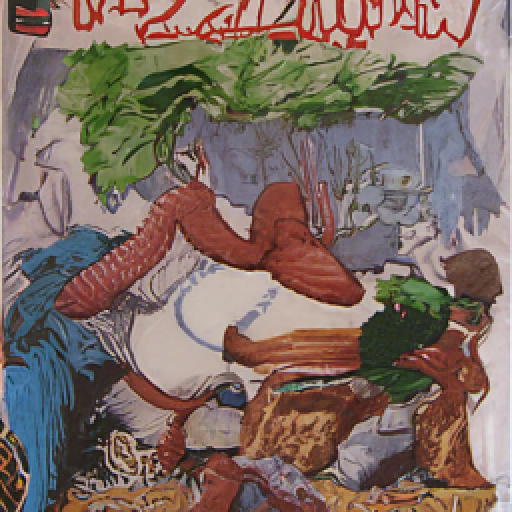} &
\includegraphics[width=0.2\textwidth]{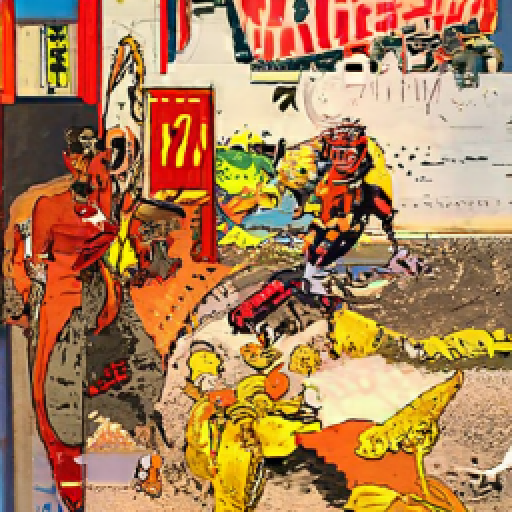} &
\includegraphics[width=0.2\textwidth]{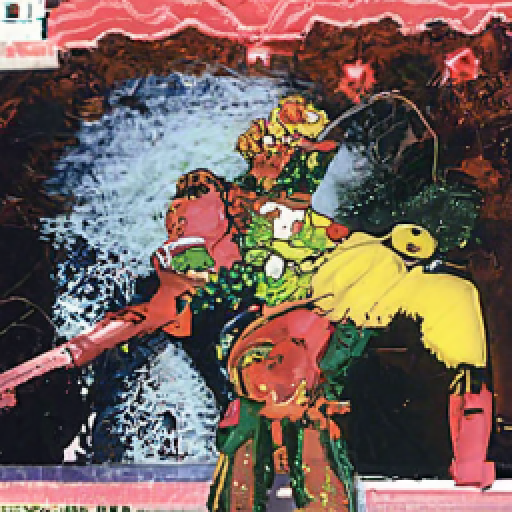} &
\includegraphics[width=0.2\textwidth]{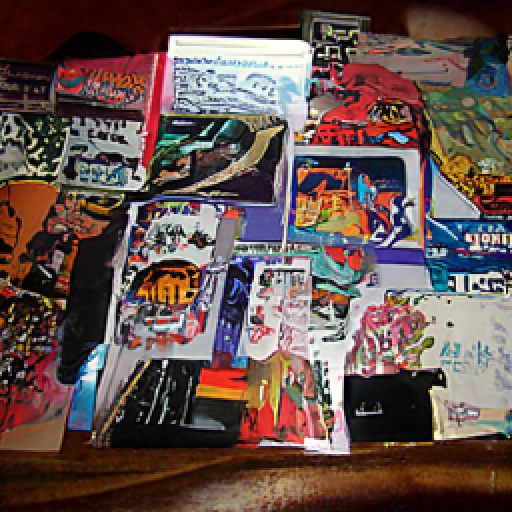} &
\includegraphics[width=0.2\textwidth]{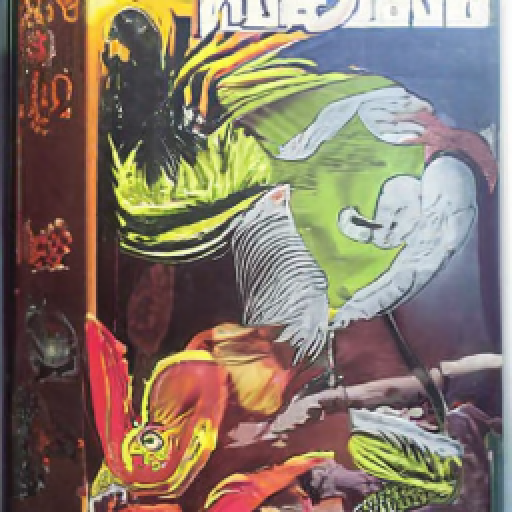} \\

\includegraphics[width=0.2\textwidth]{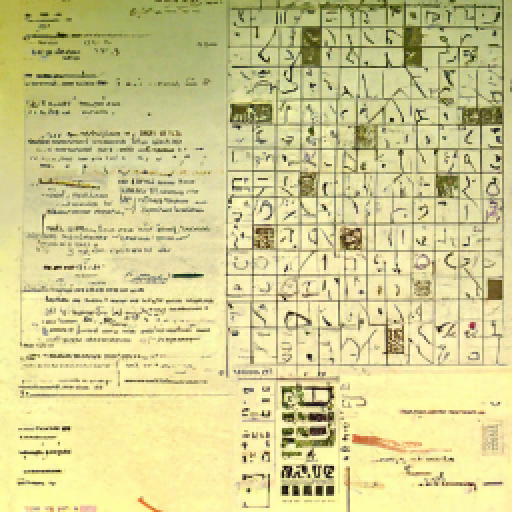} &
\includegraphics[width=0.2\textwidth]{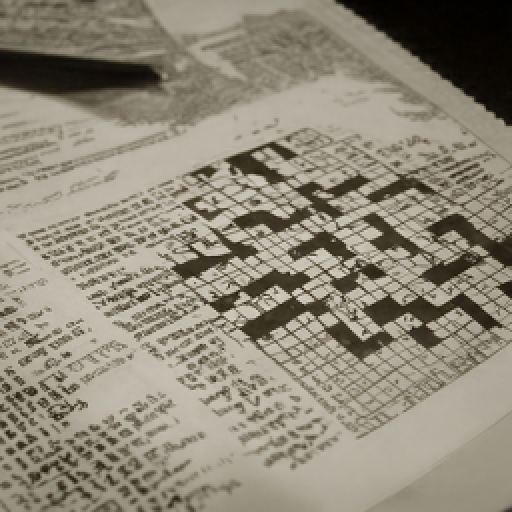} &
\includegraphics[width=0.2\textwidth]{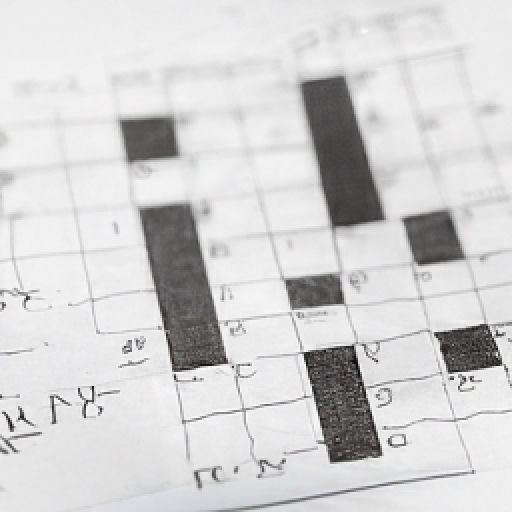} &
\includegraphics[width=0.2\textwidth]{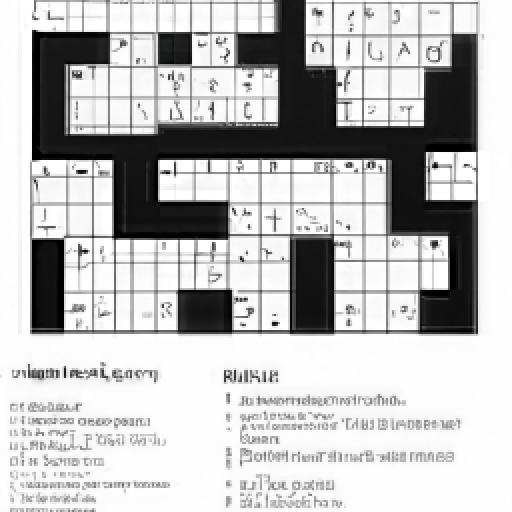} &
\includegraphics[width=0.2\textwidth]{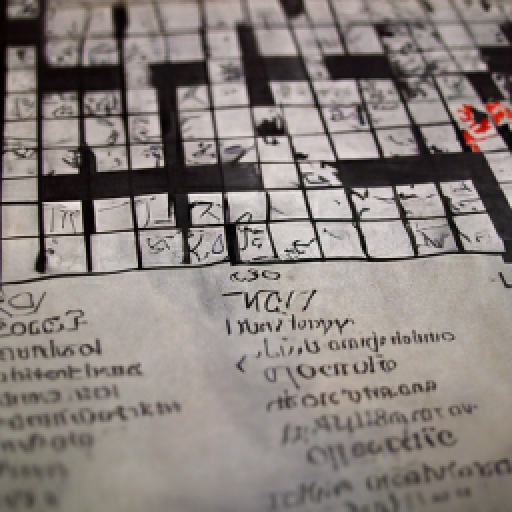} \\

\includegraphics[width=0.2\textwidth]{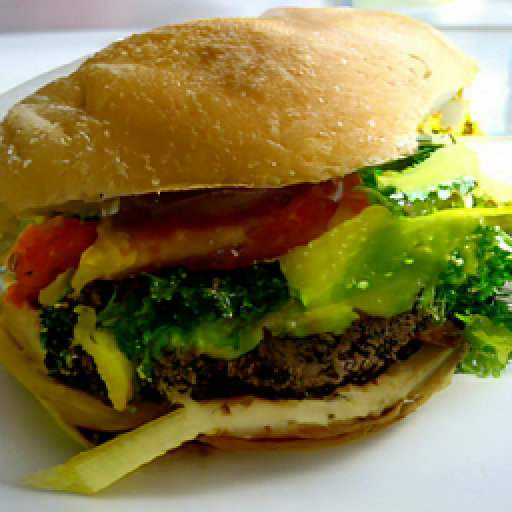} &
\includegraphics[width=0.2\textwidth]{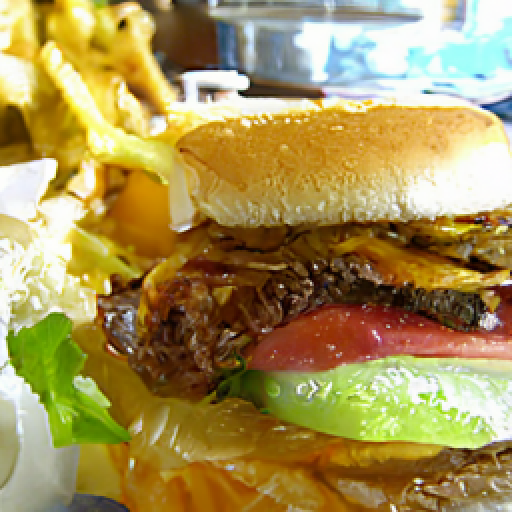} &
\includegraphics[width=0.2\textwidth]{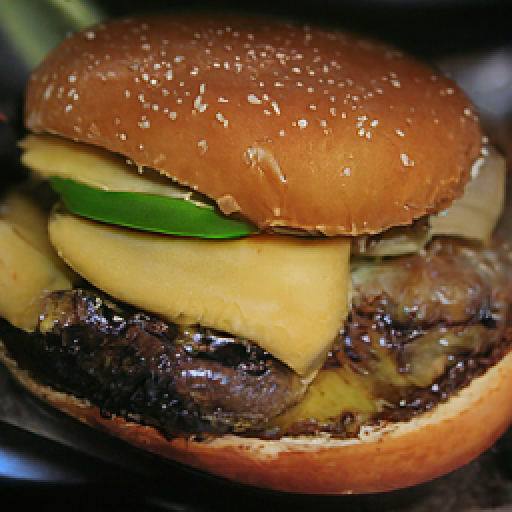} &
\includegraphics[width=0.2\textwidth]{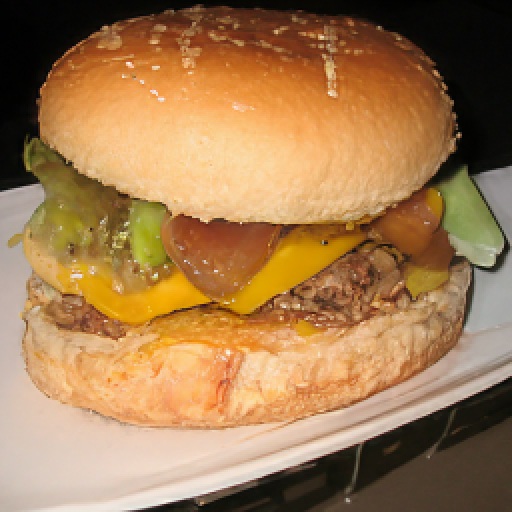} &
\includegraphics[width=0.2\textwidth]{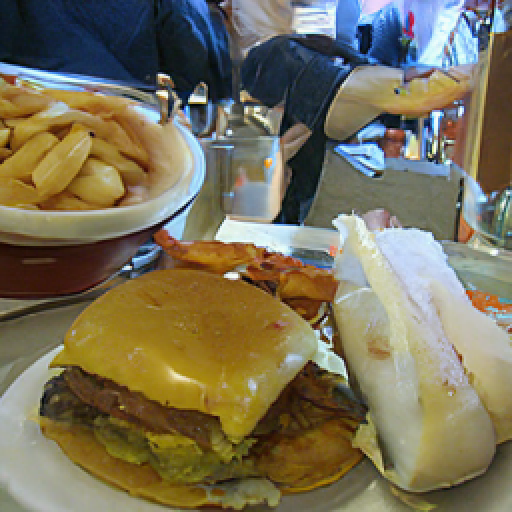}
\end{tabular}
\vspace*{-0.4cm}
\caption{Classwise Synthetic 256$\times$256 ImageNet images. Each row represents a specific ImageNet class. Classes from top to bottom - Greenhouse (580), Model T (661), Streetcar (829), Comic Book (917), Crossword Puzzle (918), Cheeseburger (933).}
\vspace*{-0.2cm}
\label{fig:imagenet_classwise_2}
\end{figure}

We designed experiments to improve the sample quality metrics of cascaded diffusion models on class-conditional ImageNet generation. Our cascading pipelines consist of class-conditional diffusion models at all resolutions, so class information is injected at all resolutions: see~\cref{fig:cascade_unet_fig}. Our final ImageNet results are described in~\cref{sec:main_results}.

To give insight into our cascading pipelines, we begin with improvements on a baseline non-cascaded model at the 64$\times$64 resolution~(\cref{sec:baseline_improvements}), then we show that cascading up to 64$\times$64 improves upon our best non-cascaded 64$\times$64 model, but only in conjunction with conditioning augmentation.
We also show that truncated and non-truncated conditioning augmentation perform equally well~(\cref{sec:cond_aug_ablations}), and we study random Gaussian blur augmentation to train super-resolution models to
resolutions of 128$\times$128 and 256$\times$256~(\cref{sec:experiments_128_256}). Finally, we verify that conditioning augmentation is also effective on the LSUN dataset~\citep{yu15lsun} and therefore is not specific to ImageNet~(\cref{sec:experiments_lsun}).

We cropped and resized the ImageNet dataset~\citep{russakovsky2015imagenet} in the same manner as BigGAN~\citep{brock2018large}.
We report Inception scores using the standard practice of generating 50k samples and calculating the mean and standard deviation over 10 splits~\citep{salimans2016improved}. Generally, throughout our experiments, we selected models and performed early stopping based on FID score calculated over 10k samples, but all reported FID scores are calculated over 50k samples for comparison with other work~\citep{heusel2017gans}. The FID scores we used for model selection and reporting model performance are calculated against training set statistics according to common practice, but since this can be seen as overfitting on the performance metric, we additionally report model performance using FID scores calculated against validation set statistics. We also report results on Classification Accuracy Score (CAS), which was proposed by \citet{ravuris2019cas} due to their findings that non-GAN models may score poorly on FID and IS despite generating visually appealing samples and that FID and IS are not correlated (sometimes anti-correlated) with performance on downstream tasks.

\subsection{Main Cascading Pipeline Results} \label{sec:main_results}

\begin{figure}[htbp] \centering
\setlength{\tabcolsep}{1.25pt}
\begin{tabular}{ccc}
\adjincludegraphics[width=0.325\textwidth]{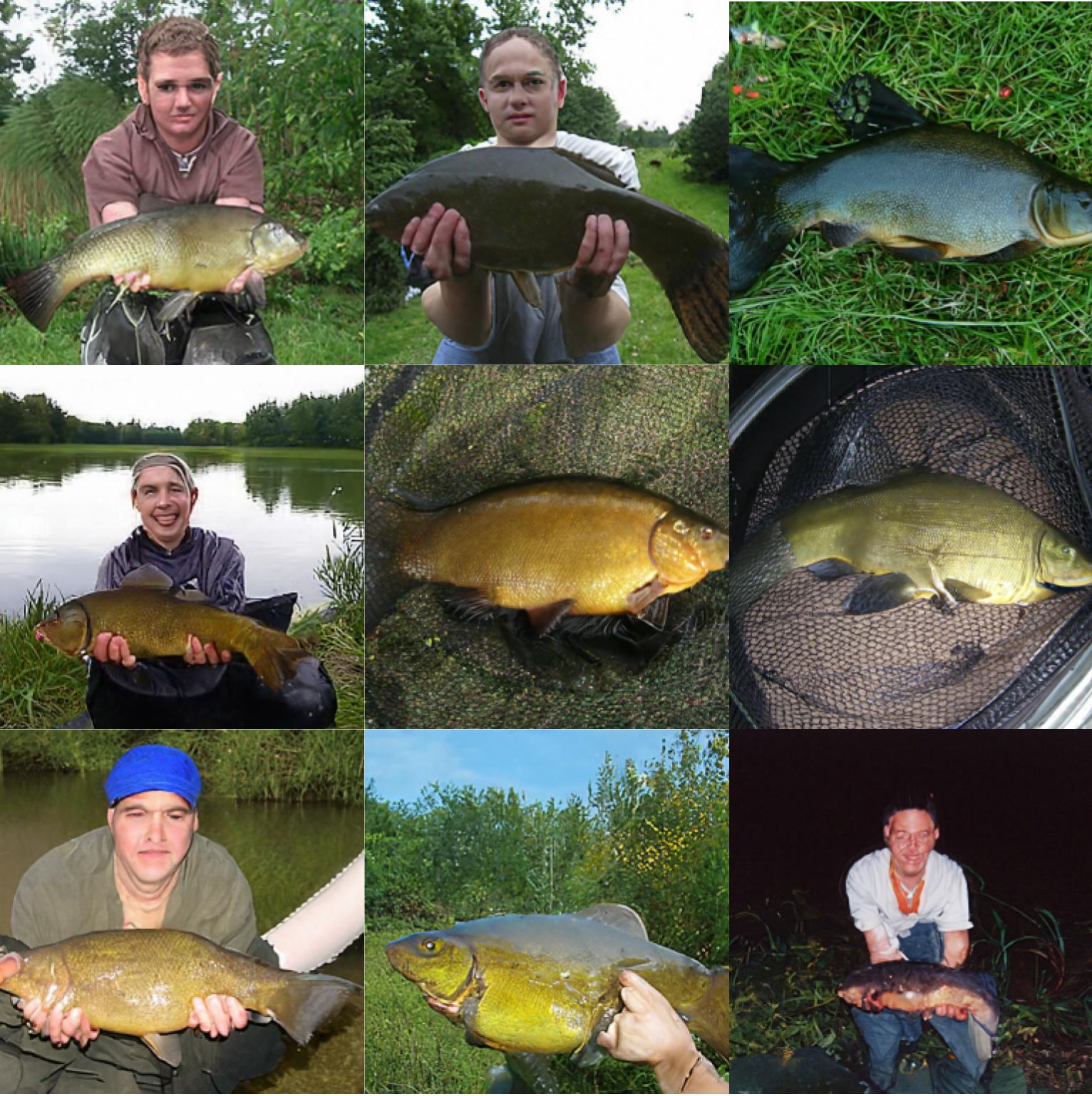} &
\adjincludegraphics[width=0.325\textwidth,trim={0 0 {.25\width} {.25\width}},clip]{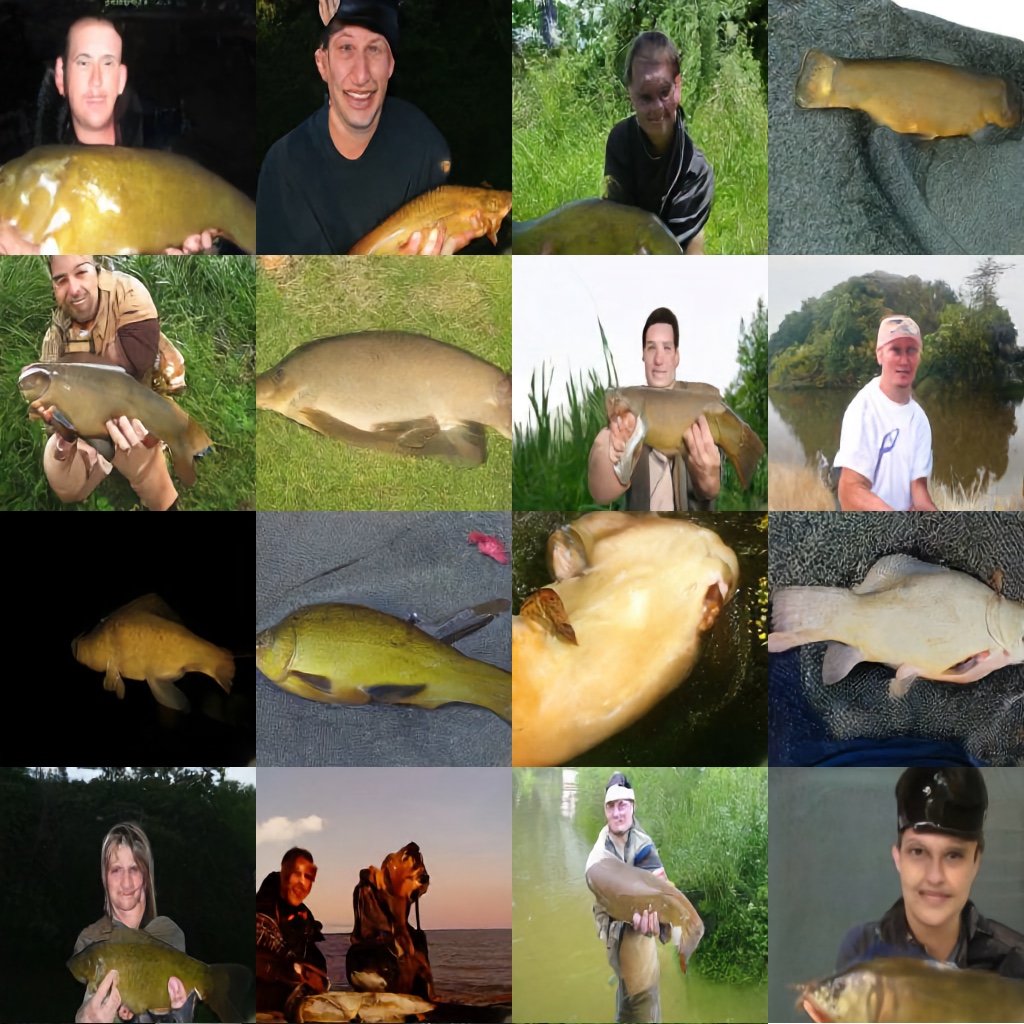} &
\adjincludegraphics[width=0.325\textwidth,trim={0 0 {.25\width} {.25\width}},clip]{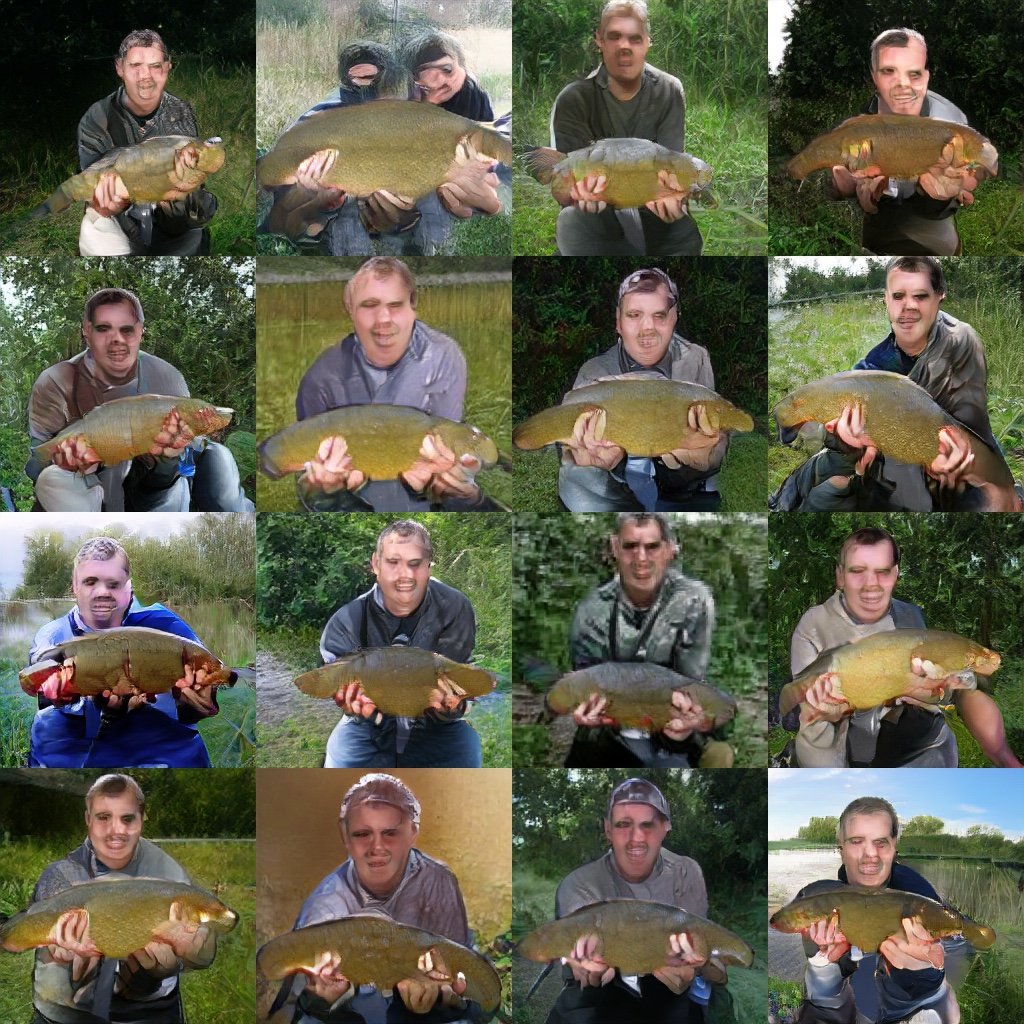} \\

\adjincludegraphics[width=0.325\textwidth]{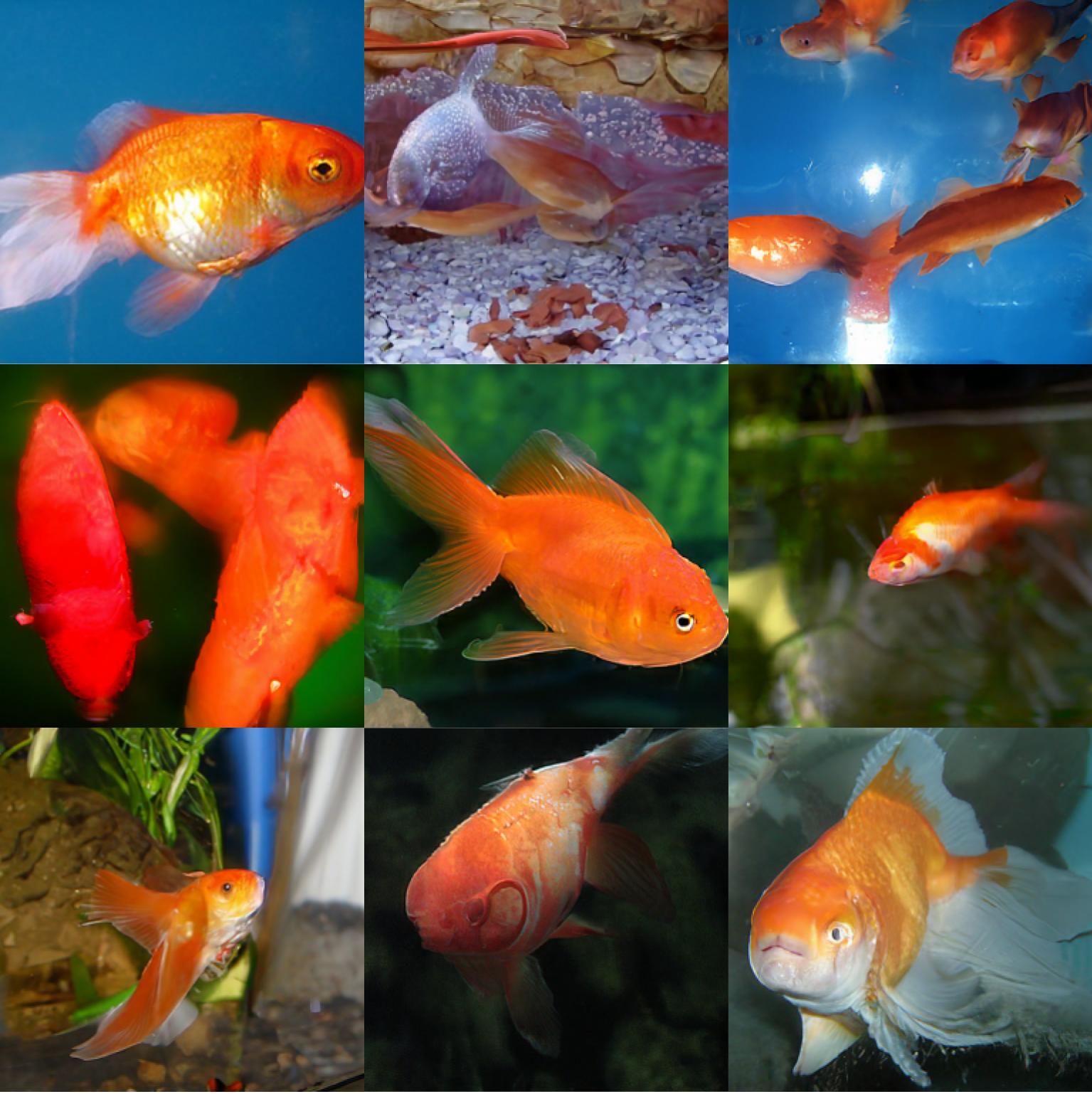} &
\adjincludegraphics[width=0.325\textwidth,trim={0 0 {.25\width} {.25\width}},clip]{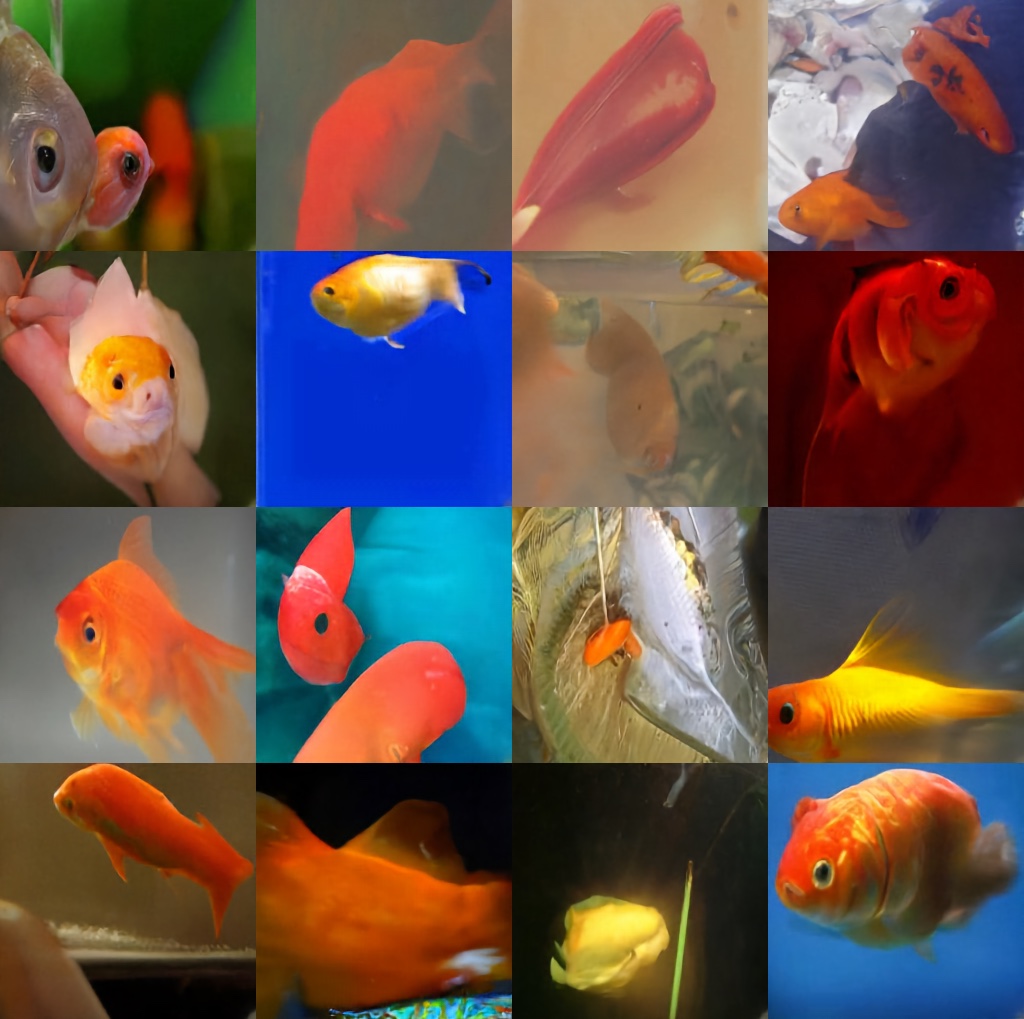} &
\adjincludegraphics[width=0.325\textwidth,trim={0 0 {.25\width} {.25\width}},clip]{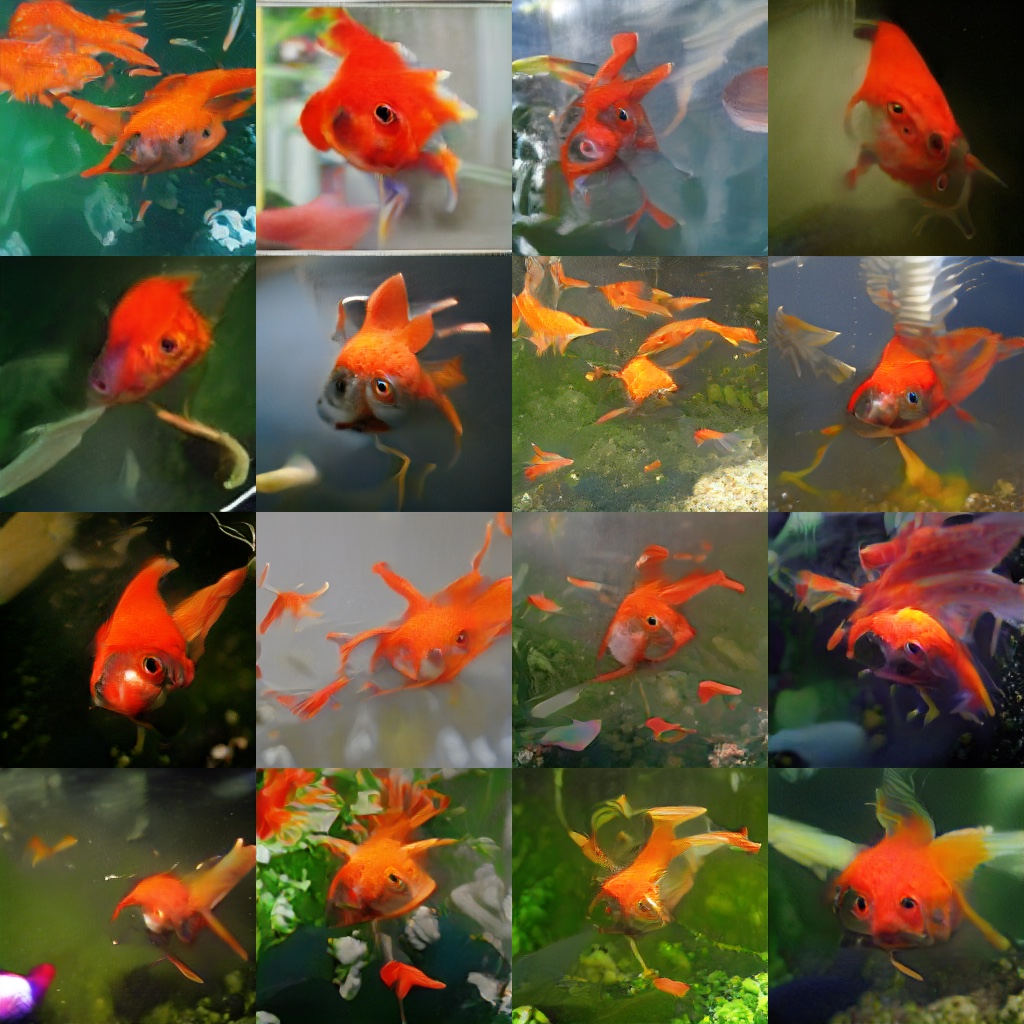} \\

\adjincludegraphics[width=0.325\textwidth]{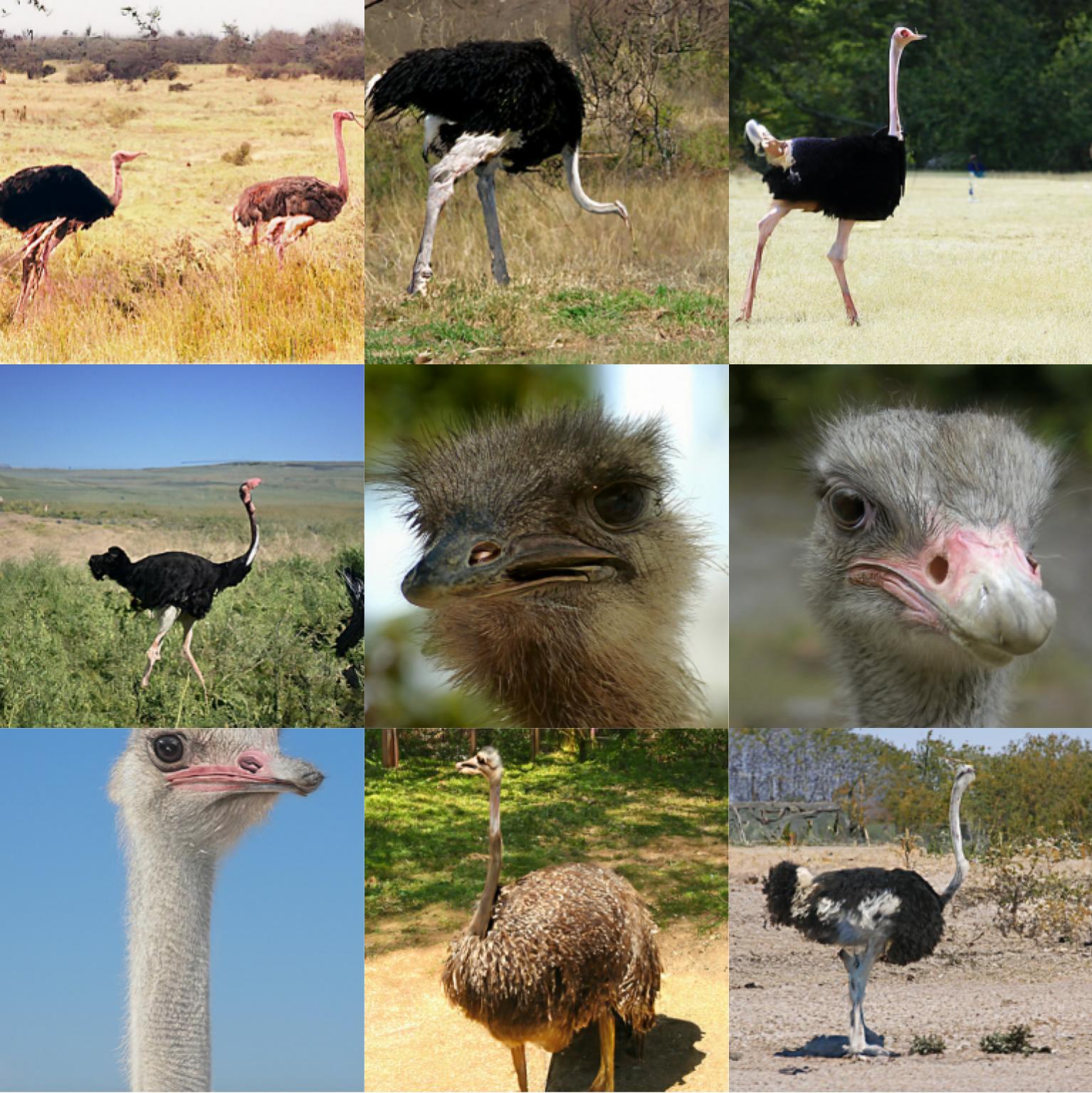} &
\adjincludegraphics[width=0.325\textwidth,trim={0 0 {.25\width} {.25\width}},clip]{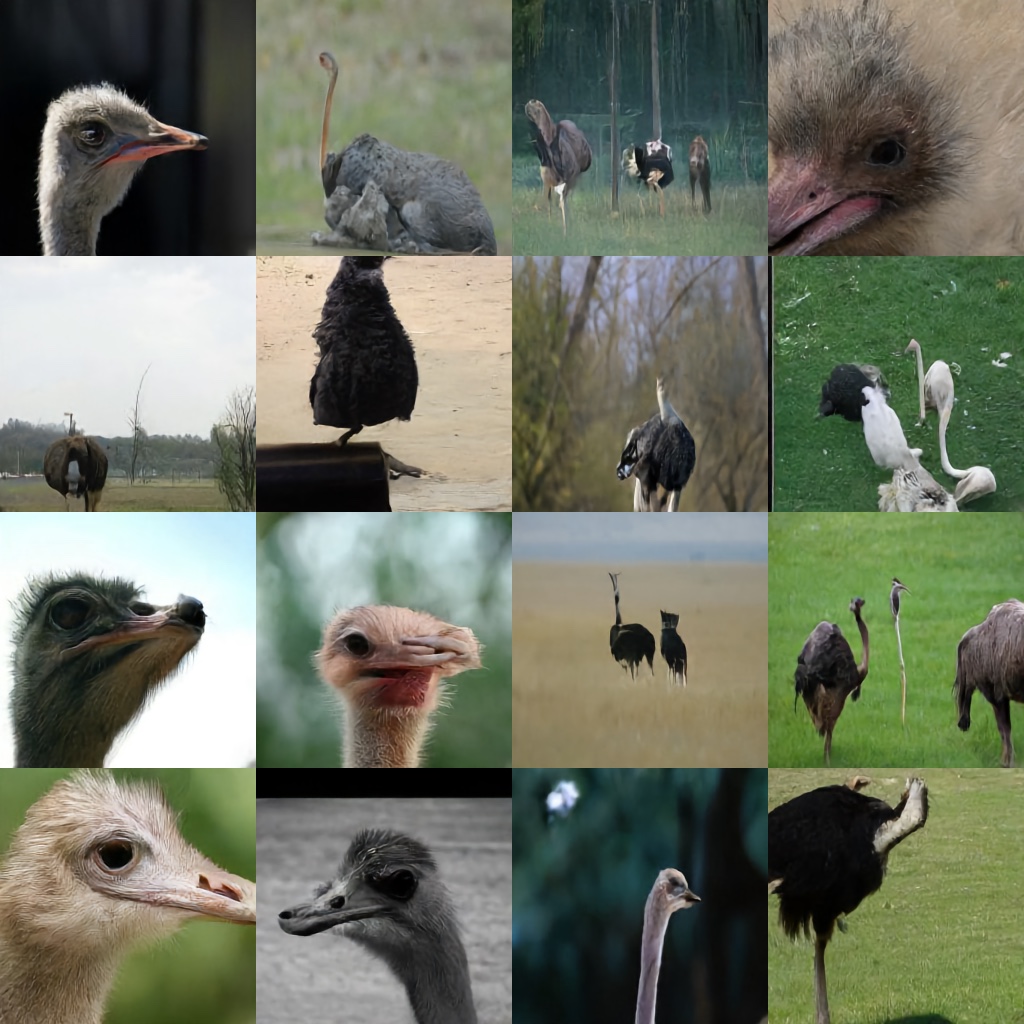} &
\adjincludegraphics[width=0.325\textwidth,trim={0 0 {.25\width} {.25\width}},clip]{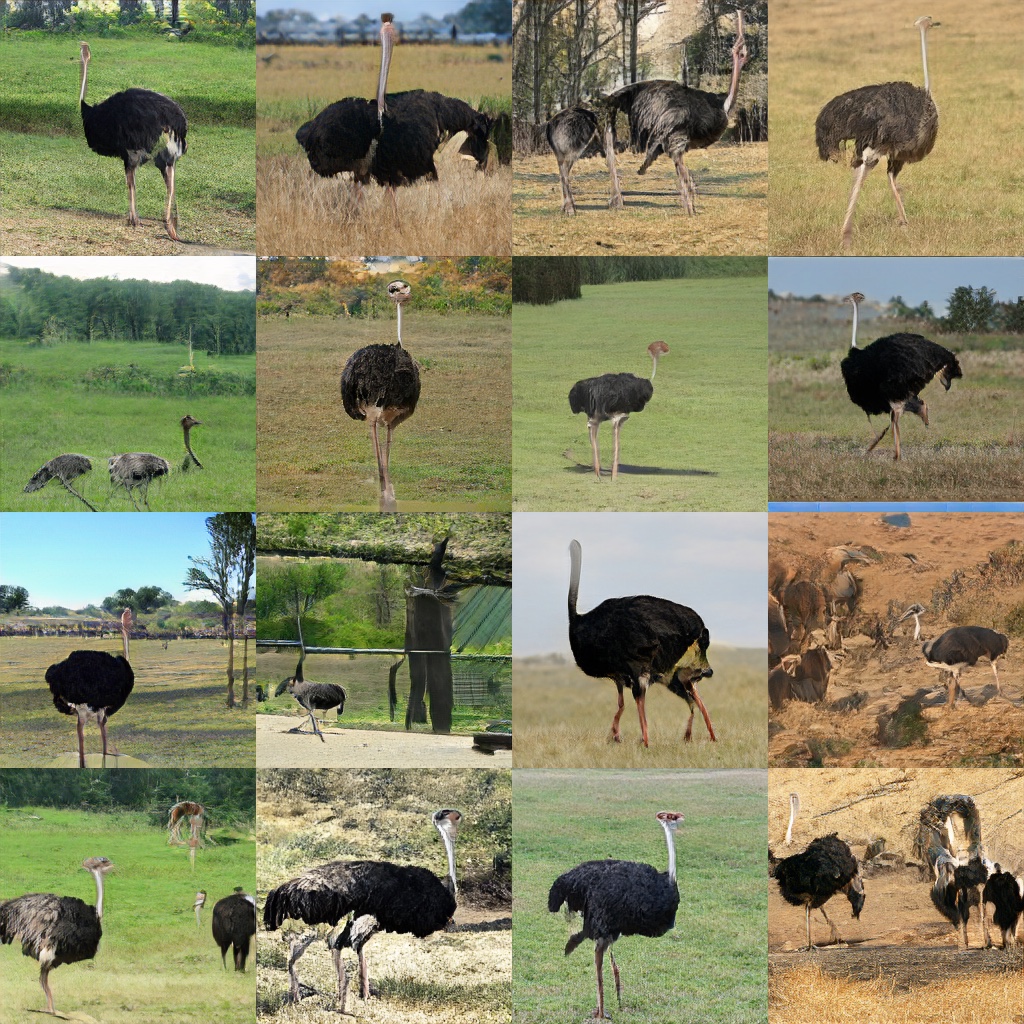} \\

CDM (ours) & VQ-VAE-2 & BigGAN-deep \\
\end{tabular}
\vspace*{-0.2cm}
\caption{Comparing the quality and diversity of model samples in selected 256$\times$256 ImageNet classes \{Tench(0), Goldfish(1) and Ostrich(9)\}. VQVAE-2 and BigGAN samples are taken from \citet{razavi2019generating}.}
\label{fig:imagenet_256x_montage_comparison}
\end{figure}

\cref{table:main_results} reports the main results on the cascaded diffusion model (\emph{CDM}), for the 64$\times$64, 128$\times$128, and 256$\times$256 ImageNet dataset resolutions, along with baselines. 
CDM outperforms BigGAN-deep in terms of FID score on the image resolutions considered, but GANs perform better in terms of Inception score when their truncation parameter is optimized for Inception score~\citep{brock2018large}. We also outperform concurrently released diffusion models that do not use classifier guidance to boost sample quality scores~\citep{dhariwal2021diffusion}. See~\cref{fig:imagenet_256x_montage_comparison} for a qualitative assessment of sample quality and diversity compared to VQ-VAE-2~\citep{razavi2019generating} and BigGAN-deep~\citep{brock2018large}, and see \cref{fig:imagenet_classwise_1,fig:imagenet_classwise_2} for examples of generated images.

\begin{figure}[tbh]\centering
    \resizebox{0.6\linewidth}{!}{{\input{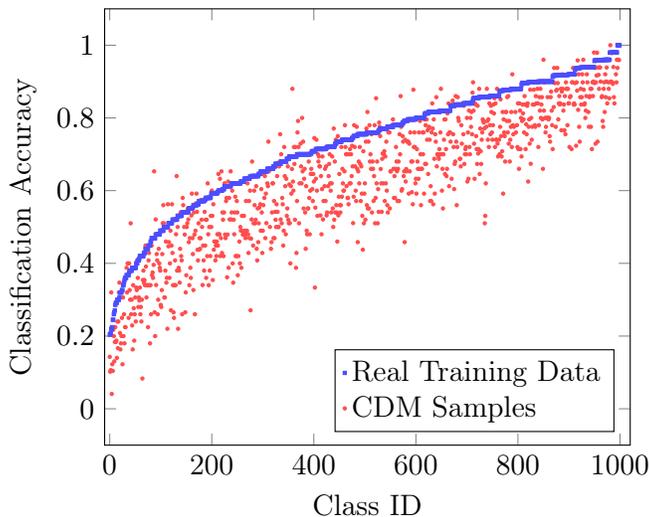}}}
    \caption{Classwise Classification Accuracy Score comparison between real data (blue) and generated data (red) at the 256$\times$256 resolution. Accompanies~\cref{table:cas_results}.}
    \label{fig:cas_acc_dist}
\end{figure}

\cref{table:cas_results} reports the results on Classification Accuracy Score (CAS) \citep{ravuris2019cas} for our models at the 128$\times$128 and 256$\times$256 resolutions.
We find that CDM outperforms VQ-VAE-2 and BigGAN-deep at both resolutions by a significant margin on the CAS metric, suggesting better potential performance on downstream tasks. Figure \ref{fig:cas_acc_dist} compares class-wise classification accuracy scores between classifiers trained on real training data, and CDM samples. The CDM classifier outperforms real data on 96 classes compared to 6 and 31 classes by BigGAN-deep and VQ-VAE-2 respectively. We also show samples from classes with best and worst accuracy scores in Appendix Figure \ref{fig:cas_classwise_1} and \ref{fig:cas_classwise_2}.

Our cascading pipelines are structured as a 32$\times$32 base model, a 32$\times$32$\rightarrow$64$\times$64 super-resolution model, followed by  64$\times$64$\rightarrow$128$\times$128 or 64$\times$64$\rightarrow$256$\times$256 super-resolution models. Models at 32$\times$32 and 64$\times$64 resolutions use 4000 diffusion timesteps and architectures similar to DDPM~\citep{ho2020denoising} and Improved DDPM~\citep{nichol2021improved}.
Models at 128$\times$128 and 256$\times$256 resolutions use 100 sampling steps, determined by post-training hyperparameter search (\cref{sec:experiments_128_256}), and they use architectures similar to SR3~\citep{saharia2021image}. All base resolution and super-resolution models are conditioned on class labels. See~\cref{appendix:hyperparams} for details.

\begin{table}[p]\small
\begin{subtable}[c]{\textwidth}\centering
\begin{tabular}{lccc} \toprule
Model & \makecell{FID \\ vs train} & \makecell{FID \\ vs validation} & IS \\ \midrule
32$\times$32 resolution \\ \cmidrule(r){1-1}
    CDM (ours) & 1.11	& 1.99	& 26.01 $\pm$ 0.59 \\ \midrule
64$\times$64 resolution \\ \cmidrule(r){1-1}
    BigGAN-deep, by~\citep{dhariwal2021diffusion} & 4.06 &  &  \\
    Improved DDPM~\citep{nichol2021improved} & 2.92 &  &  \\
    ADM~\citep{dhariwal2021diffusion} & 2.07 &  &  \\
    CDM (ours) & \textbf{1.48} & 2.48 & 67.95 $\pm$ 1.97 \\ \midrule
128$\times$128 resolution \\ \cmidrule(r){1-1}
    BigGAN-deep~\citep{brock2018large} & 5.7 &  & 124.5 \\
    BigGAN-deep, max IS~\citep{brock2018large} & 25 &  & \textbf{253} \\
    LOGAN~\citep{wu2019logan} & \textbf{3.36} &  & 148.2 \\
    ADM~\citep{dhariwal2021diffusion} & 5.91 &  &  \\
    CDM (ours) & 3.52 & 3.76  & 128.80 $\pm$ 2.51 \\ \midrule
256$\times$256 resolution \\ \cmidrule(r){1-1}
    BigGAN-deep~\citep{brock2018large} & 6.9 &  & 171.4 \\
    BigGAN-deep, max IS~\citep{brock2018large} & 27 &  & \textbf{317} \\
    VQ-VAE-2~\citep{razavi2019generating} & 31.11 &  &  \\
    Improved DDPM~\citep{nichol2021improved} & 12.26 &  &  \\
    SR3~\citep{saharia2021image} & 11.30 &  &  \\
    ADM~\citep{dhariwal2021diffusion} & 10.94 &  & 100.98 \\
    ADM+upsampling~\citep{dhariwal2021diffusion} & 7.49 &  & 127.49 \\
    CDM (ours) & \textbf{4.88} & 4.63 & 158.71 $\pm$ 2.26 \\ \bottomrule
\end{tabular}
\subcaption{Class-conditional ImageNet sample quality results for classifier guidance-free methods}
\label{table:main_results}
\end{subtable}

\vspace{1em}

\begin{subtable}[c]{\textwidth}\centering
\begin{tabular}{lcc} \toprule
Model & \makecell{Top-1 Accuracy} & \makecell{Top-5 Accuracy} \\ \midrule
128$\times$128 resolution \\ \cmidrule(r){1-1}
    Real & 68.82\% & 88.79\% \\
    BigGAN-deep~\citep{brock2018large} & 40.64\% &  64.44\% \\
    HAM~\citep{de2019hierarchical} & 54.05\% & 77.33\% \\
    CDM (ours) & \textbf{59.84\%} & \textbf{81.79\%}  \\ \midrule
256$\times$256 resolution \\ \cmidrule(r){1-1}
    Real & 73.09\% & 91.47\% \\
    BigGAN-deep~\citep{brock2018large} &  42.65\% & 65.92\% \\
    VQ-VAE-2~\citep{razavi2019generating} &  54.83\% & 77.59\%  \\
    CDM (ours) & \textbf{63.02\%} & \textbf{84.06\%}  \\ \bottomrule
\end{tabular}
\subcaption{Classification Accuracy Score (CAS) results}
\label{table:cas_results}
\end{subtable}
\caption{Main results. Numbers are bolded only when at least two are available for comparison. CAS for real data and other models are from~\citet{ravuris2019cas}.}
\end{table}

\subsection{Baseline Model Improvements}
\label{sec:baseline_improvements}

To set a strong baseline for class-conditional ImageNet generation at the 64$\times$64 resolution, we reproduced and improved upon a 4000 timestep non-cascaded 64$\times$64 class-conditional diffusion model from Improved DDPM~\citep{nichol2021improved}. Our reimplementation used dropout and was trained longer than reported by~\citeauthor{nichol2021improved}; we found that adding dropout generally slowed down convergence of FID and Inception scores, but improved their best values over the course of a longer training period. We further improved the training set FID score and Inception score by adding noise to the trained model's samples using the forward process to the 2000 timestep point, then restarting the reverse process from that point. See~\cref{table:improvements} for the resulting sample quality metrics.

\subsection{Conditioning Augmentation Experiments up to 64$\times$64}
\label{sec:cond_aug_ablations}

Building on our reimplementation in \cref{sec:baseline_improvements}, we verify in a small scale experiment that cascading improves sample quality at the 64$\times$64 resolution. We train a two-stage cascading pipeline that comprises a 16$\times$16 base model and a 16$\times$16$\rightarrow$64$\times$64 super-resolution model. The super-resolution model architecture is identical to the best 64$\times$64 non-cascaded baseline model in \cref{sec:baseline_improvements}, except for the trivial modification of adding in the low resolution image conditioning information by channelwise concatenation at the input~(see \cref{sec:background}).

\begin{figure}[htbp] \centering
\begin{subfigure}[b]{.49\linewidth}
\includegraphics[width=\textwidth]{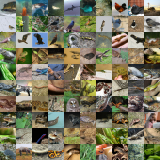}
\caption{16$\times$16 base}
\end{subfigure}
\begin{subfigure}[b]{.49\linewidth}
\includegraphics[width=\textwidth]{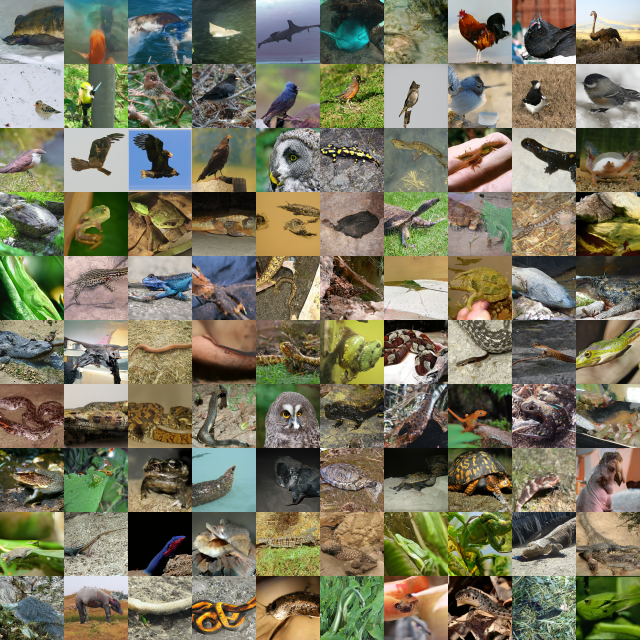}
\caption{16$\times$16$\rightarrow$64$\times$64 super-resolution, $s=0$}
\end{subfigure}
\begin{subfigure}[b]{.49\linewidth}
\includegraphics[width=\textwidth]{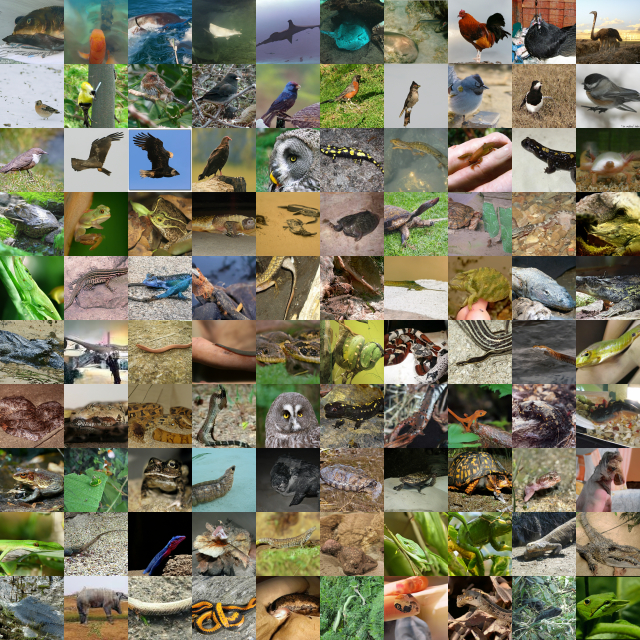}
\caption{16$\times$16$\rightarrow$64$\times$64 super-resolution, $s=101$}
\end{subfigure}
\begin{subfigure}[b]{.49\linewidth}
\adjincludegraphics[width=\textwidth]{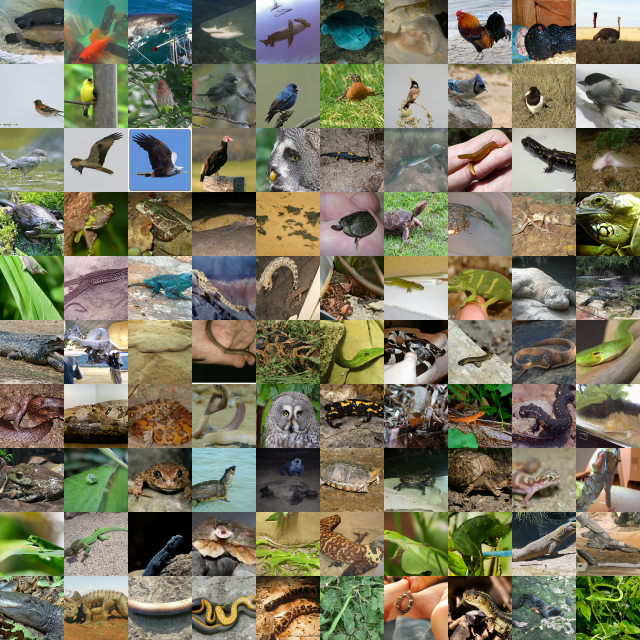}
\caption{16$\times$16$\rightarrow$64$\times$64 super-resolution, $s=1001$}
\end{subfigure}
\caption{Generated images for varying amounts of conditioning augmentation (non-truncated) in a small-scale 16$\times$16$\rightarrow$64$\times$64 pipeline for ablation purposes. Accompanies \cref{table:small_scale_16_64}.}
\label{fig:noisy_conditioning}
\end{figure}

See \cref{table:small_scale_16_64} and \cref{fig:noisy_conditioning} for the results of this 16$\times$16$\rightarrow$64$\times$64  cascading pipeline. Interestingly, we find that without conditioning augmentation, the cascading pipeline attains lower sample quality than the non-cascaded baseline 64$\times$64 model; the FID score, for example, degrades from 2.35 to 6.02. With sufficient conditioning augmentation, however, the sample quality of the cascading pipeline becomes better than the non-cascaded baseline. We train two super-resolution models with non-truncated conditioning augmentation, one at truncation time $s=101$ and another at $s=1001$ (we could have amortized both into a single model, but we chose not to do so in this experiment to prevent potential model capacity issues from confounding the results). The first model achieves better sample quality than the non-augmented model but is still worse than the non-cascaded baseline. The second model achieves a FID score of 2.13, outperforming the non-cascaded baseline. 
Conditioning augmentation is therefore crucial to improve sample quality in this particular cascading pipeline.

\begin{table}[tb]\small
\begin{subtable}[c]{\textwidth}\centering
\begin{tabular}{@{\hspace{.1cm}}l@{\hspace{.2cm}}c@{\hspace{.2cm}}c@{}c@{\hspace{.1cm}}} \toprule
Model & \makecell{FID \\ vs train} & \makecell{FID \\ vs validation} & IS \\ \midrule
Improved DDPM~\citep{nichol2021improved} &   2.92 &  &  \\ \midrule
Our reimplementation & 2.44 & \textbf{2.91} &  49.81 $\pm$ 0.65 \\
+ more sampling steps &  \textbf{2.35} & \textbf{2.91} & \textbf{52.72 $\pm$ 1.15} \\
\bottomrule
\end{tabular}
\subcaption{Improvements to a non-cascaded baseline}
\label{table:improvements}
\end{subtable}

\vspace{1em}

\begin{subtable}[c]{\textwidth}\centering
\begin{tabular}{@{\hspace{.1cm}}l@{\hspace{.2cm}}c@{\hspace{.2cm}}c@{\hspace{.2cm}}c@{\hspace{.1cm}}} \toprule
Conditioning  & \makecell{FID \\ vs train} & \makecell{FID \\ vs validation} & IS \\ \midrule
No cascading  & 2.35 & 2.91 & 52.72 $\pm$ 1.15 \\ \midrule
\multicolumn{4}{l}{16$\times$16$\rightarrow$64$\times$64 cascading} \\ \midrule
$s = 0$ & 6.02 & 5.84 & 35.59 $\pm$ 1.19 \\
$s = 101$ & 3.41 & 3.67 & 44.72 $\pm$ 1.12 \\
$s = 1001$ & \textbf{2.13} & \textbf{2.79} & \textbf{54.47 $\pm$ 1.05} \\
\bottomrule
\end{tabular}
\subcaption{Small-scale ablation comparing no cascading to 16$\times$16$\rightarrow$64$\times$64 cascading}
\label{table:small_scale_16_64}
\end{subtable}
\caption{64$\times$64 ImageNet sample quality: ablations.}
\end{table}

To further improve sample quality at the 64$\times$64 resolution, we found it helpful to increase model sizes and to switch to a cascading pipeline starting with a 32$\times$32 base resolution model. We train a 32$\times$32 base model applying random left-right flips, which we found to help 32$\times$32 scores at the expense of longer training times.  Training without random flips, the best 32$\times$32 resolution FID score is 1.25 at 300k training steps, while training with random flips it is 1.11 at 700k training steps. The 32$\times$32$\rightarrow$64$\times$64 super-resolution model is now amortized over the truncation time $s$ by providing $s$ as an extra time embedding input to the network (\cref{sec:background}), allowing us to perform a more fine grained search over $s$ without retraining the model.

\begin{table}[p]\small
\begin{subtable}[c]{\textwidth}\centering
\begin{tabular}{@{\hspace{.1cm}}l@{\hspace{.2cm}}c@{\hspace{.2cm}}c@{\hspace{.2cm}}c@{\hspace{.1cm}}} \toprule
Conditioning  & \makecell{FID \\ vs train} & \makecell{FID \\ vs validation} & IS \\ \midrule
\multicolumn{4}{l}{No conditioning augmentation (baseline)} \\ \midrule
$s=0$        & 1.71 & 2.46 & 61.34 $\pm$ 1.58 \\ \midrule
\multicolumn{4}{l}{Truncated conditioning augmentation} \\ \midrule
$s=251$    & 1.50  & \textbf{2.44}	& 66.76 $\pm$ 1.76 \\
$s=501$    & \textbf{1.48} & 2.48	& 67.95 $\pm$ 1.97 \\
$s=751$    & \textbf{1.48}	& 2.51	& \textbf{68.48 $\pm$ 1.77} \\
$s=1001$   & 1.49	& 2.51	& 67.95 $\pm$ 1.51 \\
$s=1251$   & 1.51	& 2.54	& 67.20 $\pm$ 1.94  \\
$s=1501$   & 1.54	& 2.56	& 67.09 $\pm$ 1.67 \\ \midrule
\multicolumn{4}{l}{Non-truncated conditioning augmentation} \\ \midrule
$s=251$	   & 1.58  & 2.50	& 66.21 $\pm$ 1.51 \\
$s=501$    & 1.53  & 2.51	& \textbf{67.59 $\pm$ 1.85} \\
$s=751$   	& \textbf{1.48}	& 2.47	& 67.48 $\pm$ 1.31 \\
$s=1001$ 	& 1.49	& 2.48	& 66.51 $\pm$ 1.59 \\
$s=1251$ 	& \textbf{1.48}	& \textbf{2.46}	& 66.28 $\pm$ 1.49 \\
$s=1501$ 	& 1.50	& 2.47	& 65.59 $\pm$ 0.86 \\
\bottomrule
\end{tabular}
\subcaption{Base model for low resolution conditioning}
\label{table:cond_q_steps}
\end{subtable}

\vspace{1em}

\begin{subtable}[c]{\textwidth}\centering
\begin{tabular}{@{\hspace{.1cm}}l@{\hspace{.2cm}}c@{\hspace{.2cm}}c@{\hspace{.2cm}}c@{\hspace{.1cm}}} \toprule
Conditioning  & \makecell{FID \\ vs train} & \makecell{FID \\ vs validation} & IS  \\ \midrule
\multicolumn{4}{l}{Ground truth training data} \\ \midrule
$s=0$             & \textbf{0.76} & \textbf{1.76} & \textbf{74.84 $\pm$ 1.43} \\
$s=251$         & 0.87 & 1.85 & 71.79 $\pm$ 0.89 \\
$s=501$         & 0.92 & 1.91 & 70.68 $\pm$ 1.26 \\
$s=751$         & 0.95  & 1.94  & 69.93 $\pm$ 1.40 \\
$s=1001$        & 0.98 & 1.97 & 69.03 $\pm$ 1.26 \\
$s=1251$        & 1.03  & 1.99 & 67.92 $\pm$ 1.65 \\
$s=1501$        & 1.11  & 2.04 & 66.7 $\pm$ 1.21  \\ \midrule
\multicolumn{4}{l}{Ground truth validation data} \\ \midrule
$s=0$           & \textbf{1.20}   & \textbf{0.59} & \textbf{64.33 $\pm$ 1.24} \\
$s=251$         & 1.27   & 0.96 & 63.17 $\pm$ 1.19 \\
$s=501$         & 1.32   & 1.17  & 62.65 $\pm$ 0.76 \\
$s=751$         & 1.38    & 1.32  & 62.21 $\pm$ 0.94 \\
$s=1001$        & 1.42   & 1.44  & 61.53 $\pm$ 1.39 \\
$s=1251$        & 1.47   & 1.54  & 60.58 $\pm$ 0.93 \\
$s=1501$        & 1.53   & 1.64  & 60.02 $\pm$ 0.84 \\
\bottomrule
\end{tabular}
\subcaption{Ground truth for low resolution conditioning}
\label{table:cond_q_steps_groundtruth32}
\end{subtable}
\caption{64$\times$64 ImageNet sample quality: large scale experiment comparing truncated and non-truncated conditioning augmentation for 32$\times$32$\rightarrow$64$\times$64 cascading, using amortized truncation time conditioning.}
\end{table}

\Cref{table:cond_q_steps} displays the resulting sample quality scores for both truncated and non-truncated augmentation. The sample quality metrics improve and then degrade non-monotonically as the truncation time is increased. This indicates that moderate amounts of conditioning augmentation are beneficial to sample quality of the cascading pipeline, but too much conditioning augmentation causes the super-resolution model to behave as a non-conditioned model unable to benefit from cascading.
For comparison, \cref{table:cond_q_steps_groundtruth32} shows sample quality when the super-resolution model is conditioned on ground truth data instead of generated data. Here, sample quality monotonically degrades as truncation time is increased. Conditioning augmentation is therefore useful precisely when conditioning on generated samples, so as a technique it is uniquely suited to cascading pipelines.

Based on these findings on non-monotonicity of sample quality with respect to truncation time, we conclude that conditioning augmentation works because it alleviates compounding error from a train-test mismatch for the super-resolution model. This occurs when low-resolution model samples are out of distribution compared to the ground truth data on which the super-resolution model is trained.  A sufficient amount of Gaussian conditioning augmentation prevents the super-resolution model from attempting to upsample erroneous, out-of-distribution details in the low resolution generated samples. In contrast, sample quality degrades monotonically with respect to truncation time when conditioning the super-resolution model on ground truth data, because there is no such train-test mismatch.

\Cref{table:cond_q_steps} additionally shows that truncated and non-truncated conditioning augmentation are approximately equally effective at improving sample quality of the cascading pipeline, albeit at different values of the truncation time parameter. Thus we generally recommend non-truncated augmentation due to its practical benefits described in~\cref{sec:non_truncated_gaussian_aug}.

\subsection{Experiments at 128$\times$128 and 256$\times$256} \label{sec:experiments_128_256}

\begin{figure}[p]\centering
    \resizebox{0.5\linewidth}{!}{{\pgfplotsset{compat=1.3}
\begin{tikzpicture}
\begin{axis}[
    xlabel={Inference Steps},
    ylabel={FID},
    ymin=4.5, ymax=5.8,
    xmin=0, xmax=7,
    xticklabels={4, 8, 16, 25, 100, 1000}, xtick={1,2,3,4,5,6},
    ytick={4.6, 4.9, 5.3, 5.6},,
    ymajorgrids=true,
    grid style=dashed,
    legend pos=south west,
    legend cell align=left
]
\addplot[
    color=blue,
    mark=*
    ]
    coordinates {
    (1,5.42)(2,5.45)(3,5.55)(4,5.51)(5,4.87)(6,4.85)
    };
\addplot[
    color=red,
    mark=square*,
    very thick,
    dotted,
    mark options={solid}
    ]
    coordinates {
    (1,5.52)(2,5.50)(3,5.48)(4,5.38)(5,4.63)(6,4.54)
    };
    \legend{FID vs Train,
    FID vs Validation}
\end{axis}
\end{tikzpicture}}}
    \caption{FID on 256$\times$256 images vs inference steps in 64$\times$64 $\rightarrow$ 256$\times$256 super-resolution.}
    \label{fig:score_vs_step}
\end{figure}
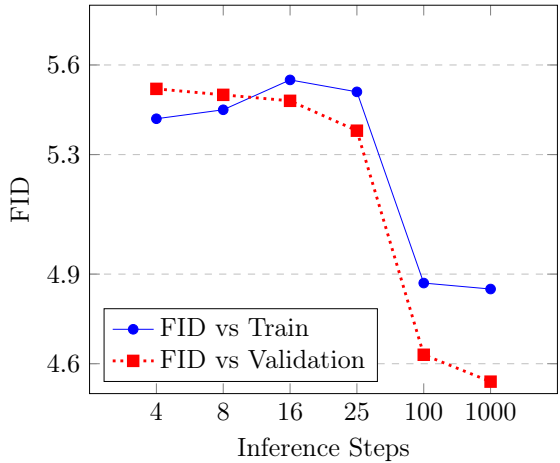

While we found Gaussian noise augmentation to be a key ingredient to boost the performance of our cascaded models at low resolutions, our initial experiments with similar augmentations for 128$\times$128 and 256$\times$256 upsampling yielded negative results. Hence, we explore Gaussian blurring augmentation for these resolutions. As mentioned in \cref{sec:blur_aug}, we apply the blurring augmentation 50\% of the time during training, and use no blurring during inference. We explored other settings (e.g. applying blurring to all training examples, and using varying amounts of blurring during inference), but found this to be most effective in our initial experiments.

\begin{table}[p]\small
\begin{subtable}[c]{\textwidth}\centering
\begin{tabular}{@{\hspace{.1cm}}l@{\hspace{.2cm}}c@{\hspace{.2cm}}c@{\hspace{.2cm}}c@{\hspace{.1cm}}} \toprule
Blur $\sigma$ & \makecell{FID \\ vs train} & \makecell{FID \\ vs validation} & IS \\ \midrule
$\sigma = 0$ (no blur)                        & 7.26 & 6.42 & 134.53 $\pm$ 2.97\\ \midrule
$\sigma \sim \mathcal{U}(0.4, 0.6)$           & \textbf{6.18} & \textbf{5.57} & \textbf{142.71 $\pm$ 2.83}\\
$\sigma \sim \mathcal{U}(0.4, 0.8)$           & 6.90 & 6.31 & 136.57 $\pm$ 4.34\\
$\sigma \sim \mathcal{U}(0.4, 1.0)$           & 6.35 & 5.76 & 141.40 $\pm$ 4.34\\
\bottomrule
\end{tabular}
\caption{Gaussian blur noise in conditioning}
\label{table:cond_blur}
\end{subtable}

\vspace{1cm}

\begin{subtable}[c]{\textwidth}\centering
\begin{tabular}{@{\hspace{.1cm}}l@{\hspace{.2cm}}c@{\hspace{.2cm}}c@{\hspace{.2cm}}c@{\hspace{.1cm}}} \toprule
Model  & \makecell{FID \\ vs train} & \makecell{FID \\ vs validation} & IS \\ \midrule
Baseline                         & 6.18 & 5.57 & 142.71 $\pm$ 2.83\\ \midrule
+ Class Conditioning             & 5.75 & 5.27 & 152.17 $\pm$ 2.29 \\
+ Large Batch Training           & 5.00 & 4.71 & 157.84 $\pm$ 2.60 \\
+ Flip LR                        & \textbf{4.88} & \textbf{4.63} & \textbf{158.71 $\pm$ 2.26} \\
\bottomrule
\end{tabular}
\caption{Further improvements on super-resolution}
\label{table:cond_blur_2}
\end{subtable}
\caption{256$\times$256 ImageNet sample quality: experiments on 64$\times$64 $\rightarrow$ 256$\times$256 super-resolution.}
\end{table}

\Cref{table:cond_blur} shows the results of applying Gaussian blur augmentation to the  64$\times$64 $\rightarrow$ 256$\times$256 super-resolution model. While any amount of blurring helps improve the scores of the 256$\times$256 samples over the baseline model with no blur, we found that sampling $\sigma \sim \mathcal{U}(0.4, 0.6)$ gives the best results. 
\Cref{table:cond_blur_2} shows further improvements from class conditioning, large batch training, and random flip augmentation for the super-resolution model. While we find class conditioning helpful for upsampling at low resolution settings, it is interesting that it still gives a huge boost to the upsampling performance at high resolutions even when the low resolution inputs at 64$\times$64 can be sufficiently informative. We also found increasing the training batch size from 256 to 1024 further improved performance by a significant margin. We also obtain marginal improvements by training the super-resolution model on randomly flipped data.

Since the sampling cost increases quadratically with the target image resolution, we attempt to minimize the number of denoising iterations for our 64$\times$64 $\rightarrow$ 256$\times$256 and  64$\times$64 $\rightarrow$ 128$\times$128 super-resolution models. To this end, we train these super-resolution models with continuous noise conditioning, like \citet{saharia2021image} and \citet{chen2020wavegrad}, and tune the noise schedule for a given number of steps during inference. This tuning is relatively inexpensive as we do not need to retrain the models.  We report all results using 100 inference steps for these models.  \Cref{fig:score_vs_step} shows FID vs number of inference steps for our 64$\times$64 $\rightarrow$ 256$\times$256 model. The FID score deteriorates marginally even when using just 4 inference steps. Interestingly, we do not observe any concrete improvement in FID by increasing the number of inference steps from 100 to 1000.

\subsection{Experiments on LSUN}
\label{sec:experiments_lsun}

While the main results of this work are on class-conditional ImageNet generation, here we study the effectiveness of non-truncated conditioning augmentation for a 64$\times$64$\rightarrow$128$\times$128 cascading pipeline on the LSUN Bedroom and Church datasets~\citep{yu15lsun} in order to verify that conditioning augmentation is not an ImageNet-specific method. LSUN Bedroom and Church are two separate unconditional datasets that do not have any class labels, so our study here additionally verifies the effectiveness of conditioning augmentation for unconditional generation.

\Cref{table:lsun} displays our LSUN sample quality results, which confirm that a nonzero amount of conditioning augmentation is beneficial to sample quality. (The relatively large FID scores between generated examples and the validation sets are explained by the fact that the LSUN Church and Bedroom validation sets are extremely small, consisting of only 300 examples each.) We observe a similar effect as our ImageNet results in \cref{table:cond_q_steps_groundtruth32}: because the super-resolution model is conditioned on base model samples, the sample quality improves then degrades non-monotonically as the truncation time $s$ is increased. See \cref{appendix:samples} for examples of images generated by our LSUN models.

\begin{table}[htb]\small \centering
\begin{tabular}{@{\hspace{.1cm}}l@{\hspace{.2cm}}c@{\hspace{.2cm}}c} \toprule
Conditioning  & \makecell{FID \\ vs train} & \makecell{FID \\ vs validation}  \\ \midrule
\multicolumn{3}{l}{LSUN Bedroom} \\ \midrule
$s=0$           & 2.30 & 40.68 \\
$s=251$         & \textbf{2.06} & 40.47 \\
$s=501$         & 2.08 & \textbf{40.44} \\
$s=751$         & 2.14 & 40.45 \\
$s=1001$        & 2.18 & 40.53 \\
$s=1251$        & 2.24 & 40.58 \\
$s=1501$        & 2.28 & 40.58 \\ \midrule
\multicolumn{3}{l}{LSUN Church} \\ \midrule
$s=0$           & 3.29 & 42.21 \\
$s=251$         & 2.97 & \textbf{42.14} \\
$s=501$         & 2.93 & 42.17 \\
$s=751$         & 2.89 & 42.20 \\
$s=1001$        & 2.86 & 42.26 \\
$s=1251$        & \textbf{2.83} & 42.28 \\
$s=1501$        & 2.84 & 42.31 \\
\bottomrule
\end{tabular}
\caption{128$\times$128 LSUN sample quality: non-truncated conditioning augmentation for a 64$\times$64$\rightarrow$128$\times$128 cascading pipeline using the base model for low resolution conditioning.}
\label{table:lsun}
\end{table}

\section{Related Work}
\label{sec:related_work}

One way to formulate cascaded diffusion models is to modify the original diffusion formalism of a forward process $q(\bx_{0:T})$ at single resolution so that the transition $q(\bx_{t}|\bx_{t-1})$ performs downsampling at certain intermediate timesteps, for example at $t \in S \defeq \{T/4, 2T/4, 3T/4\}$. The reverse process would then be required to perform upsampling at those timesteps, similar to our cascaded models here. However, there is no guarantee that the reverse transitions at the timesteps in $S$ are conditional Gaussian, unlike the guarantee for reverse transitions at other timesteps for sufficiently slow diffusion. By contrast, our cascaded diffusion model construction dedicates entire conditional diffusion models for these upsampling steps, so it is specified more flexibly.

Recent interest in diffusion models~\citep{sohl2015deep} started with work connecting diffusion models to denoising score matching over multiple noise scales~\citep{ho2020denoising,song2019generative}. There have been a number of improvements and alternatives proposed to the diffusion framework, for example generalization to continuous time~\citep{song2020score}, deterministic sampling~\citep{song2020denoising}, adversarial training~\citep{jolicoeur2020adversarial}, and others~\citep{gao2020learning}. For simplicity, we base our models on DDPM~\citep{ho2020denoising} with modifications from Improved DDPM~\citep{nichol2021improved} to stay close to the original diffusion framework.

Cascading pipelines have been investigated in work on VQ-VAEs~\citep{oord2016conditional,razavi2019generating} and autoregressive models~\citep{menick2018generating}. Cascading pipelines have also been investigated for diffusion models, such as SR3~\citep{saharia2021image}, Improved DDPM~\citep{nichol2021improved}, and concurrently in ADM~\citep{dhariwal2021diffusion}. Our work here focuses on improving cascaded diffusion models for ImageNet generation and is distinguished by the extensive study on conditioning augmentation and deeper cascading pipelines. Our conditioning augmentation work also resembles scheduled sampling in autoregressive sequence generation \citep{bengio2015scheduled}, where noise is used to alleviate the mismatch between train and inference conditions.

Concurrent work~\citep{dhariwal2021diffusion} showed that diffusion models are capable of generating high quality ImageNet samples using an improved architecture, named ADM, and a classifier guidance technique in which a class-conditional diffusion model sampler is modified to simultaneously take gradient steps to maximize the score of an extra trained image classifier. By contrast, our work focuses solely on improving sample quality by cascading, so we avoid introducing extra model elements such as the image classifier. We are interested in avoiding classifier guidance because the FID and Inception score sample quality metrics that we use to evaluate our models are themselves computed on activations of an image classifier trained on ImageNet, and therefore classifier guidance runs the risk of cheating these metrics.

Avoiding classifier guidance comes at the expense of using thousands of diffusion timesteps in our low resolution models, where ADM uses hundreds. ADM with classifier guidance outperforms our models in terms of FID and Inception scores, while our models outperform ADM without classifier guidance as reported by~\citeauthor{dhariwal2021diffusion}. Our work is a showcase of the effectiveness of cascading alone in a pure generative model, and since classifier guidance and cascading complement each other as techniques to improve sample quality and can be applied together, we expect classifier guidance would improve our results too.

\section{Conclusion}
\label{sec:conclusion}

We have shown that cascaded diffusion models are capable of outperforming state-of-the-art generative models on the ImageNet class-conditional generation benchmark when paired with conditioning augmentation, our technique of introducing data augmentation into the conditioning information of super-resolution models. Our models outperform BigGAN-deep and VQ-VAE-2 as measured by FID score and classification accuracy score. We found that conditioning augmentation helps sample quality because it combats compounding error in cascading pipelines due to  train-test mismatch in super-resolution models, and we proposed practical methods to train and test models amortized over varying levels of conditioning augmentation.

Although there could be negative impact of our work in the form of malicious uses of image generation, our work has the potential to improve beneficial downstream applications such as data compression while advancing the state of knowledge in fundamental machine learning problems. We see our results as a conceptual study of the image synthesis capabilities of diffusion models in their original form with minimal extra techniques, and we hope our work serves as inspiration for future advances in the capabilities of diffusion models.

\acks{We thank Jascha Sohl-Dickstein, Douglas Eck and the Google Brain team for feedback, research discussions and technical assistance.}

\newpage

\appendix

\section{Samples}
\label{appendix:samples}

\begin{figure}[H]
\vspace*{-0.5cm}
\setlength{\tabcolsep}{1.25pt}
\centering
\begin{tabular}{ccccc}
\includegraphics[width=0.2\textwidth]{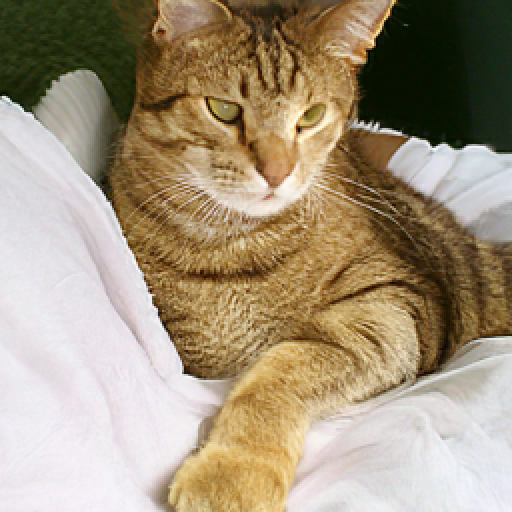} &
\includegraphics[width=0.2\textwidth]{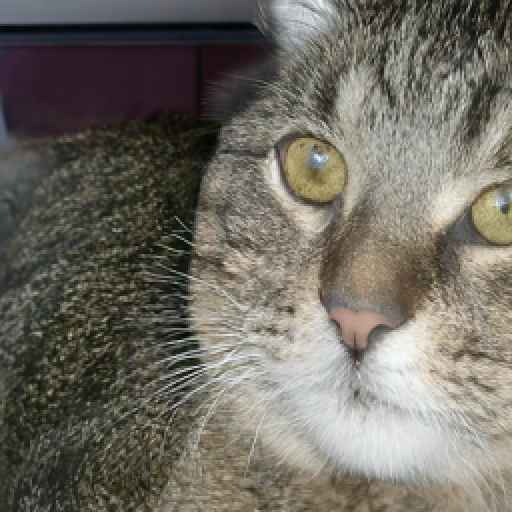} &
\includegraphics[width=0.2\textwidth]{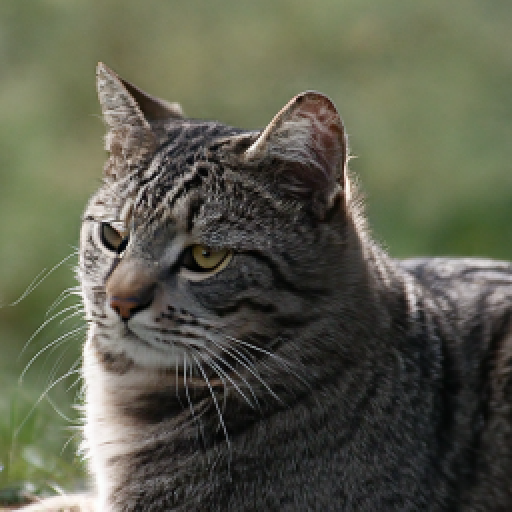} &
\includegraphics[width=0.2\textwidth]{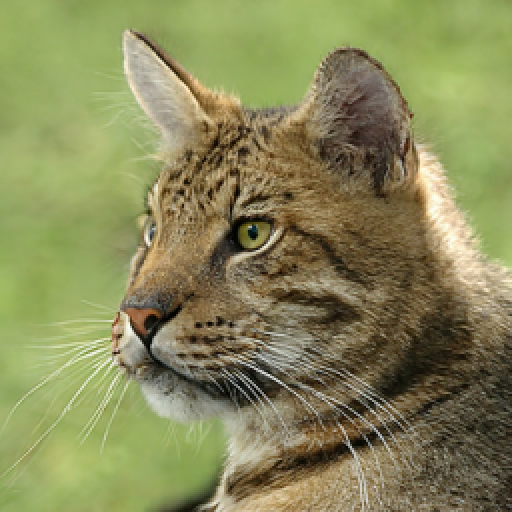} &
\includegraphics[width=0.2\textwidth]{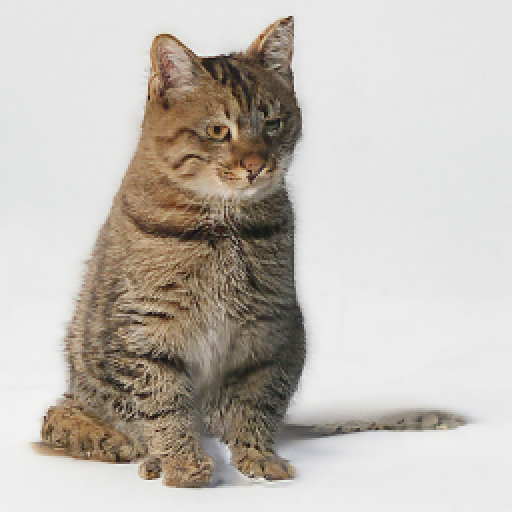} \\

\includegraphics[width=0.2\textwidth]{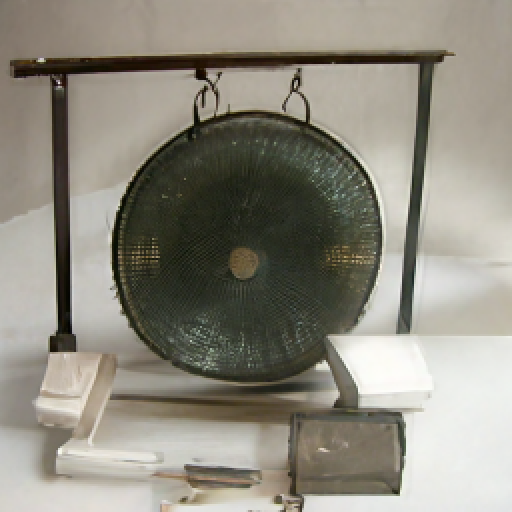} &
\includegraphics[width=0.2\textwidth]{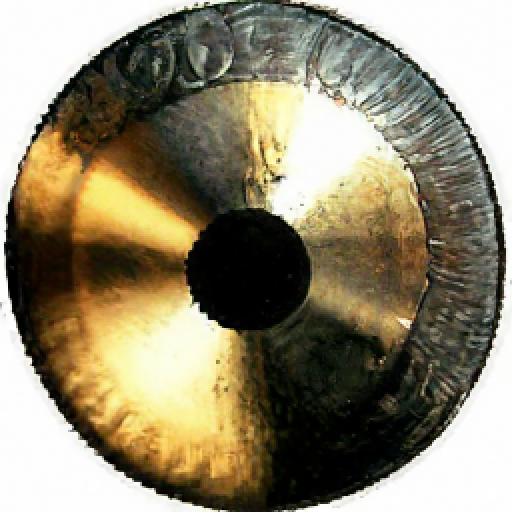} &
\includegraphics[width=0.2\textwidth]{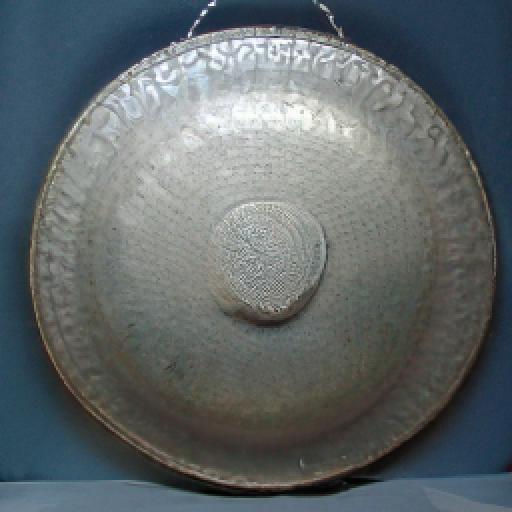} &
\includegraphics[width=0.2\textwidth]{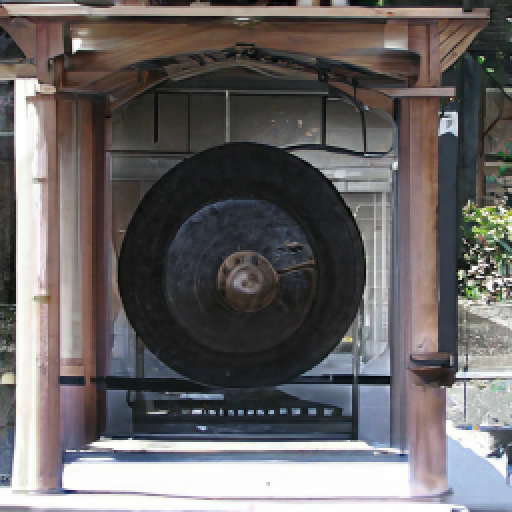} &
\includegraphics[width=0.2\textwidth]{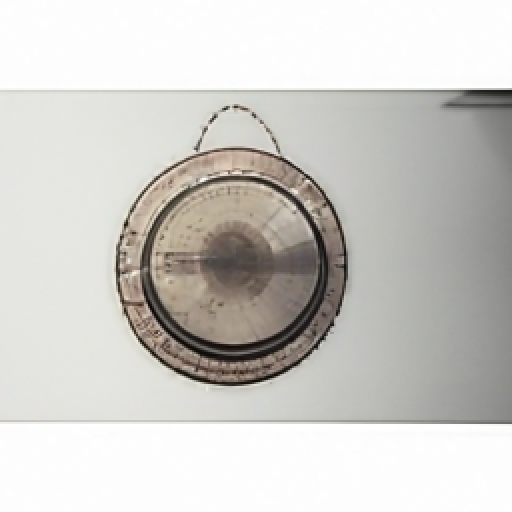} \\

\includegraphics[width=0.2\textwidth]{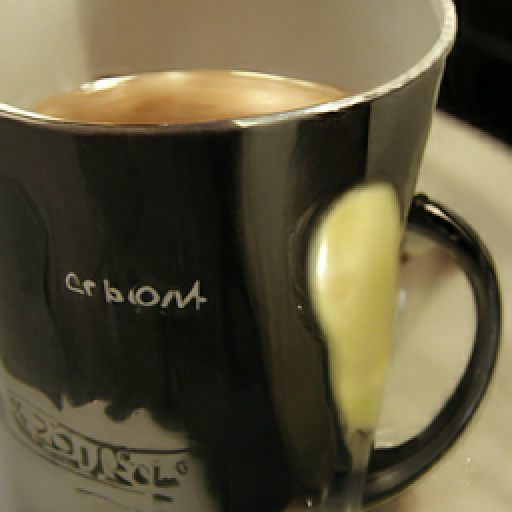} &
\includegraphics[width=0.2\textwidth]{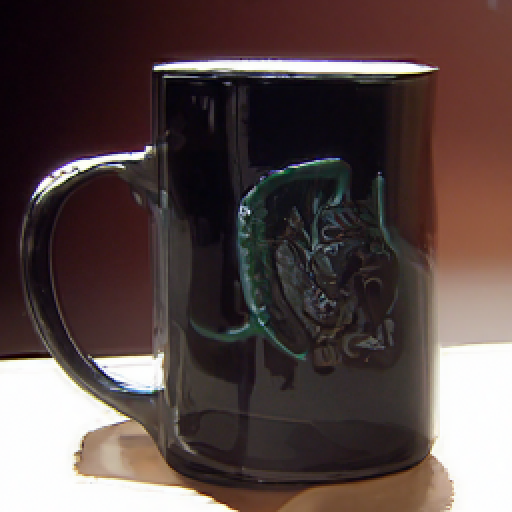} &
\includegraphics[width=0.2\textwidth]{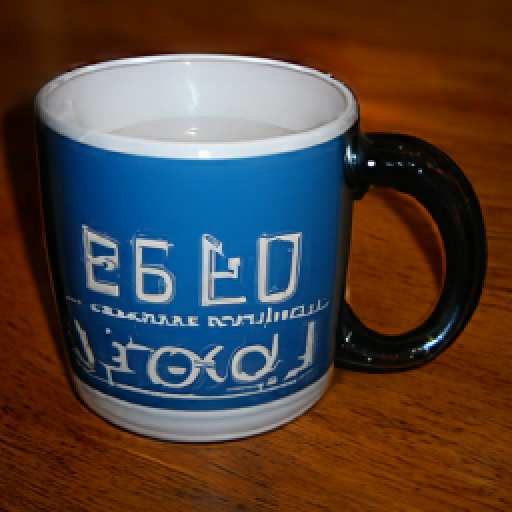} &
\includegraphics[width=0.2\textwidth]{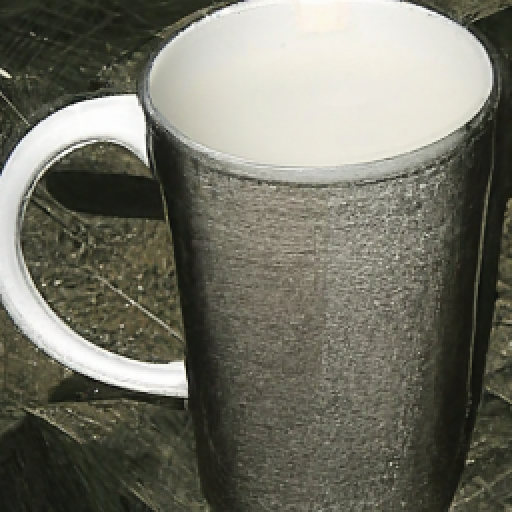} &
\includegraphics[width=0.2\textwidth]{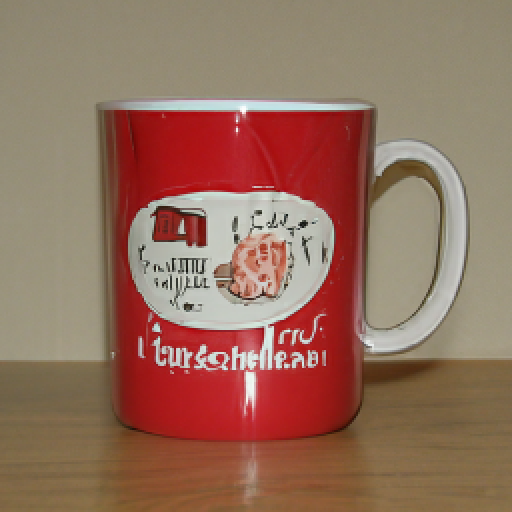} \\

\includegraphics[width=0.2\textwidth]{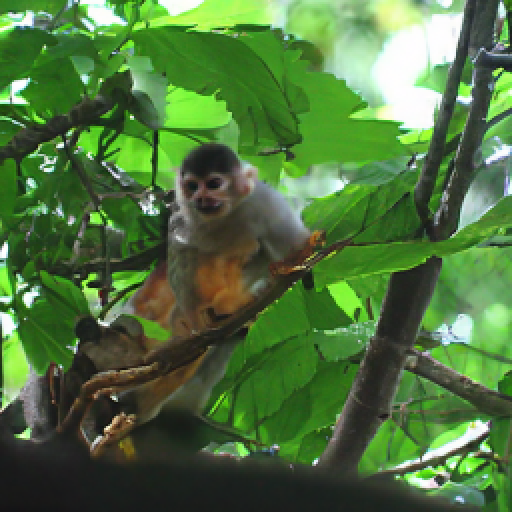} &
\includegraphics[width=0.2\textwidth]{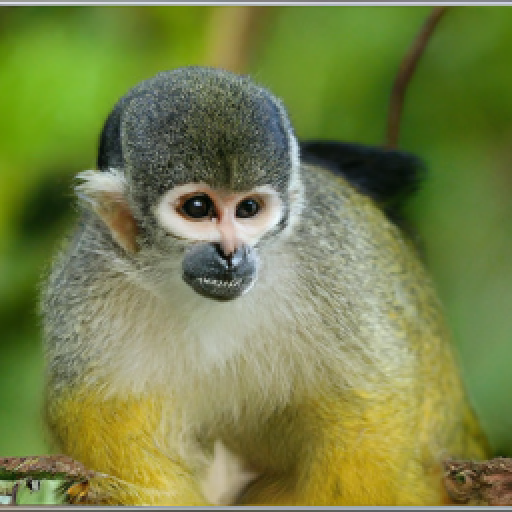} &
\includegraphics[width=0.2\textwidth]{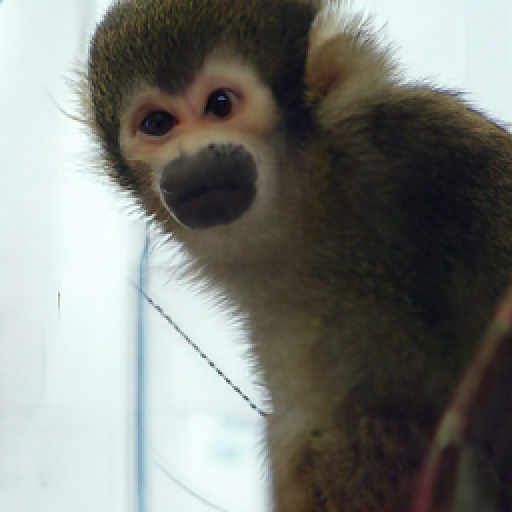} &
\includegraphics[width=0.2\textwidth]{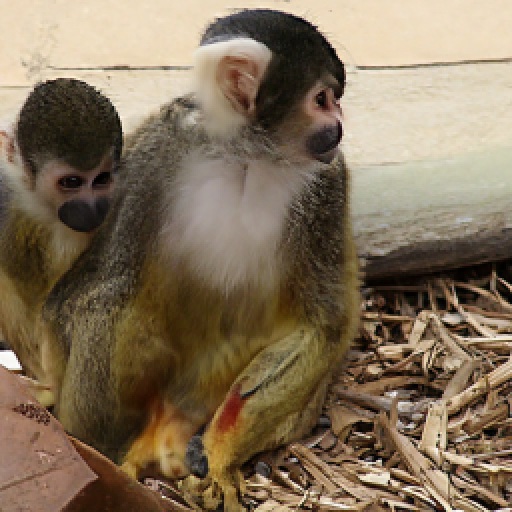} &
\includegraphics[width=0.2\textwidth]{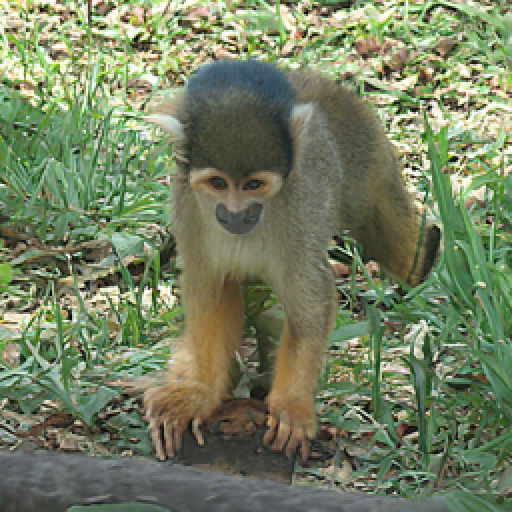} \\

\includegraphics[width=0.2\textwidth]{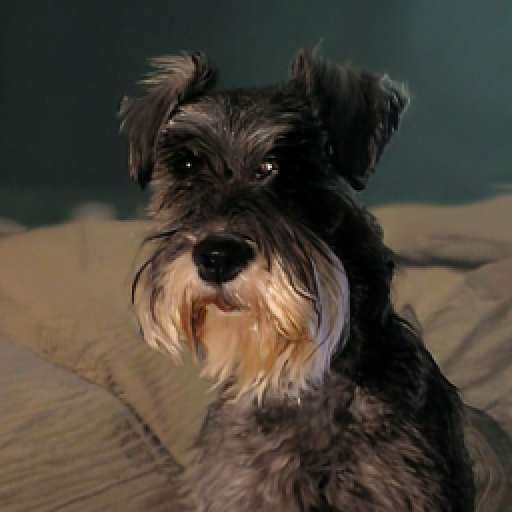} &
\includegraphics[width=0.2\textwidth]{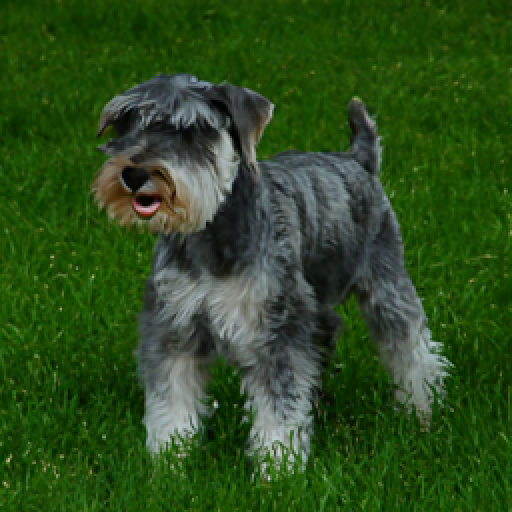} &
\includegraphics[width=0.2\textwidth]{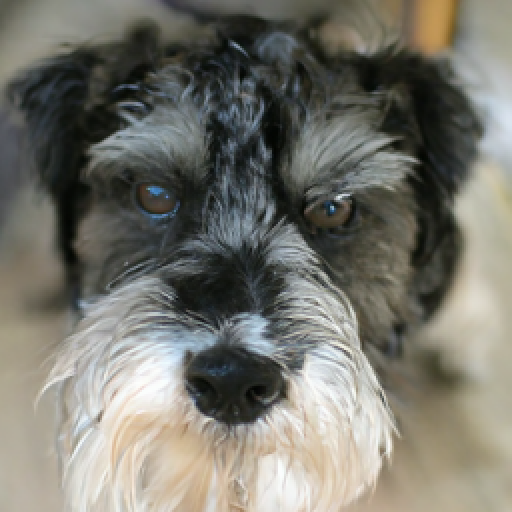} &
\includegraphics[width=0.2\textwidth]{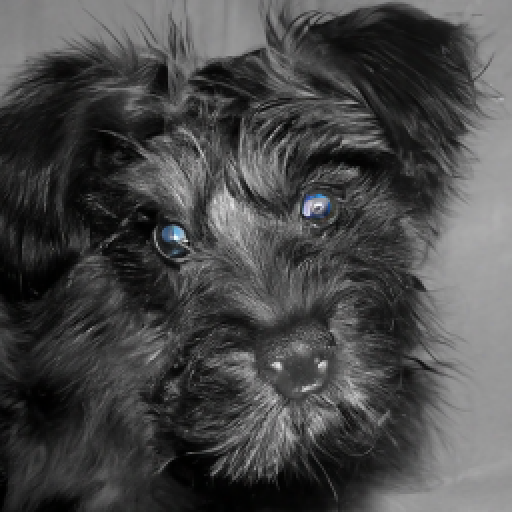} &
\includegraphics[width=0.2\textwidth]{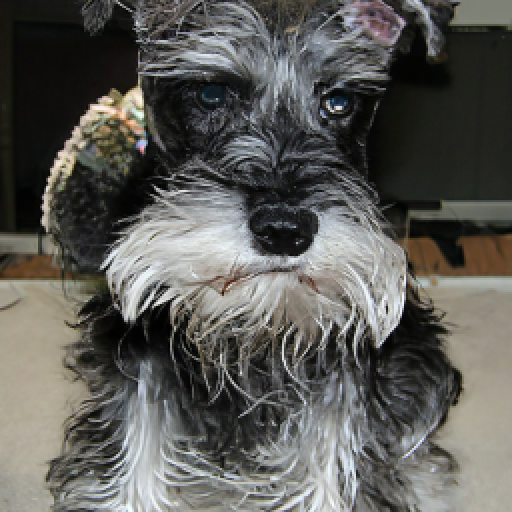} \\

\includegraphics[width=0.2\textwidth]{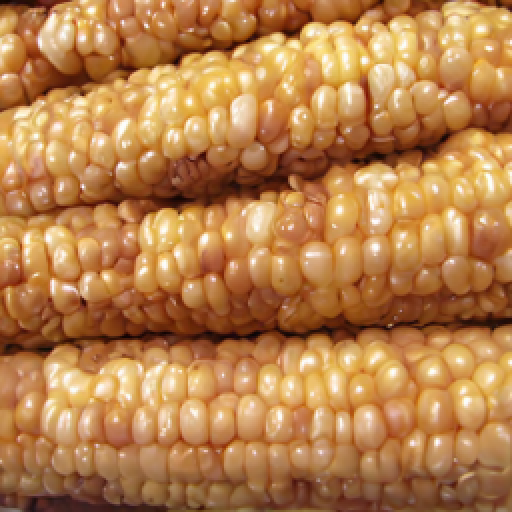} &
\includegraphics[width=0.2\textwidth]{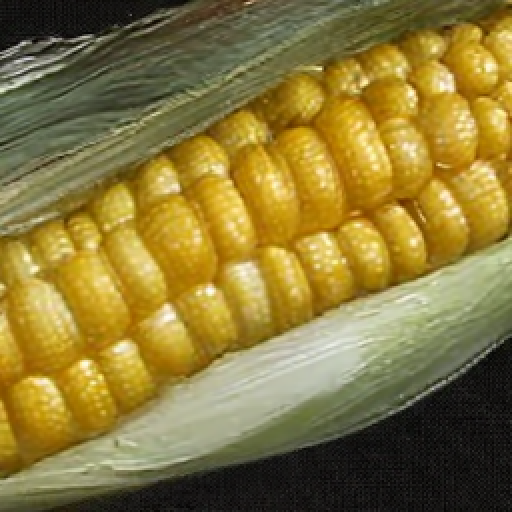} &
\includegraphics[width=0.2\textwidth]{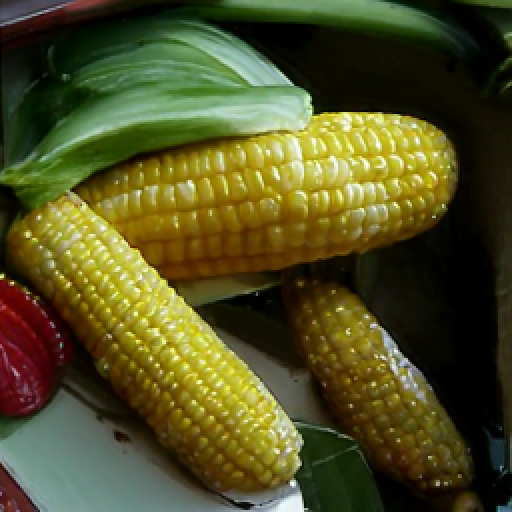} &
\includegraphics[width=0.2\textwidth]{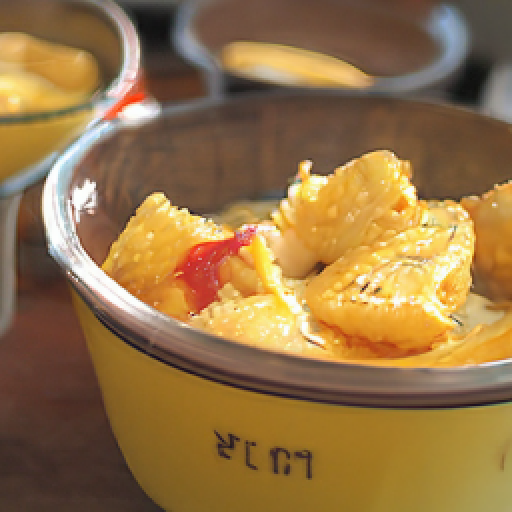} &
\includegraphics[width=0.2\textwidth]{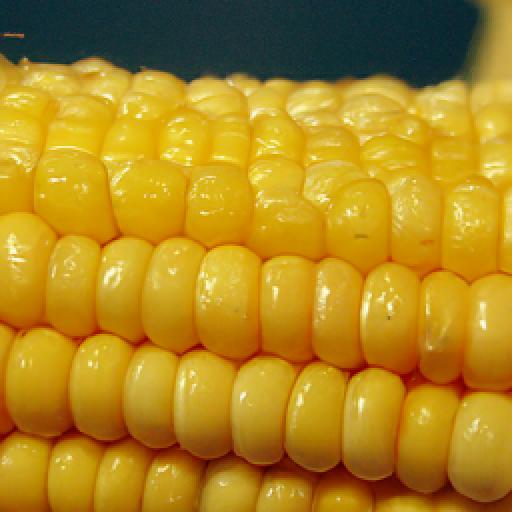}
\end{tabular}
\vspace*{-0.4cm}
\caption{\small Samples from classes with best relative classification accuracy score. Each row represents a specific ImageNet class. Classes from top to bottom - Tiger Cat (282), Gong (577), Coffee Mug (504), Squirrel Monkey (382), Miniature Schnauzer (196) and Corn (987).}
\vspace*{-0.2cm}
\label{fig:cas_classwise_1}
\end{figure}

\begin{figure}[H]
\vspace*{-0.5cm}
\setlength{\tabcolsep}{1.25pt}
\centering
\begin{tabular}{ccccc}
\includegraphics[width=0.2\textwidth]{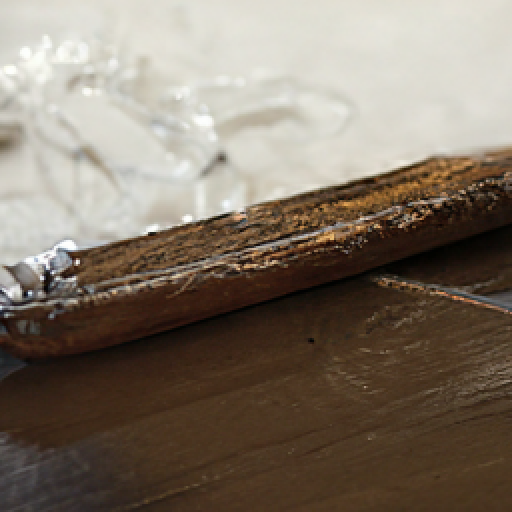} &
\includegraphics[width=0.2\textwidth]{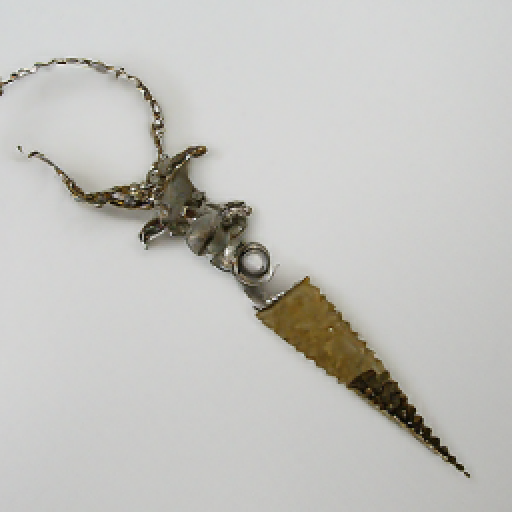} &
\includegraphics[width=0.2\textwidth]{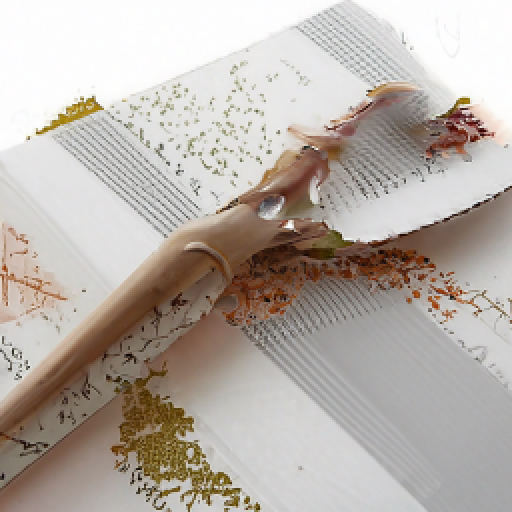} &
\includegraphics[width=0.2\textwidth]{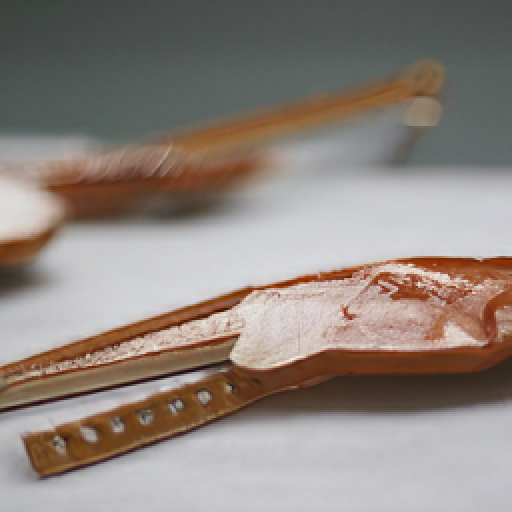} &
\includegraphics[width=0.2\textwidth]{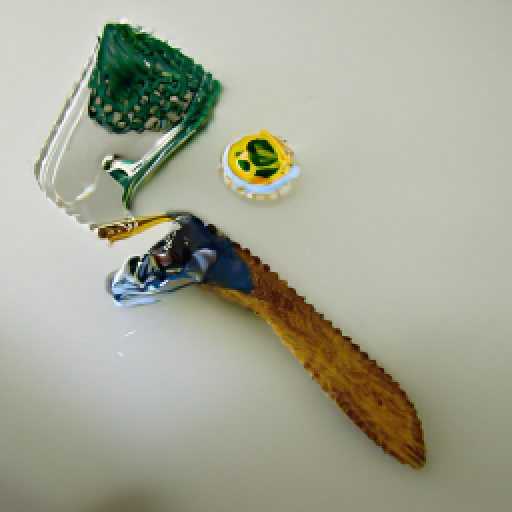} \\

\includegraphics[width=0.2\textwidth]{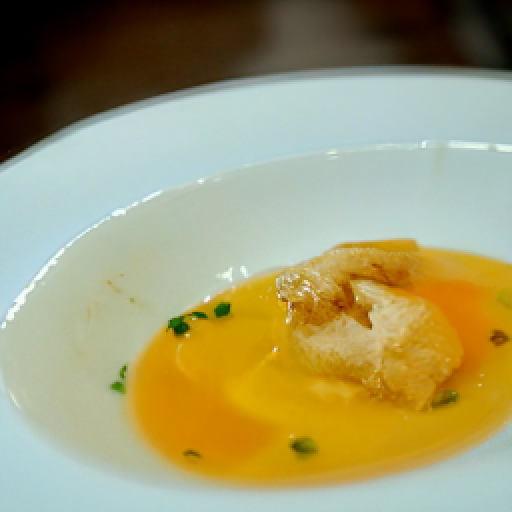} &
\includegraphics[width=0.2\textwidth]{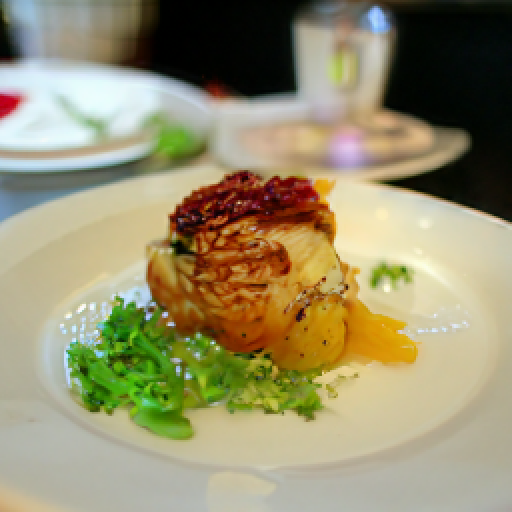} &
\includegraphics[width=0.2\textwidth]{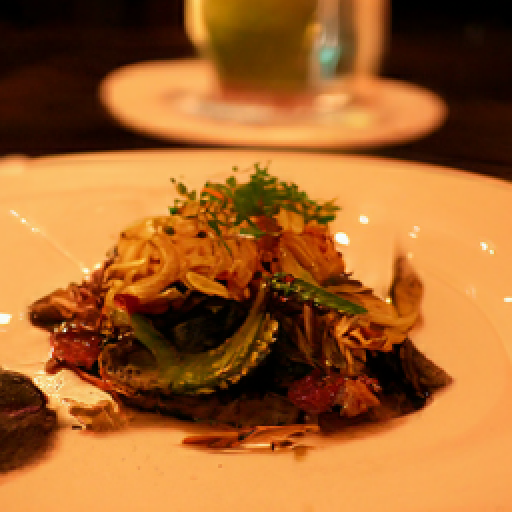} &
\includegraphics[width=0.2\textwidth]{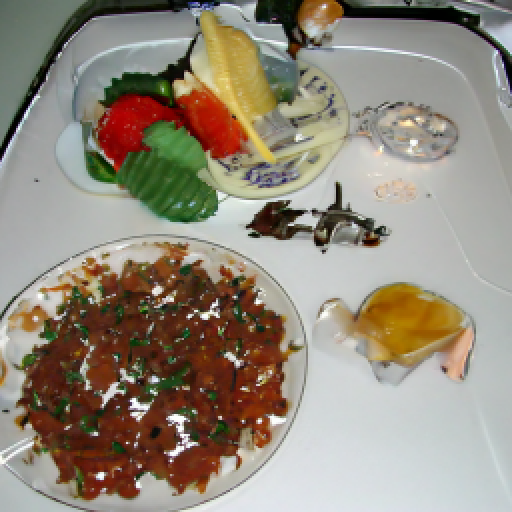} &
\includegraphics[width=0.2\textwidth]{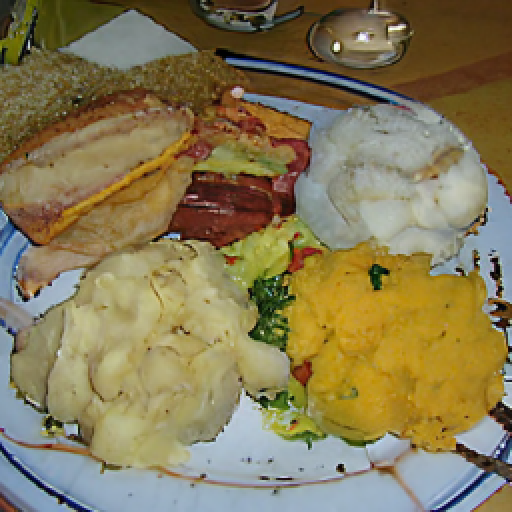} \\

\includegraphics[width=0.2\textwidth]{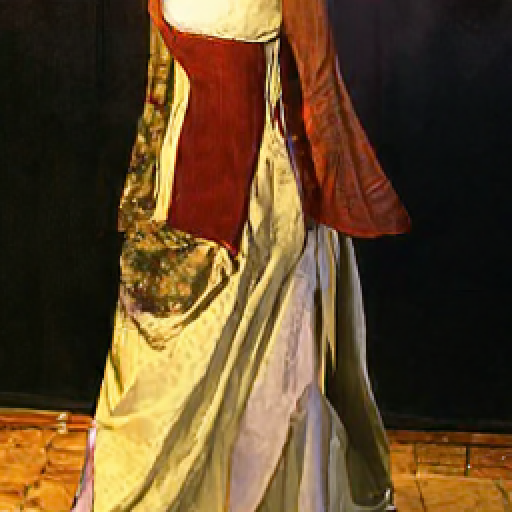} &
\includegraphics[width=0.2\textwidth]{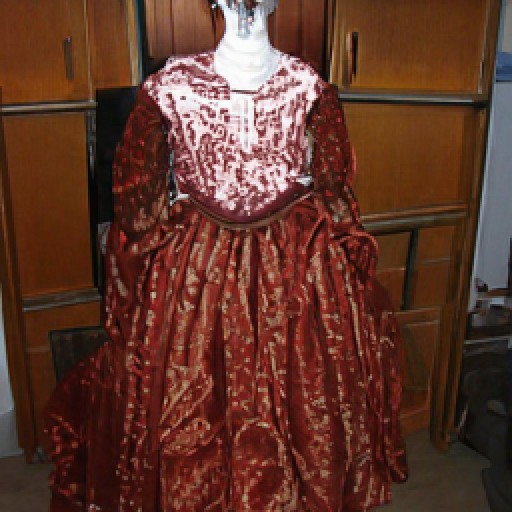} &
\includegraphics[width=0.2\textwidth]{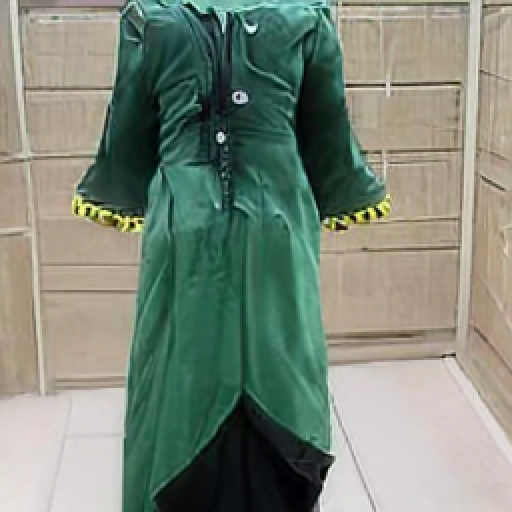} &
\includegraphics[width=0.2\textwidth]{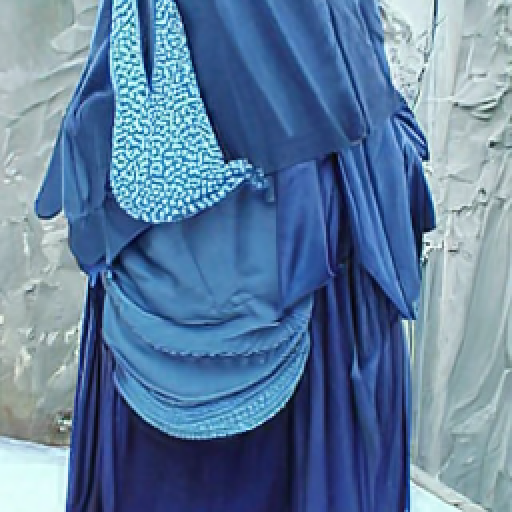} &
\includegraphics[width=0.2\textwidth]{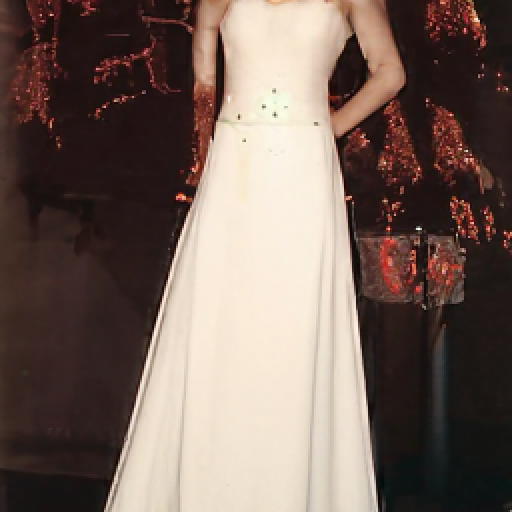} \\

\includegraphics[width=0.2\textwidth]{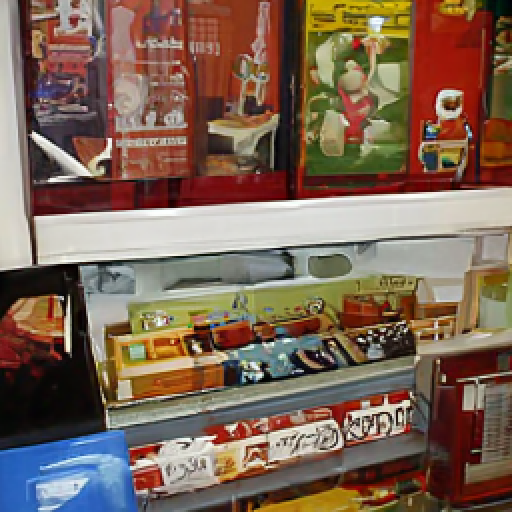} &
\includegraphics[width=0.2\textwidth]{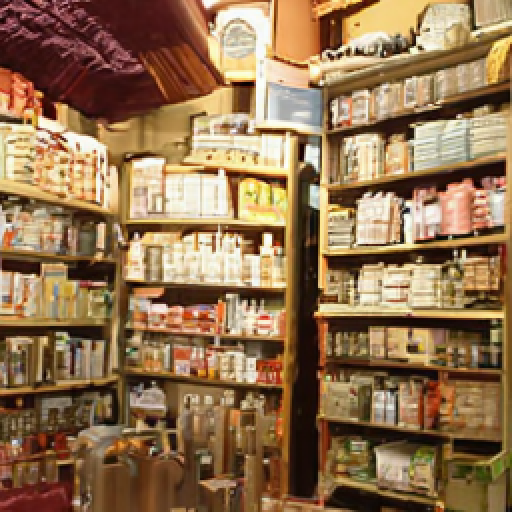} &
\includegraphics[width=0.2\textwidth]{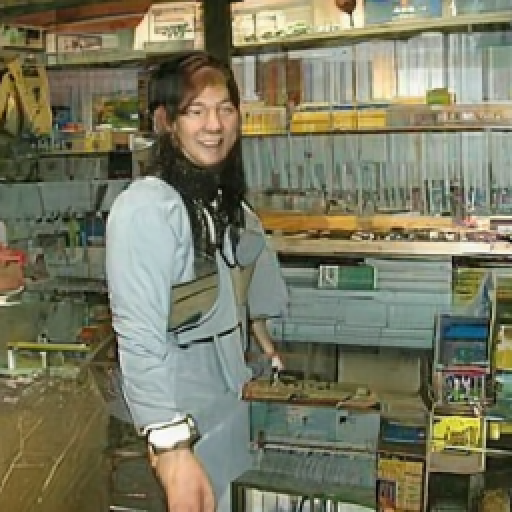} &
\includegraphics[width=0.2\textwidth]{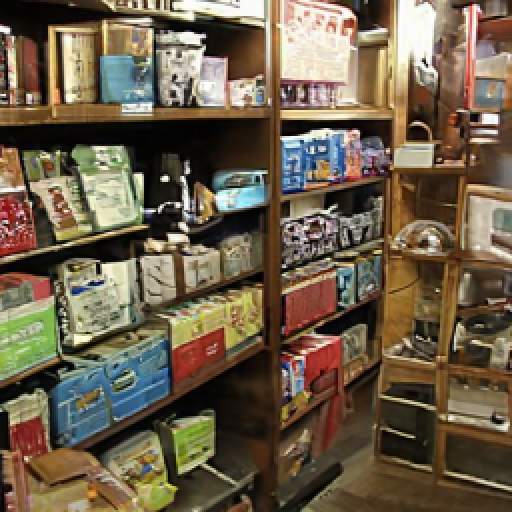} &
\includegraphics[width=0.2\textwidth]{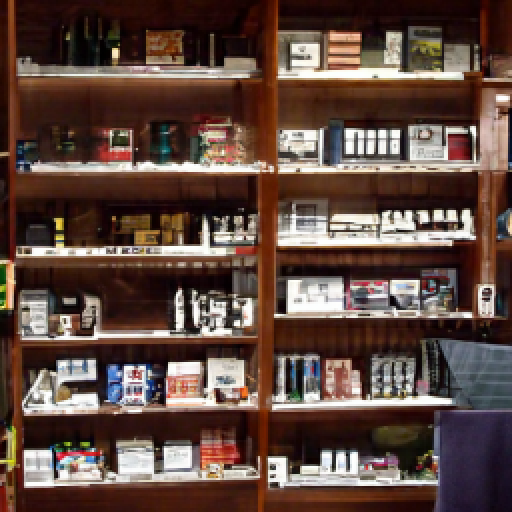} \\

\includegraphics[width=0.2\textwidth]{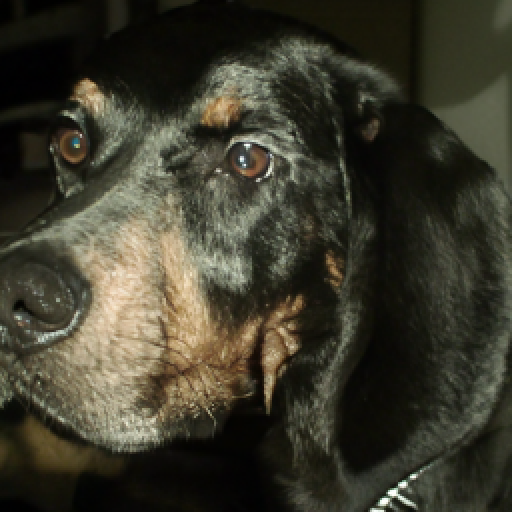} &
\includegraphics[width=0.2\textwidth]{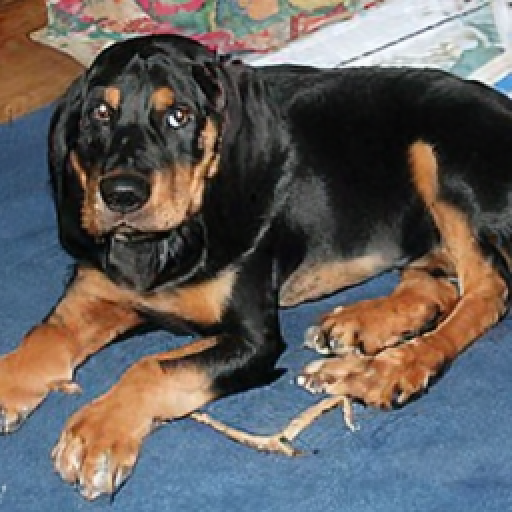} &
\includegraphics[width=0.2\textwidth]{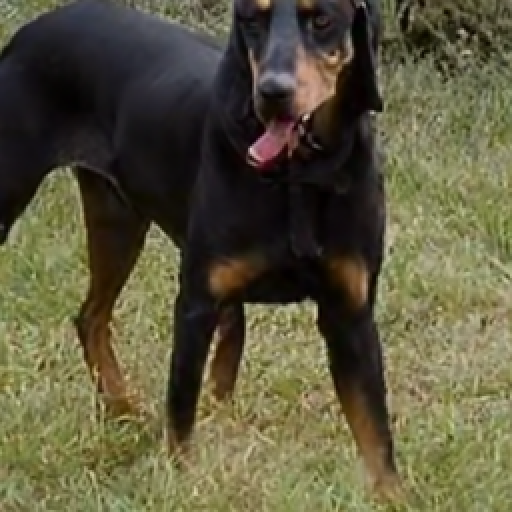} &
\includegraphics[width=0.2\textwidth]{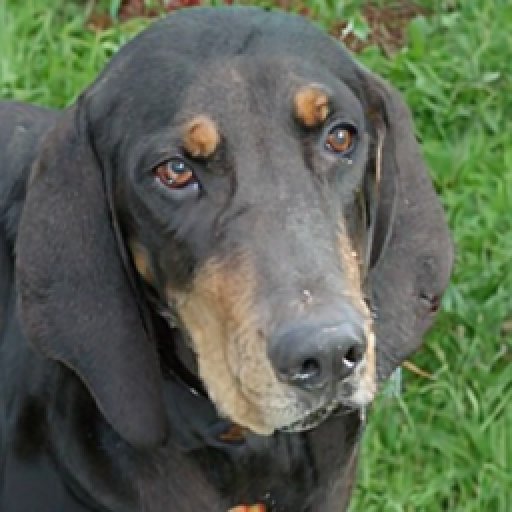} &
\includegraphics[width=0.2\textwidth]{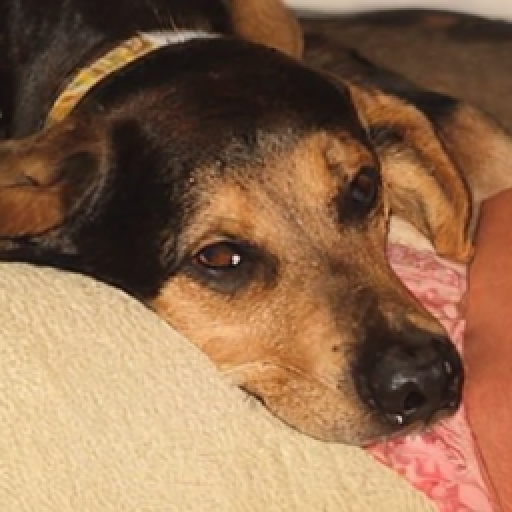} \\

\includegraphics[width=0.2\textwidth]{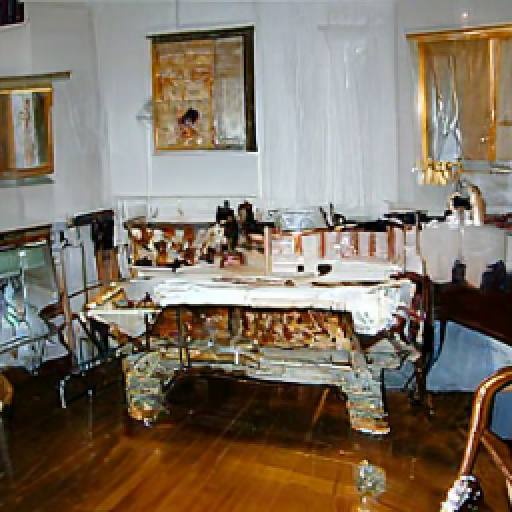} &
\includegraphics[width=0.2\textwidth]{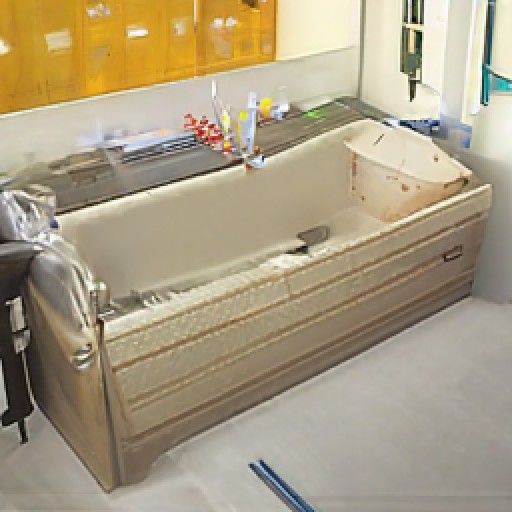} &
\includegraphics[width=0.2\textwidth]{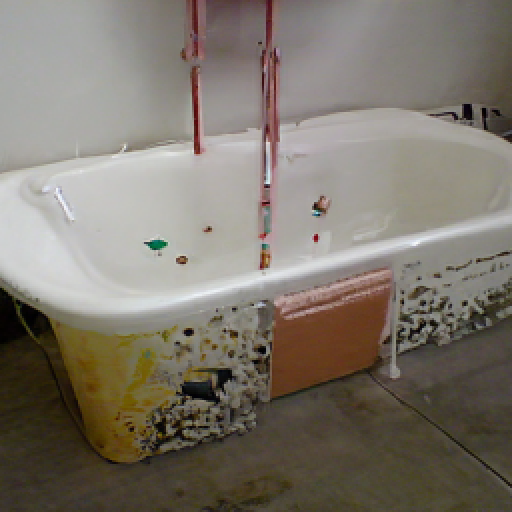} &
\includegraphics[width=0.2\textwidth]{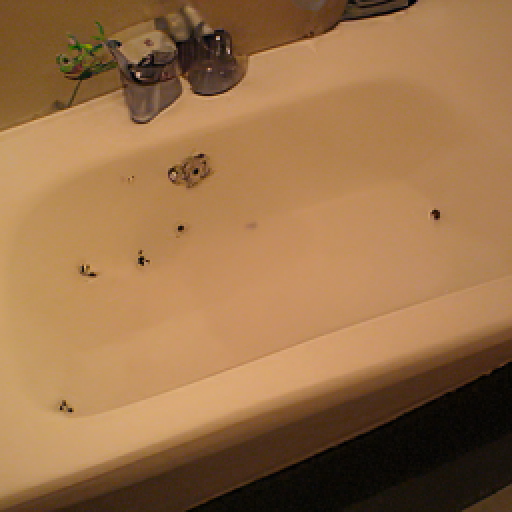} &
\includegraphics[width=0.2\textwidth]{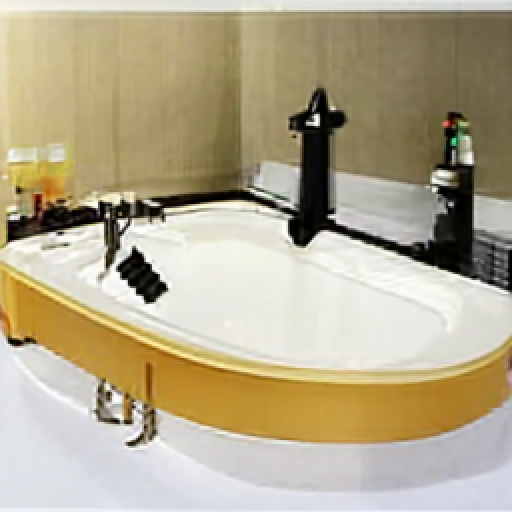}
\end{tabular}
\vspace*{-0.4cm}
\caption{\small Samples from classes with worst relative classification accuracy score. Each row represents a specific ImageNet class. Classes from top to bottom - Letter Opener (623), Plate (923), Overskirt (689), Tobacco Shop (860), Black-and-tan Coonhound (165) and Bathtub (435).}
\vspace*{-0.2cm}
\label{fig:cas_classwise_2}
\end{figure}

\begin{figure}[htbp] \centering
\includegraphics[width=\linewidth]{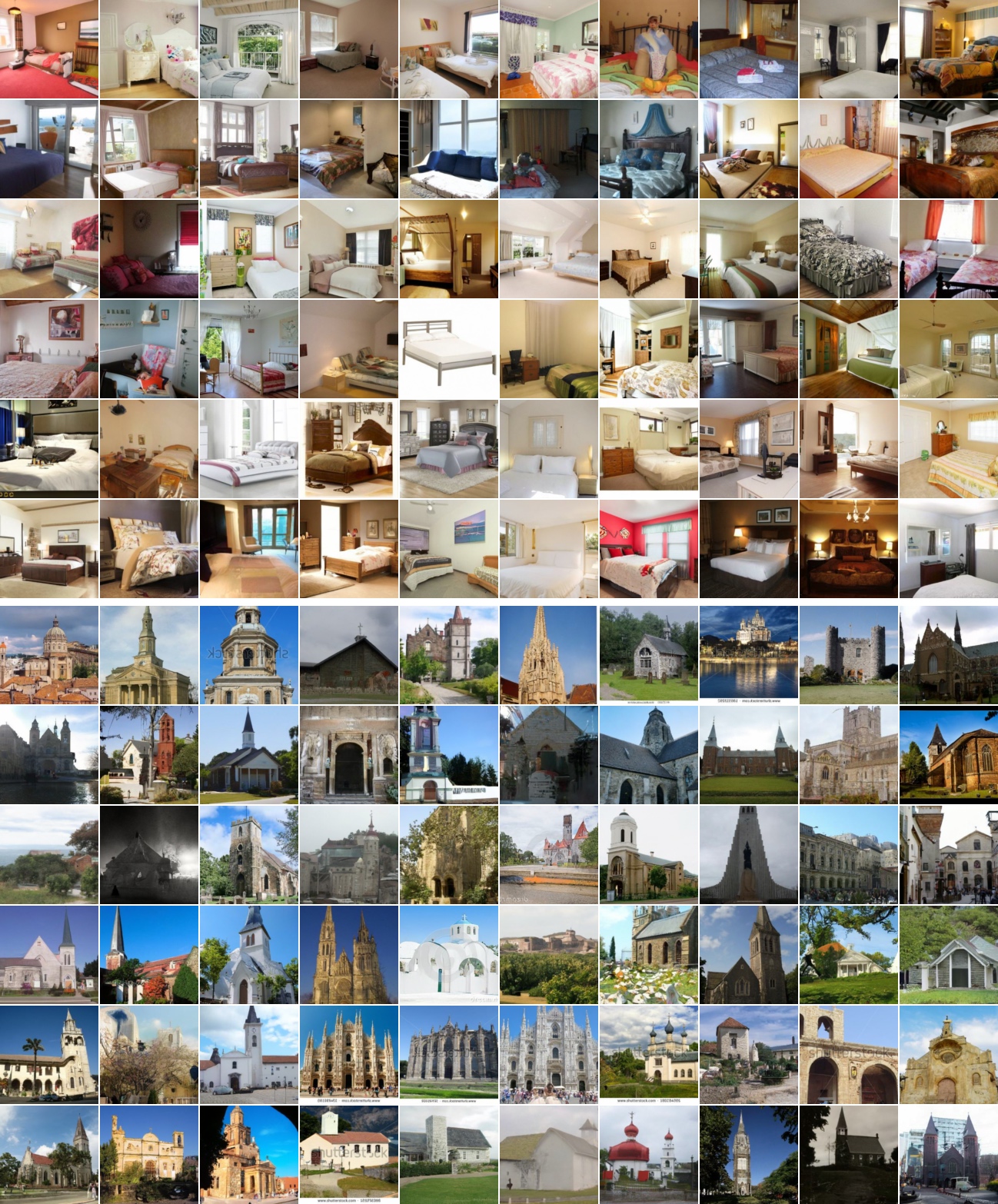}
\vspace*{-0.2cm}
\caption{\small Samples from LSUN 128x128: bedroom subset (first six rows) and church subset (last six rows).}
\label{fig:lsun_samples}
\end{figure}

\section{Hyperparameters}
\label{appendix:hyperparams}

\subsection{ImageNet}

Here we give the hyperparameters of the models in our ImageNet cascading pipelines. Each model in the pipeline is described by its diffusion process, its neural network architecture, and its training hyperparameters. Architecture hyperparameters, such as the base channel count and the list of channel multipliers per resolution, refer to hyperparameters of the U-Net in DDPM and related models~\citep{ho2020denoising,nichol2021improved,saharia2021image,salimans2017pixelcnn++}. The cosine noise schedule and the hybrid loss method of learning reverse process variances are from Improved DDPM~\citep{nichol2021improved}. Some models are conditioned on $\bar\alpha_t$ for post-training sampler tuning~\citep{chen2020wavegrad,saharia2021image}. \\

\setlist[itemize]{leftmargin=*}

{\small
\paragraph{32$\times$32 base model}\mbox{}\\
\begin{tabular}{p{0.45\textwidth}p{0.45\textwidth}}
\begin{itemize}
    \item Architecture
    \begin{itemize}
        \item Base channels: 256
        \item Channel multipliers: 1, 2, 3, 4
        \item Residual blocks per resolution: 6
        \item Attention resolutions: 8, 16
        \item Attention heads: 4
    \end{itemize}
    \item Training
    \begin{itemize}
        \item Optimizer: Adam
        \item Batch size: 2048
        \item Learning rate: 1e-4
        \item Steps: 700000
        \item Dropout: 0.1
        \item EMA: 0.9999
        \item Hardware: 256 TPU-v3 cores %
    \end{itemize}
\end{itemize} &
\begin{itemize}
    \item Diffusion
    \begin{itemize}
        \item Timesteps: 4000
        \item Noise schedule: cosine
        \item Learned variances: yes
        \item Loss: hybrid
    \end{itemize}
\end{itemize}
\end{tabular}
\vfill

\paragraph{32$\times$32$\rightarrow$64$\times$64 super-resolution}\mbox{}\\
\begin{tabular}{p{0.45\textwidth}p{0.45\textwidth}}
\begin{itemize}
    \item Architecture
    \begin{itemize}
        \item Base channels: 256
        \item Channel multipliers: 1, 2, 3, 4
        \item Residual blocks per resolution: 5
        \item Attention resolutions: 8, 16
        \item Attention heads: 4
    \end{itemize}

    \item Training
    \begin{itemize}
        \item Optimizer: Adam
        \item Batch size: 2048
        \item Learning rate: 1e-4
        \item Steps: 400000
        \item Dropout: 0.1
        \item EMA: 0.9999
        \item Hardware: 256 TPU-v3 cores %
    \end{itemize}
\end{itemize} &
\begin{itemize}
    \item Diffusion
    \begin{itemize}
        \item Timesteps: 4000
        \item Noise schedule: cosine
        \item Learned variances: yes
        \item Loss: hybrid
    \end{itemize}
\end{itemize}
\end{tabular}

\paragraph{64$\times$64$\rightarrow$128$\times$128 super-resolution}\mbox{}\\
\begin{tabular}{p{0.45\textwidth}p{0.45\textwidth}}
\begin{itemize}
    \item Architecture
    \begin{itemize}
        \item Base channels: 128
        \item Channel multipliers: 1, 2, 4, 8, 8
        \item Residual blocks per resolution: 3
        \item Attention resolutions: 16
        \item Attention heads: 1
    \end{itemize}
    \item Training
    \begin{itemize}
        \item Optimizer: Adam
        \item Batch size: 1024
        \item Learning rate: 1e-4
        \item Steps: 500000
        \item Dropout: 0.0
        \item EMA: 0.9999
        \item Hardware: 128 TPU-v3 cores %
    \end{itemize}
\end{itemize} &
\begin{itemize}
    \item Diffusion (Training)
    \begin{itemize}
        \item Timesteps: 2000
        \item Noise schedule: linear
        \item Learned variances: no
        \item Loss: simple
        \item Continuous noise conditioning
    \end{itemize}
     \item Diffusion (Inference)
    \begin{itemize}
        \item Timesteps: 100
        \item Noise schedule: linear
    \end{itemize}
\end{itemize}
\end{tabular}
\vfill

\paragraph{64$\times$64$\rightarrow$256$\times$256 super-resolution}\mbox{}\\
\begin{tabular}{p{0.45\textwidth}p{0.45\textwidth}}
\begin{itemize}
    \item Architecture
    \begin{itemize}
        \item Base channels: 128
        \item Channel multipliers: 1, 2, 4, 4, 8, 8
        \item Residual blocks per resolution: 3
        \item Attention resolutions: 16
        \item Attention heads: 1
    \end{itemize}
    \item Training
    \begin{itemize}
        \item Optimizer: Adam
        \item Batch size: 1024
        \item Learning rate: 1e-4
        \item Steps: 500000
        \item Dropout: 0.0
        \item EMA: 0.9999
        \item Hardware: 128 TPU-v3 cores %
    \end{itemize}
\end{itemize} &
\begin{itemize}
    \item Diffusion (Training)
    \begin{itemize}
        \item Timesteps: 2000
        \item Noise schedule: linear
        \item Learned variances: no
        \item Loss: simple
        \item Continuous noise conditioning
    \end{itemize}
    \item Diffusion (Inference)
    \begin{itemize}
        \item Timesteps: 100
        \item Noise schedule: linear
    \end{itemize}
\end{itemize}
\end{tabular}
} %
\newpage

\subsection{LSUN}

Here we give the hyperparameters of our LSUN Bedroom and Church pipelines. We used the same hyperparameters for both datasets.

{\small
\paragraph{64$\times$64 base model}\mbox{}\\
\begin{tabular}{p{0.45\textwidth}p{0.45\textwidth}}
\begin{itemize}
    \item Architecture
    \begin{itemize}
        \item Base channels: 128
        \item Channel multipliers: 1, 2, 3, 4
        \item Residual blocks per resolution: 3
        \item Attention resolutions: 8, 16, 32
        \item Attention heads dimension: 64
    \end{itemize}
    \item Training
    \begin{itemize}
        \item Optimizer: Adam
        \item Batch size: 2048
        \item Learning rate: 3e-4
        \item Steps: 100000
        \item Dropout: 0.1
        \item EMA: 0.9999
        \item Hardware: 64 TPU-v3 cores  %
    \end{itemize}
\end{itemize} &
\begin{itemize}
    \item Diffusion (Training)
    \begin{itemize}
        \item Noise schedule: cosine
        \item Learned variances: no
        \item Loss: simple
        \item Continuous noise conditioning
    \end{itemize}
    \item Diffusion (Inference)
    \begin{itemize}
        \item Timesteps: 256
        \item Noise schedule: cosine
    \end{itemize}
\end{itemize}
\end{tabular}

\paragraph{64$\times$64$\rightarrow$128$\times$128 super-resolution}\mbox{}\\
\begin{tabular}{p{0.45\textwidth}p{0.45\textwidth}}
\begin{itemize}
    \item Architecture
    \begin{itemize}
        \item Base channels: 64
        \item Channel multipliers: 1, 2, 4, 6, 8
        \item Residual blocks per resolution: 3
        \item Attention resolutions: 8, 16, 32
        \item Attention heads dimension: 64
    \end{itemize}
    \item Training
    \begin{itemize}
        \item Optimizer: Adam
        \item Batch size: 1024
        \item Learning rate: 2e-4
        \item Steps: 220000
        \item Dropout: 0.1
        \item EMA: 0.9999
        \item Hardware: 64 TPU-v3 cores  %
    \end{itemize}
\end{itemize} &
\begin{itemize}
     \item Diffusion (Training)
    \begin{itemize}
        \item Noise schedule: cosine
        \item Learned variances: no
        \item Loss: simple
        \item Continuous noise conditioning
    \end{itemize}
     \item Diffusion (Inference)
    \begin{itemize}
        \item Timesteps: 256
        \item Noise schedule: cosine
    \end{itemize}
\end{itemize}
\end{tabular}
} %

\newpage

\bibliography{main.bib}

\begin{thebibliography}{36}
\providecommand{\natexlab}[1]{#1}
\providecommand{\url}[1]{\texttt{#1}}
\expandafter\ifx\csname urlstyle\endcsname\relax
  \providecommand{\doi}[1]{doi: #1}\else
  \providecommand{\doi}{doi: \begingroup \urlstyle{rm}\Url}\fi

\bibitem[Bengio et~al.(2015)Bengio, Vinyals, Jaitly, and
  Shazeer]{bengio2015scheduled}
S.~Bengio, O.~Vinyals, N.~Jaitly, and N.~Shazeer.
\newblock Scheduled sampling for sequence prediction with recurrent neural
  networks.
\newblock \emph{{Advances in Neural Information Processing Systems}}, 2015.

\bibitem[Brock et~al.(2019)Brock, Donahue, and Simonyan]{brock2018large}
A.~Brock, J.~Donahue, and K.~Simonyan.
\newblock Large scale {GAN} training for high fidelity natural image synthesis.
\newblock In \emph{International Conference on Learning Representations}, 2019.

\bibitem[Chen et~al.(2021)Chen, Zhang, Zen, Weiss, Norouzi, and
  Chan]{chen2020wavegrad}
N.~Chen, Y.~Zhang, H.~Zen, R.~J. Weiss, M.~Norouzi, and W.~Chan.
\newblock {WaveGrad}: Estimating gradients for waveform generation.
\newblock \emph{{International Conference on Learning Representations}}, 2021.

\bibitem[De~Fauw et~al.(2019)De~Fauw, Dieleman, and
  Simonyan]{de2019hierarchical}
J.~De~Fauw, S.~Dieleman, and K.~Simonyan.
\newblock Hierarchical autoregressive image models with auxiliary decoders.
\newblock \emph{arXiv preprint arXiv:1903.04933}, 2019.

\bibitem[Dhariwal and Nichol(2021)]{dhariwal2021diffusion}
P.~Dhariwal and A.~Nichol.
\newblock Diffusion models beat gans on image synthesis.
\newblock \emph{arXiv preprint arXiv:2105.05233}, 2021.

\bibitem[Dinh et~al.(2017)Dinh, Sohl-Dickstein, and Bengio]{dinh2016density}
L.~Dinh, J.~Sohl-Dickstein, and S.~Bengio.
\newblock Density estimation using {R}eal {NVP}.
\newblock \emph{{International Conference on Learning}}, 2017.

\bibitem[Gao et~al.(2021)Gao, Song, Poole, Wu, and Kingma]{gao2020learning}
R.~Gao, Y.~Song, B.~Poole, Y.~N. Wu, and D.~P. Kingma.
\newblock Learning energy-based models by diffusion recovery likelihood.
\newblock \emph{{International Conference on Learning Representations}}, 2021.

\bibitem[Goodfellow et~al.(2014)Goodfellow, Pouget-Abadie, Mirza, Xu,
  Warde-Farley, Ozair, Courville, and Bengio]{goodfellow2014generative}
I.~Goodfellow, J.~Pouget-Abadie, M.~Mirza, B.~Xu, D.~Warde-Farley, S.~Ozair,
  A.~Courville, and Y.~Bengio.
\newblock Generative adversarial nets.
\newblock In \emph{Advances in Neural Information Processing Systems}, pages
  2672--2680, 2014.

\bibitem[Heusel et~al.(2017)Heusel, Ramsauer, Unterthiner, Nessler, and
  Hochreiter]{heusel2017gans}
M.~Heusel, H.~Ramsauer, T.~Unterthiner, B.~Nessler, and S.~Hochreiter.
\newblock {GANs} trained by a two time-scale update rule converge to a local
  {Nash} equilibrium.
\newblock In \emph{Advances in Neural Information Processing Systems}, pages
  6626--6637, 2017.

\bibitem[Ho et~al.(2019)Ho, Chen, Srinivas, Duan, and Abbeel]{ho2019flow++}
J.~Ho, X.~Chen, A.~Srinivas, Y.~Duan, and P.~Abbeel.
\newblock Flow++: Improving flow-based generative models with variational
  dequantization and architecture design.
\newblock In \emph{International Conference on Machine Learning}, 2019.

\bibitem[Ho et~al.(2020)Ho, Jain, and Abbeel]{ho2020denoising}
J.~Ho, A.~Jain, and P.~Abbeel.
\newblock Denoising diffusion probabilistic models.
\newblock In \emph{Advances in Neural Information Processing Systems}, pages
  6840--6851, 2020.

\bibitem[Jolicoeur-Martineau et~al.(2021)Jolicoeur-Martineau,
  Pich{\'e}-Taillefer, Combes, and Mitliagkas]{jolicoeur2020adversarial}
A.~Jolicoeur-Martineau, R.~Pich{\'e}-Taillefer, R.~T.~d. Combes, and
  I.~Mitliagkas.
\newblock Adversarial score matching and improved sampling for image
  generation.
\newblock \emph{{International Conference on Learning Representations}}, 2021.

\bibitem[Kingma and Dhariwal(2018)]{kingma2018glow}
D.~P. Kingma and P.~Dhariwal.
\newblock Glow: Generative flow with invertible 1x1 convolutions.
\newblock In \emph{Advances in Neural Information Processing Systems}, pages
  10215--10224, 2018.

\bibitem[Kingma and Welling(2014)]{kingma2013auto}
D.~P. Kingma and M.~Welling.
\newblock Auto-encoding variational {B}ayes.
\newblock \emph{{International Conference on Learning Representations}}, 2014.

\bibitem[Kong et~al.(2021)Kong, Ping, Huang, Zhao, and
  Catanzaro]{kong2020diffwave}
Z.~Kong, W.~Ping, J.~Huang, K.~Zhao, and B.~Catanzaro.
\newblock {DiffWave: A Versatile Diffusion Model for Audio Synthesis}.
\newblock \emph{{International Conference on Learning Representations}}, 2021.

\bibitem[Menick and Kalchbrenner(2019)]{menick2018generating}
J.~Menick and N.~Kalchbrenner.
\newblock Generating high fidelity images with subscale pixel networks and
  multidimensional upscaling.
\newblock In \emph{International Conference on Learning Representations}, 2019.

\bibitem[Nichol and Dhariwal(2021)]{nichol2021improved}
A.~Nichol and P.~Dhariwal.
\newblock Improved denoising diffusion probabilistic models.
\newblock \emph{{International Conference on Machine Learning}}, 2021.

\bibitem[Ranzato et~al.(2016)Ranzato, Chopra, Auli, and
  Zaremba]{ranzato2015sequence}
M.~Ranzato, S.~Chopra, M.~Auli, and W.~Zaremba.
\newblock Sequence level training with recurrent neural networks.
\newblock \emph{International Conference on Learning Representations}, 2016.

\bibitem[Ravuri and Vinyals(2019)]{ravuris2019cas}
S.~Ravuri and O.~Vinyals.
\newblock {Classification Accuracy Score for Conditional Generative Models}.
\newblock In \emph{Advances in Neural Information Processing Systems},
  volume~32, 2019.

\bibitem[Razavi et~al.(2019)Razavi, van~den Oord, and
  Vinyals]{razavi2019generating}
A.~Razavi, A.~van~den Oord, and O.~Vinyals.
\newblock Generating diverse high-fidelity images with {VQ-VAE-2}.
\newblock In \emph{Advances in Neural Information Processing Systems}, pages
  14837--14847, 2019.

\bibitem[Ronneberger et~al.(2015)Ronneberger, Fischer, and
  Brox]{ronneberger2015unet}
O.~Ronneberger, P.~Fischer, and T.~Brox.
\newblock {U-Net}: Convolutional networks for biomedical image segmentation.
\newblock In \emph{International Conference on Medical Image Computing and
  Computer-Assisted Intervention}, pages 234--241. Springer, 2015.

\bibitem[Russakovsky et~al.(2015)Russakovsky, Deng, Su, Krause, Satheesh, Ma,
  Huang, Karpathy, Khosla, Bernstein, et~al.]{russakovsky2015imagenet}
O.~Russakovsky, J.~Deng, H.~Su, J.~Krause, S.~Satheesh, S.~Ma, Z.~Huang,
  A.~Karpathy, A.~Khosla, M.~Bernstein, et~al.
\newblock {ImageNet} large scale visual recognition challenge.
\newblock \emph{International Journal of Computer Vision}, 115\penalty0
  (3):\penalty0 211--252, 2015.

\bibitem[Saharia et~al.(2021)Saharia, Ho, Chan, Salimans, Fleet, and
  Norouzi]{saharia2021image}
C.~Saharia, J.~Ho, W.~Chan, T.~Salimans, D.~J. Fleet, and M.~Norouzi.
\newblock Image super-resolution via iterative refinement.
\newblock \emph{arXiv preprint arXiv:2104.07636}, 2021.

\bibitem[Salimans et~al.(2016)Salimans, Goodfellow, Zaremba, Cheung, Radford,
  and Chen]{salimans2016improved}
T.~Salimans, I.~Goodfellow, W.~Zaremba, V.~Cheung, A.~Radford, and X.~Chen.
\newblock Improved techniques for training gans.
\newblock In \emph{Advances in Neural Information Processing Systems}, pages
  2234--2242, 2016.

\bibitem[Salimans et~al.(2017)Salimans, Karpathy, Chen, and
  Kingma]{salimans2017pixelcnn++}
T.~Salimans, A.~Karpathy, X.~Chen, and D.~P. Kingma.
\newblock Pixel{CNN}++: Improving the {PixelCNN} with discretized logistic
  mixture likelihood and other modifications.
\newblock In \emph{International Conference on Learning Representations}, 2017.

\bibitem[Sohl-Dickstein et~al.(2015)Sohl-Dickstein, Weiss, Maheswaranathan, and
  Ganguli]{sohl2015deep}
J.~Sohl-Dickstein, E.~Weiss, N.~Maheswaranathan, and S.~Ganguli.
\newblock Deep unsupervised learning using nonequilibrium thermodynamics.
\newblock In \emph{International Conference on Machine Learning}, pages
  2256--2265, 2015.

\bibitem[Song et~al.(2021{\natexlab{a}})Song, Meng, and
  Ermon]{song2020denoising}
J.~Song, C.~Meng, and S.~Ermon.
\newblock Denoising diffusion implicit models.
\newblock \emph{{International Conference on Learning Representations}},
  2021{\natexlab{a}}.

\bibitem[Song and Ermon(2019)]{song2019generative}
Y.~Song and S.~Ermon.
\newblock Generative modeling by estimating gradients of the data distribution.
\newblock In \emph{Advances in Neural Information Processing Systems}, pages
  11895--11907, 2019.

\bibitem[Song and Ermon(2020)]{song2020improved}
Y.~Song and S.~Ermon.
\newblock Improved techniques for training score-based generative.
\newblock \emph{{Advances in Neural Information Processing Systems}}, 2020.

\bibitem[Song et~al.(2021{\natexlab{b}})Song, Sohl-Dickstein, Kingma, Kumar,
  Ermon, and Poole]{song2020score}
Y.~Song, J.~Sohl-Dickstein, D.~P. Kingma, A.~Kumar, S.~Ermon, and B.~Poole.
\newblock Score-based generative modeling through stochastic differential
  equations.
\newblock \emph{{International Conference on Learning Representations}},
  2021{\natexlab{b}}.

\bibitem[van~den Oord et~al.(2016{\natexlab{a}})van~den Oord, Dieleman, Zen,
  Simonyan, Vinyals, Graves, Kalchbrenner, Senior, and
  Kavukcuoglu]{oord2016wavenet}
A.~van~den Oord, S.~Dieleman, H.~Zen, K.~Simonyan, O.~Vinyals, A.~Graves,
  N.~Kalchbrenner, A.~Senior, and K.~Kavukcuoglu.
\newblock {WaveNet}: A generative model for raw audio.
\newblock \emph{arXiv preprint arXiv:1609.03499}, 2016{\natexlab{a}}.

\bibitem[van~den Oord et~al.(2016{\natexlab{b}})van~den Oord, Kalchbrenner, and
  Kavukcuoglu]{oord2016pixel}
A.~van~den Oord, N.~Kalchbrenner, and K.~Kavukcuoglu.
\newblock Pixel recurrent neural networks.
\newblock \emph{International Conference on Machine Learning},
  2016{\natexlab{b}}.

\bibitem[van~den Oord et~al.(2016{\natexlab{c}})van~den Oord, Kalchbrenner,
  Vinyals, Espeholt, Graves, and Kavukcuoglu]{oord2016conditional}
A.~van~den Oord, N.~Kalchbrenner, O.~Vinyals, L.~Espeholt, A.~Graves, and
  K.~Kavukcuoglu.
\newblock Conditional image generation with {PixelCNN} decoders.
\newblock In \emph{Advances in Neural Information Processing Systems}, pages
  4790--4798, 2016{\natexlab{c}}.

\bibitem[van~den Oord et~al.(2017)van~den Oord, Vinyals, and
  Kavukcuoglu]{oord2017neural}
A.~van~den Oord, O.~Vinyals, and K.~Kavukcuoglu.
\newblock Neural discrete representation learning.
\newblock \emph{{Advances in Neural Information Processing Systems}}, 2017.

\bibitem[Wu et~al.(2019)Wu, Donahue, Balduzzi, Simonyan, and
  Lillicrap]{wu2019logan}
Y.~Wu, J.~Donahue, D.~Balduzzi, K.~Simonyan, and T.~Lillicrap.
\newblock Logan: Latent optimisation for generative adversarial networks.
\newblock \emph{arXiv preprint arXiv:1912.00953}, 2019.

\bibitem[Yu et~al.(2015)Yu, Zhang, Song, Seff, and Xiao]{yu15lsun}
F.~Yu, Y.~Zhang, S.~Song, A.~Seff, and J.~Xiao.
\newblock {LSUN}: Construction of a large-scale image dataset using deep
  learning with humans in the loop.
\newblock \emph{arXiv preprint arXiv:1506.03365}, 2015.

\end{thebibliography}

\end{document}